\documentclass{article}

\usepackage[preprint]{neurips_2026}

\usepackage[utf8]{inputenc} 
\usepackage[T1]{fontenc}    
\usepackage{hyperref}       
\usepackage{url}            
\usepackage{tcolorbox}
\usepackage{booktabs}       
\usepackage{amsmath}        
\usepackage{amsfonts}       
\usepackage{amssymb}        
\usepackage{nicefrac}       
\usepackage{microtype}      
\usepackage[table]{xcolor}  
\usepackage{graphicx}       
\usepackage{longtable}      
\usepackage{tabularx}       
\usepackage{makecell}       
\usepackage{caption}        
\usepackage{subcaption}     
\usepackage{wrapfig}
\usepackage{pifont} 
\usepackage{multirow}
\definecolor{darkgreen}{RGB}{0,150,0}
\definecolor{darkred}{RGB}{255,0,0}
\definecolor{tabheader}{RGB}{218,228,242}
\definecolor{tabrowalt}{RGB}{240,244,250}
\definecolor{tabours}{RGB}{255,243,205}
\definecolor{tabrule}{RGB}{38,70,108}
\definecolor{cellyes}{RGB}{220,240,225}
\definecolor{cellno}{RGB}{248,220,220}
\definecolor{cellpartial}{RGB}{252,236,210}
\definecolor{yescell}{HTML}{E1EFE5}     
\definecolor{yesmark}{HTML}{1F7A47}     
\definecolor{nomark}{HTML}{C9CDD2}      
\definecolor{ourscell}{HTML}{FBE3C5}    
\definecolor{oursmark}{HTML}{AE540B}    
\definecolor{oursname}{HTML}{FFF6E8}    
\definecolor{group1}{HTML}{ECF2FA}      
\definecolor{group2}{HTML}{F4EDF7}      
\definecolor{citegray}{HTML}{8A8E96}
\definecolor{sectionbg}{HTML}{E8EEF4}

\definecolor{tierT}{HTML}{A5C8AB}      
\definecolor{tierH}{HTML}{C9DFCC}      
\definecolor{tierM}{HTML}{E6EFE7}      
\definecolor{oursrow}{HTML}{FFF4E5}    
\definecolor{zebrarow}{HTML}{E4EAF2}   

\newcommand{\cT}{\cellcolor{tierT}}
\newcommand{\cH}{\cellcolor{tierH}}
\newcommand{\cM}{\cellcolor{tierM}}

\title{A Benchmark for Omni-Modal Reasoning in Long Videos}

\author{%
  Mohammed Irfan Kurpath\thanks{Equal contribution.}\,$^{1}$ \quad
  Jaseel Muhammad Kaithakkodan\footnotemark[1]\,$^{1}$ \\
  \textbf{Jinxing Zhou}$^{1}$ \quad
  \textbf{Sahal Shaji Mullappilly}$^{1}$ \quad
  \textbf{Mohammad Almansoori}$^{1}$ \\
  \textbf{Noor Ahsan}$^{1}$ \quad
  \textbf{Beknur Kalmakhanbet}$^{1}$ \quad
  \textbf{Sambal Shikhar}$^{1}$ \quad
  \textbf{Rishabh Lalla}$^{1}$ \\
  \textbf{Jean Lahoud}$^{1}$ \quad
  \textbf{Mariette Awad}$^{2}$ \quad
  \textbf{Fahad Shahbaz Khan}$^{1,3}$ \\
  \textbf{Salman Khan}$^{1}$ \quad
  \textbf{Rao Muhammad Anwer}$^{1}$ \quad
  \textbf{Hisham Cholakkal}$^{1}$ \\
  \vspace{4pt} \\
  \normalfont\normalsize
  $^{1}$Mohamed Bin Zayed University of Artificial Intelligence (MBZUAI), UAE \\
  $^{2}$American University of Beirut \quad
  $^{3}$Link\"oping University \\
}

\begin{document}

\maketitle
\vspace{-5ex}

\begin{figure}[!h]
  \centering
  {\includegraphics[width=\textwidth]{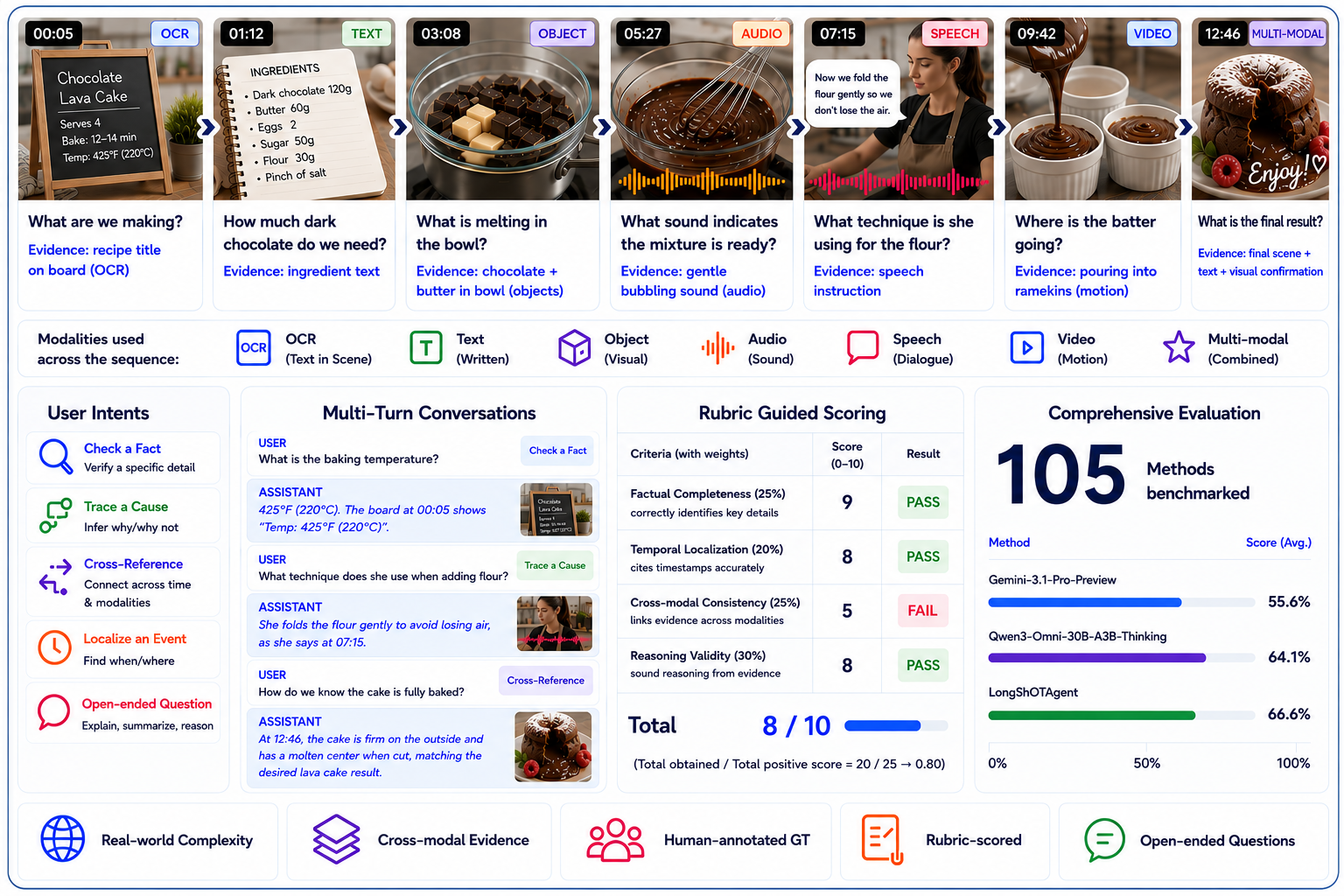}}
  \vspace{-4ex}
  \caption{\textbf{LongShOTBench overview.} A representative cooking-video sequence demonstrates the benchmark's omni-modal reasoning challenge, where answering questions requires combining OCR, text, visual, speech, audio, motion, and temporal evidence distributed across a long video. LongShOTBench evaluates intent-driven single- and multi-turn QA using rubric-guided scoring against human-annotated ground truth and provides comprehensive evaluation across 105 methods.}
  \label{fig:teaser-fig}
  \vspace{-2ex}
\end{figure}

\begin{abstract}
\vspace{-1ex}
Long-form omni-modal video understanding requires models to integrate vision, speech, and ambient audio with coherent long-context reasoning. Existing video benchmarks often trade off temporal scale, modality coverage, open-ended interaction, and interpretable scoring.
To address this gap, we introduce \textbf{LongShOTBench}, a long video understanding benchmark designed around three coupled goals: holistic omni-modal integration, intent-driven open-ended interaction, and rubric-level diagnosis.
It constructs single- and multi-turn questions from practical viewing scenarios, involving systematical tasks for probing visual, speech, ambient-audio, temporal, and cross-modal reasoning. Each item includes a reference answer and a weighted criterion-level rubric, enabling evaluation to identify which perceptual facts, temporal links, modality-grounding requirements, and reasoning steps are satisfied or missed. All samples are manually verified and corrected to improve grounding, clarity, and rubric reliability. 
We also introduce \textbf{LongShOTAgent}, a training-free omni-modal evidence-seeking agent that couples full-video preprocessing with targeted retrieval, query-adaptive segment refinement, and explicit claim verification over visual, speech, and non-speech audio evidence. Its iterative search-refine-verify loop exposes intermediate evidence and lets modality-specific specialists re-analyze relevant moments before answering. 
We perform comprehensive evaluation of 105 video-capable models spanning open-source omni-modal models, vision-language systems, audio LLMs, agentic pipelines and closed-source APIs. Across this broad evaluation, current MLLMs remain far from saturating LongShOTBench, while our LongShOTAgent emerges as the strongest training-free system, reaching 66.64\% overall. 
By releasing the benchmark, leaderboard, and agentic method, our work provides the community with a shared, interpretable testbed for evaluating and advancing long-form omni-modal video reasoning. Code, data, the benchmark, and the leaderboard are available at \url{https://longshot.cvmbzuai.com/}.
\end{abstract}

\section{Introduction}
\label{sec:intro}

Video is among the most demanding settings for multimodal AI. Unlike a static image or a transcript, a video weaves together what we see, what is said, and what we hear into a single timeline. Any system that claims to understand it must integrate all three modalities as events unfold, making video the natural stress test for today's omni-modal large language models~\citep{bai2025qwen2,yang2025qwen3,wang2025internvl3_5,google2024gemini,
xu2025qwen25omni,ai2025mingomni}.

Yet the way we evaluate these models has not kept pace with their growing capabilities. Earlier benchmarks focused on short, trimmed clips and proved useful for testing event recognition and brief temporal reasoning~\citep{li2024mvbench}, but they no longer capture the demands placed on video understanding systems in realistic use cases. In practice, questions often require connecting events that are minutes apart, resolving a spoken reference to an off-screen event, or recognizing that an ambient sound changes the meaning of an on-screen action. Motivated by this gap, recent work has moved toward longer videos, but extending duration alone has not been enough. Most long-video benchmarks still rely primarily on visual frames or subtitle-level speech and leave non-speech audio largely untouched~\citep{Wang2024LVBench}. Those that do incorporate richer audio-visual-language signals tend to focus on shorter clips or narrow tasks such as captioning and segment retrieval~\citep{fu2024video}.

Beyond the \textit{modality-duration trade-off}, {question construction} also limits what existing benchmarks can reveal. Many benchmarks rely on fixed templates or predefined task taxonomies\citep{li2024mvbench,jang2017tgif}, while even scenario-grounded benchmarks often restrict outputs to short factoids or predetermined tool sequences. In real video-watching scenarios, however, human queries are usually driven by diverse goals and intents, such as seeking facts, understanding causes, comparing events, or planning actions. Even for the same video, different users may ask substantially different questions depending on their needs. Existing long-video benchmarks often overlook this \textit{goal-driven nature of user queries}. Moreover, most adopt a single-turn Q\&A format, making it difficult to assess whether models can maintain context, resolve follow-up questions, and reason consistently across
\textit{multi-turn interactions}.

\newcommand{\yY}{\textcolor{yesmark}{\ding{51}}}                              
\newcommand{\yN}{\textcolor{nomark}{\rule[0.45ex]{0.6em}{0.9pt}}}            
\newcommand{\yS}{\textcolor{yesmark}{\ding{51}}\textsuperscript{\,$*$}}      
\newcommand{\bcite}[1]{{\scriptsize\textcolor{citegray}{\citep{#1}}}}

\begin{table}[t]
\centering
\small
\setlength{\tabcolsep}{6.5pt}
\renewcommand{\arraystretch}{1.18}
\caption{\textbf{Long-video benchmark comparison.} LongShOTBench is the only benchmark to pair full omni-modal coverage (visual, audio, speech) with open-ended, multi-turn, intent-driven evaluation and weighted criterion-level rubrics. \yS\ marks speech available only through subtitles.}
\label{tab:benchmark_comparison}
\begin{tabular}{l ccc cccc}
\toprule
\rowcolor{tabheader}
\textbf{Benchmark} & \textbf{Visual} & \textbf{Audio} & \textbf{Speech}
 & \shortstack{\textbf{Open-}\\\textbf{Ended}} & \shortstack{\textbf{Multi-}\\\textbf{Turn}}
 & \shortstack{\textbf{Intent-}\\\textbf{Driven}} & \textbf{Rubrics} \\
\midrule
\rowcolor{tabours}\textbf{LongShOTBench} & \yY & \yY & \yY & \yY & \yY & \yY & \yY \\
\rowcolor{zebrarow}VideoOdyssey\bcite{he2026videoodysseybenchmarkultralongcontextomnimodal}   & \yY & \yY & \yY & \yN & \yN & \yN & \yN \\
LVOmniBench\bcite{tao2026lvomnibenchpioneeringlongaudiovideo}              & \yY & \yY & \yY & \yN & \yN & \yN & \yN \\
\rowcolor{zebrarow}WorldSense\bcite{hong2025worldsense}                    & \yY & \yY & \yY & \yN & \yN & \yN & \yN \\
OmniVideoBench\bcite{li2026omnivideobenchaudiovisualunderstandingevaluation} & \yY & \yY & \yY & \yN & \yN & \yN & \yN \\
\rowcolor{zebrarow}Daily-Omni\bcite{zhou2025dailyomni}                     & \yY & \yY & \yY & \yN & \yN & \yN & \yN \\
TriSense-2M\bcite{li2025TriSense}                                          & \yY & \yY & \yY & \yY & \yN & \yN & \yN \\
\rowcolor{zebrarow}LongVALE\bcite{geng2025longvale}                        & \yY & \yY & \yY & \yY & \yN & \yN & \yN \\
Video-MME\bcite{fu2024video}                                              & \yY & \yY & \yY & \yN & \yN & \yN & \yN \\
\rowcolor{zebrarow}InfiniBench\bcite{ataallah2025infinibenchbenchmarklargemultimodal} & \yY & \yN & \yS & \yY & \yN & \yN & \yN \\
Video-Holmes\bcite{cheng2025video}                                        & \yY & \yY & \yY & \yN & \yN & \yN & \yN \\
\rowcolor{zebrarow}MoVQA\bcite{zhang2025lvbench}                           & \yY & \yN & \yS & \yN & \yN & \yN & \yN \\
LVBench\bcite{Wang2024LVBench}                                             & \yY & \yN & \yN & \yN & \yN & \yN & \yN \\
\rowcolor{zebrarow}SVBench\bcite{yang2025svbenchbenchmarktemporalmultiturn} & \yY & \yN & \yN & \yY & \yY & \yN & \yN \\
MLVU\bcite{zhou2024mlvu}                                                   & \yY & \yN & \yN & \yY & \yN & \yN & \yN \\
\rowcolor{zebrarow}MovieChat\bcite{song2024moviechatdensetokensparse}      & \yY & \yN & \yN & \yY & \yN & \yN & \yN \\
LongVideoBench\bcite{Wu2024LongVideoBench}                                 & \yY & \yN & \yS & \yN & \yN & \yN & \yN \\
\rowcolor{zebrarow}EgoSchema\bcite{mangalam2023egoschema}                  & \yY & \yN & \yN & \yN & \yN & \yN & \yN \\
MV-Bench\bcite{li2024mvbench}                                              & \yY & \yN & \yN & \yN & \yN & \yN & \yN \\
\bottomrule
\end{tabular}
\end{table}

Another notable limitation lies in how answers are scored.
Multiple-choice question (MCQ) formats remain widely used, reducing evaluation to option accuracy, while open-ended setups are typically handled by a single LLM that produces a holistic free-form score, a setting where reliability concerns are well documented~\citep{gu2024survey}. 
In either case, the scoring mechanism provides limited insights into the source of failure: whether a model missed visual evidence, misheard speech, ignored an ambient cue, mislocalized an event in time, or reasoned poorly over otherwise correct observations~\citep{mangalam2023egoschema,zhang2025lvbench}.
Moreover, we find that MCQ scoring can create an ``{illusion of comprehension}'', substantially overestimating model capability: on Video-MME~\citep{fu2024video} and WorldSense~\citep{hong2025worldsense}, simply removing options or perturbing their order causes accuracy drops exceeding 20\% for strong monolithic models, revealing heavy reliance on option-level shortcuts rather than genuine understanding (Appendix~\ref{sec:robustness}).

To address these shortcomings, we introduce \textbf{LongShOTBench}, a diagnostic benchmark for omni-modal reasoning over long videos. LongShOTBench couples extended temporal contexts with vision, speech, and non-speech audio, and reframes question generation around \emph{user intents}: we first elicit plausible viewing scenarios, such as checking a fact, tracing a cause, or comparing options, and then derive both single-turn queries and multi-turn dialogues from those scenarios. Rather than collapsing correctness into a single number, each question is paired with a weighted, criterion-level rubric that decomposes the expected answer into verifiable components such as factual completeness, temporal localization, and cross-modal consistency; an LLM \emph{verifier} performs an atomic binary check per criterion, comparing the candidate answer against the human-validated reference answer rather than relying on the verifier's own world knowledge, which enables partial credit and traces every score to specific evidence. The benchmark is built through a multi-stage, human-validated pipeline. LongShOTBench therefore addresses the limitations of prior work by contributing three key elements: long-form omni-modal QA, intent-driven interactions spanning single- and multi-turn dialogue, and rubric-guided diagnostic scoring.

Alongside the benchmark, we provide \textbf{LongShOTAgent}, a simple, training-free agentic baseline that illustrates what structured retrieval and modality-aware processing can offer on long-video tasks. LongShOTAgent follows an iterative \emph{search-refine-verify} loop: after preprocessing a video into a reusable store of visual, speech, and audio signals, an orchestrator retrieves candidate moments at query time, re-examines short segments with modality-specific tools, and checks intermediate claims before producing a final answer. It is intended as a reference point that highlights the role of structured omni-modal analysis in hour-scale video understanding.

We evaluate \textbf{105} video-capable models on LongShOTBench, spanning closed-source APIs, open-source omni-modal, vision-language, and audio-only models, as well as existing agentic methods (full leaderboard in Table~\ref{tab:full_benchmark_table}). The results reveal substantial headroom across the board. The strongest closed-source system, Gemini 3.1 Pro Preview, reaches only 55.63\%, and the best open-source omni-modal model, Qwen3-Omni-30B-A3B-Thinking, reaches 64.05\%. Notably, our training-free LongShOTAgent reaches 66.64\%, suggesting that structured use of visual, speech, and audio evidence can be as important as raw model capacity for hour-length video understanding. To validate the rubric-guided evaluation itself, we conduct a human-alignment study with four expert annotators across 400 criterion decisions. The three-verifier ensemble achieves Cohen's $\kappa = 0.75$ and Pearson $r = 0.92$ against the human consensus, matching the leave-one-out inter-annotator reference ($\kappa \in [0.61, 0.76]$, $r \in [0.78, 0.90]$), and preserves the human model ranking up to near-tied adjacent pairs (Spearman $\rho = 0.80$, all flips at score gaps $<\!0.04$). These results indicate that rubric-guided scoring system can approximate expert judgment while retaining criterion-level traceability, making LongShOTBench practical for scalable and interpretable evaluation.

\section{Related Work}
\label{sec:related_works}

\textbf{Long-Video Benchmarks.}
Benchmarks have progressively moved from short trimmed clips~\citep{Li2022Learning,li2024mvbench,patraucean2023perception} to videos lasting tens of minutes or longer. LVBench~\citep{Wang2024LVBench}, MoVQA\footnote{\citet{zhang2025lvbench} published their dataset under the name `LvBench', which shares a name with but is entirely distinct from the `LVBench' by \citet{Wang2024LVBench}. We use MoVQA to avoid confusion.}~\citep{zhang2025lvbench}, and InfiniBench~\citep{ataallah2025infinibenchbenchmarklargemultimodal} extend duration but operate almost entirely over the visual stream and subtitles. LongVALE~\citep{geng2025longvale} and TriSense-2M~\citep{li2025TriSense} add speech and audio, yet average only 4 and 15 minutes respectively and focus on segment captioning and retrieval. No existing benchmark combines long temporal contexts, full omni-modal coverage, and higher-order reasoning evaluation (Table~\ref{tab:benchmark_comparison}).
Moreover, most benchmarks generate questions through human authoring or fixed templates. ALLVB~\citep{tan2025allvb} relies on predefined task taxonomies. GAIA~\citep{mialon2023gaia} and GTA~\citep{wang2024gta} restrict outputs to factoid answers or predetermined tool sequences.
Instead, our LongShOTBench uses scenario-driven, intent-aware generation that covers both single-turn and multi-turn interactions.
LongShOTBench also provides weighted, criterion-level rubrics that offer partial credit and make each score traceable to specific evidence dimensions.

\textbf{Agentic Video Understanding.}
An alternative to processing video in a single forward pass is to let an LLM orchestrate retrieval and tool calls over the video. VideoMind~\citep{liu2025videomind} introduces a chain-of-LoRA agent for long-video reasoning, Vgent~\citep{shen2025vgent} performs graph-based retrieval-reasoning-augmented generation, and Video-RAG~\citep{luo2024video} pairs visually-aligned retrieval with a base VLM. While all three target long videos, they operate predominantly over the visual stream and, where available, ASR transcripts, without first-class treatment of non-speech audio. On LongShOTBench, all three score below 15\% (Table~\ref{tab:full_benchmark_table}), indicating that visual-centric agentic recipes do not yet generalize to hour-long omni-modal scenarios.
\vspace{-2ex}
\begin{figure}[t]
  \centering
  {\includegraphics[width=\textwidth]{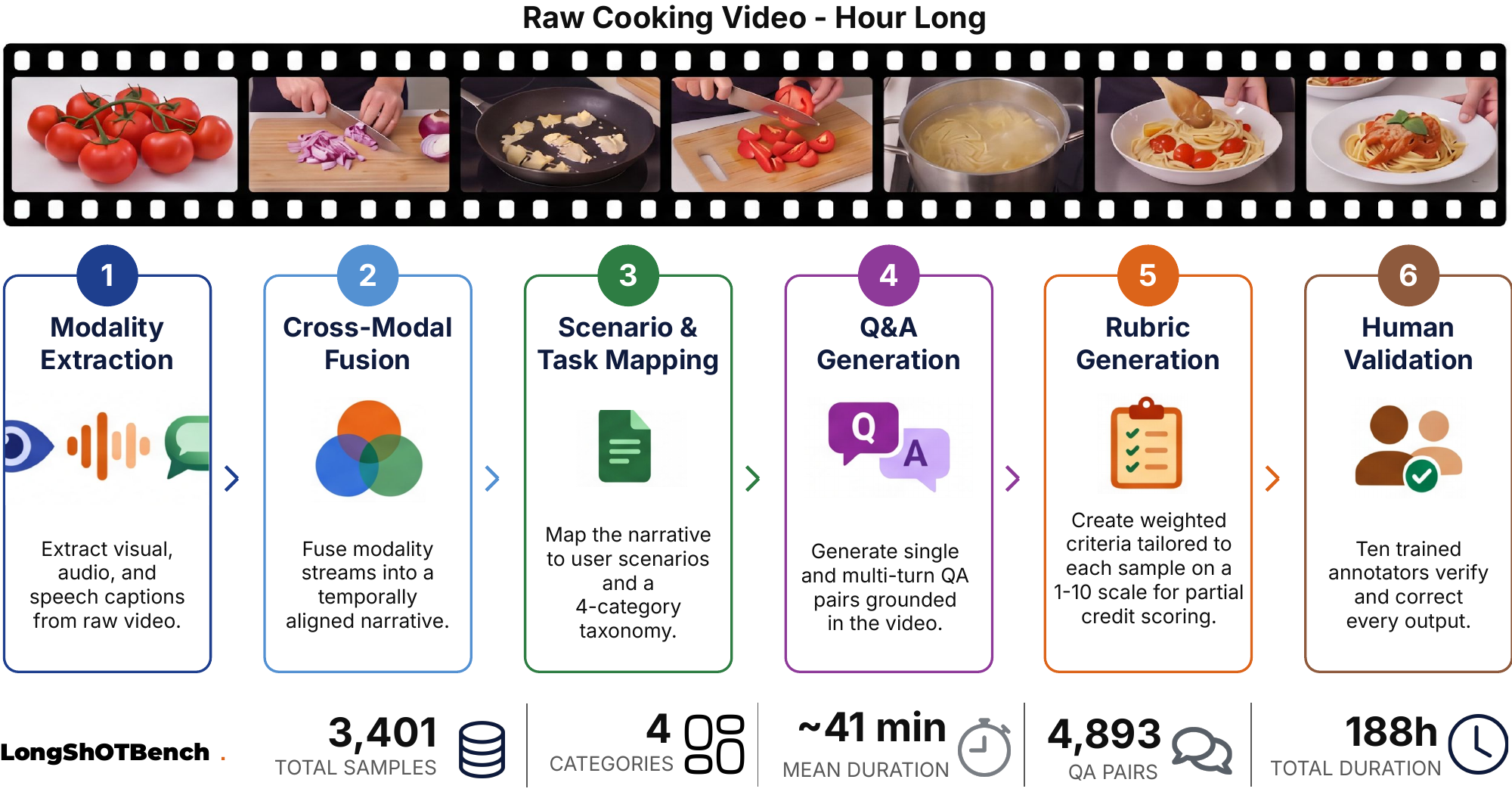}}
  \vspace{-4ex}
  \caption{\textbf{Construction pipeline of LongShOTBench.} Starting from raw long-form videos, we extract speech, visual, and audio cues, then fuse them into segment-wise aligned omni-modal metadata. The distilled metadata drives scenario and task mapping, followed by single- and multi-turn Q\&A generation and verifiable rubric construction. Finally, human validators review and correct the Q\&A pairs and tailored rubrics, ensuring a grounded and reliable benchmark.}
  \label{fig:pipeline-overview}
  \vspace{-4ex}
\end{figure}

\section{LongShOTBench}
\vspace{-1ex}
\label{sec:constructing_starbench}

LongShOTBench converts raw videos into structured QA pairs through a sequential pipeline: omni-modal caption generation, question design, answer generation, hierarchical rubric generation, and human validation (Fig.~\ref{fig:pipeline-overview}). The resulting dataset comprises 274 long-form videos averaging ${\sim}41$ minutes (median 40.9, range 7.8-89.7\,min), totaling ${\sim}188$ hours of content. These are organized into 3{,}401 evaluation samples containing 4{,}893 question-answer turns across single-turn and multi-turn interactions, roughly 12 samples and 18 questions per video. We describe each stage below.

\vspace{-2ex}
\subsection{Omni-modal Caption Generation}

\vspace{-1ex}
\textbf{Video Collection.}
We construct a benchmark of 274 long-form videos by repurposing and re-annotating existing public video collections~\citep{fu2024video}, carefully adhering to the license terms of all source material, and supplemented with additional videos manually curated from YouTube to support more complex omni-modal reasoning scenarios (Fig.~\ref{fig:duration-distribution}).
\vspace{-1ex}

\textbf{Modality-Specific Captioning.}
Each video is split into segments using speech activity, with silent intervals serving as natural boundaries. Three specialist models then process each segment in parallel. Speech is transcribed into time-stamped captions using a large speech recognition model (Whisper-Large-v3~\citep{radford2023robust}). Visual content is described through dense, segment-level scene captions generated by passing representative frames to a vision-language model (Qwen2.5-VL-32B-Instruct~\citep{bai2025qwen2}). Non-speech audio events such as music, applause, and environmental sounds are tagged by an audio understanding model (Audio Flamingo 3~\citep{goel2025audio}). This parallel strategy preserves temporal granularity even in segments with little or no speech.

\vspace{-1ex}
\textbf{Cross-Modal Integration.}
The three caption streams are then merged by a language model (Qwen3-30B-A3B-Instruct-2507~\citep{yang2025qwen3}) into coherent high-level narratives that reconcile cross-modal conflicts and align events temporally.

\vspace{-1ex}
\subsection{Question Design}

\vspace{-1ex}
\textbf{Intent Framing.}
Rather than generating questions directly, we first perform \textit{intent framing} to mimic users' practical but diverse intents, for example, a prospective buyer checking battery life versus an enthusiast comparing camera performance in a phone review. Using the cross-modal summaries, we generate up to three diverse scenarios per video, each containing up to two tasks.

\vspace{-1ex}
\textbf{Task Mapping.}
To ensure balanced coverage, every question is mapped onto a four-category taxonomy: (1)~\textit{Core Perception}, covering entities, events, temporal cues, and audio recognition, (2)~\textit{Information}, assessing retrieval, summarization, and interpretation, (3)~\textit{Multimodal}, probing integration and alignment across modalities, and (4)~\textit{Reasoning}, targeting causal, quantitative, compositional, and comparative inference.

\vspace{-1ex}
\textbf{Question Generation.}
Within each scenario, we generate both single-turn queries and multi-turn dialogues. Difficulty is controlled on a five-level scale, skewing toward moderate-to-hard: 23.6\% at Levels~1-2, 44.0\% at Level~3, and 32.3\% at Levels~4-5 (per-category breakdown in Fig.~\ref{fig:difficulty-ridge}; modality count vs.\ difficulty in Fig.~\ref{fig:difficulty-modality}). All questions are explicitly grounded in observable omni-modal cues.

\vspace{-2ex}
\subsection{Answer Generation}
\vspace{-1ex}

Given the omni-modal captions and generated questions, we prompt Qwen3-30B-A3B-Instruct-2507~\citep{yang2025qwen3} to produce reference answers. The generation is guided by three principles: match response detail to question complexity, integrate evidence from multiple modalities when relevant, and abstain explicitly when the available metadata is insufficient rather than hallucinate. Answers remain grounded and concise, avoiding speculation about unobservable states.

\vspace{-2ex}
\subsection{Hierarchical Rubric Generation}
\vspace{-1ex}

Each QA pair is accompanied by a rubric consisting of multiple independently verifiable criteria, each targeting a specific fact, event, or relationship from the video. Criteria are weighted on a 10-point priority scale: high-priority items (7-10) capture essential facts, medium-priority items (4-6) supporting details, and low-priority items (1-3) contextual enrichment. During evaluation, an LLM verifier checks each criterion $c_i$ against the gold reference answer and marks it satisfied or not based on the candidate's match to the gold, and the sample score is computed as the fraction of total positive weight recovered by satisfied criteria, scaled to a 0-100 range:
{
\setlength{\abovedisplayskip}{4pt}
\setlength{\belowdisplayskip}{4pt}
\begin{equation}
\label{eq:rubric_score}
\text{Score} = \frac{\sum_{i} w_i \cdot \mathbf{1}[c_i]}{\sum_{i} w_i} 
\times 100
\end{equation}
}
where $w_i$ is the corresponding weight. This avoids asking the verifier for direct numeric scores, which tend to be inconsistent~\citep{gu2024survey}, and instead leverages the strength of LLMs at binary textual matching against a fixed gold reference. The result is a scoring process that supports partial credit and links every score to the specific evidence it reflects.

\vspace{-2ex}
\subsection{Human Validation}
\vspace{-1ex}

All generated samples pass through a structured validation stage carried out by ten trained annotators with backgrounds in linguistics and computer vision. Annotators completed a calibration program covering the taxonomy, rubric format, and scoring guidelines before beginning review. During validation, samples were discarded if the source video was corrupted, ambiguous, or lacked sufficient omni-modal information. Question phrasing was refined to remove bias, answers were corrected for missing details or hallucinated content, and rubric criteria were revised to eliminate redundancy and improve verifiability. This process ensures that the final dataset maintains a consistent standard of naturalness, clarity, and factual grounding across all 274 videos. In total, this human validation effort amounted to over 450 annotation hours.

\begin{figure}[t]
    \centering{\includegraphics[width=\linewidth]{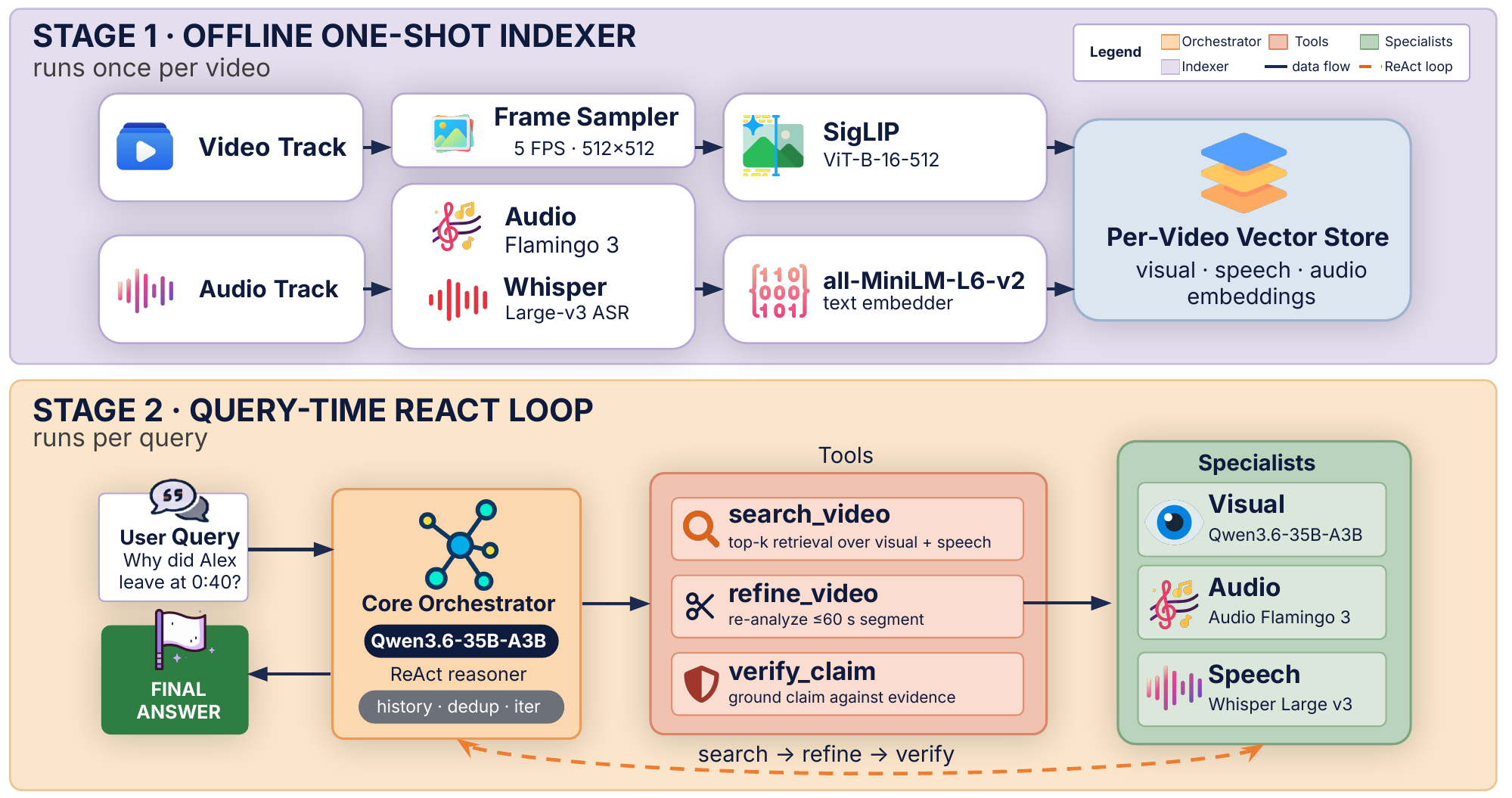}}
    \vspace{-4ex}
    \caption{\textbf{LongShOTAgent architecture.} The indexer embeds video frames, speech transcripts and audio understanding into a per-video vector store. At query time, the orchestrator LLM analyzes multimodal evidence by issuing tool calls in a ReAct-style loop: \texttt{search\_video} for semantic retrieval, \texttt{refine\_video} for fine-grained segment-level re-analysis with specialists, and \texttt{verify\_claim} for evidence grounding.}
    \label{fig:agentic-pipeline}
    \vspace{-4ex}
\end{figure}

\vspace{-1ex}
\section{LongShOTAgent}
\label{sec:staragent_pipeline}
\vspace{-1ex}

Alongside the benchmark, we provide LongShOTAgent, a training-free agentic baseline that serves as a reference point for LongShOTBench (Fig.~\ref{fig:agentic-pipeline}). Rather than processing an entire video in a single forward pass, LongShOTAgent first converts the video into a searchable multimodal store through one-time preprocessing: frames are embedded with SigLIP~\citep{zhai2023sigmoid}, speech is transcribed with Whisper~\citep{radford2023robust}, and non-speech audio is described and embedded alongside, populating a per-video vector database. At query time, an LLM orchestrator operates in a ReAct-style loop over three tools: \texttt{search\_video} retrieves relevant segments via semantic similarity, \texttt{refine\_video} re-analyses a short clip with the appropriate modality specialist (the VLM for visual content, Audio Flamingo~3~\citep{goel2025audio} for speech and non-speech audio), and \texttt{verify\_claim} grounds a tentative claim against retrieved evidence. The orchestrator iterates until the evidence is sufficient or a configurable limit is reached, then synthesizes a final answer. As the experiments in Section~\ref{sec:benchmark_results} show, this simple loop over all three modalities proves more important than model scale alone for hour-length omni-modal questions. Due to space limitations, full architectural details of LongShOTAgent are provided in Appendix~\ref{sec:agent_details}.

\vspace{-1ex}
\section{Experiments}
\label{sec:benchmarking}

\begin{table}[!h]
\centering
\caption{\textbf{LongShOTBench leaderboard (\%).} Curated set of representative models (top entries per family, plus the strongest closed-source API), grouped by paradigm. Within each group, rows are sorted by overall 3-verifier mean. CP/RE/Info/MM and Overall are 3-verifier averages (Qwen3-14B, Gemma-4-31B-it, GPT-5-mini). Cells are shaded by score (deeper = higher); the peach band is our agent. \textbf{Bold}: column best; \underline{underline}: second best (in score columns). \textbf{CP}: Core Perception, \textbf{RE}: Reasoning, \textbf{Info}: Information, \textbf{MM}: Multimodal. Input codes: \textbf{V}=video, \textbf{F}=frames, \textbf{V+A}=video+audio (omni), \textbf{A}=audio-only. Variant suffixes: -T=Thinking, -I=Instruct, -R=Reasoning. Full 105-model leaderboard with parameter counts and per-verifier breakdowns in Table~\ref{tab:full_benchmark_table}; per-modality and per-duration breakdowns in Table~\ref{tab:full_benchmark_modality}.}
\label{tab:main_benchmark_table}

\setlength{\tabcolsep}{3pt}
\renewcommand{\arraystretch}{0.95}
\fontsize{7}{8}\selectfont
\newcolumntype{Y}{>{\centering\arraybackslash}X}

\begin{tabularx}{\textwidth}{@{}lYYYYYY@{}}
\toprule
& & \multicolumn{4}{c}{\textbf{Category scores}} & \\
\cmidrule(lr){3-6}
\textbf{Model} & \textbf{Input} & \textbf{CP} & \textbf{RE} & \textbf{Info} & \textbf{MM} & \textbf{Overall} \\
\midrule
\multicolumn{7}{c}{\textit{\textbf{Video-native VLMs}}} \\
\midrule
\rowcolor{zebrarow}
Qwen3-VL 235B-T & V &  39.84 &  53.12 &  45.36 &  50.60 &  47.23 \\
Qwen3-VL 32B-T & V &  38.57 &  50.79 &  44.32 &  50.97 &  46.16 \\
\rowcolor{zebrarow}
Intern-S1-mini & V &  34.00 &  45.98 &  39.41 &  45.67 &  41.26 \\
Step3-VL 10B & V &  31.35 &  44.40 &  34.78 &  44.09 &  38.65 \\
\rowcolor{zebrarow}
GLM-4.5V & V &  31.10 &  42.50 &  34.53 &  42.53 &  37.66 \\
MiMo-VL 7B & V &  29.92 &  42.54 &  34.77 &  43.03 &  37.56 \\
\rowcolor{zebrarow}
Ovis2.6 30B & V &  26.61 &  36.60 &  31.11 &  39.88 &  33.54 \\
Video-R2 & V &  23.01 &  29.28 &  25.22 &  33.79 &  27.83 \\
\rowcolor{zebrarow}
ERNIE-4.5-VL 28B & V &  23.10 &  30.64 &  24.06 &  30.72 &  27.13 \\
Keye-VL 8B & V &  17.19 &  24.83 &  18.97 &  28.34 &  22.33 \\
\rowcolor{zebrarow}
MiniCPM-V 4.5 & V &  17.59 &  23.63 &  18.61 &  26.80 &  21.66 \\
VideoLLaMA3 7B & V & 12.47 & 16.72 & 13.62 & 23.05 & 16.47 \\
\midrule
\multicolumn{7}{c}{\textit{\textbf{Frame-based VLMs}}} \\
\midrule
\rowcolor{zebrarow}
Gemma-3 27B & F &  34.67 &  47.20 &  39.14 &  45.22 &  41.55 \\
Gemma-3 12B & F &  32.08 &  46.06 &  36.69 &  43.46 &  39.58 \\
\rowcolor{zebrarow}
Kimi-VL-T & F &  17.76 &  27.49 &  21.24 &  30.77 &  24.32 \\
Kimi-VL-I & F & 12.38 & 16.58 & 12.29 & 21.79 & 15.76 \\
\midrule
\multicolumn{7}{c}{\textit{\textbf{Audio-Only LLMs}}} \\
\midrule
\rowcolor{zebrarow}
Voxtral-Mini 3B & A &  33.92 &  53.31 &  42.90 &  40.41 &  49.27 \\
Voxtral-Small 24B & A &  32.63 &  51.65 &  41.38 &  43.53 &  48.41 \\
\rowcolor{zebrarow}
Fun-Audio-Chat 8B & A &  19.92 &  28.85 &  22.93 &  30.69 &  30.94 \\
Audio Flamingo 3 & A &  13.50 &  18.18 &  14.54 &  22.67 &  22.90 \\
\rowcolor{zebrarow}
Qwen2-Audio 7B-I & A &  6.04 &  7.61 &  6.16 &  12.47 &  13.18 \\
\midrule
\multicolumn{7}{c}{\textit{\textbf{Omni-modal Models (native video + audio)}}} \\
\midrule
\rowcolor{zebrarow}
Qwen3-Omni 30B-T & V+A &  \underline{51.44} &  \underline{72.86} &  \underline{65.40} &  \textbf{66.50} &  \underline{64.05} \\
Nemotron-3 Omni 30B-R & V+A &  30.90 &  50.39 &  42.00 &  40.55 &  40.96 \\
\rowcolor{zebrarow}
Qwen3-Omni 30B-I & V+A &  31.21 & 46.92 &  36.96 &  41.59 &  39.17 \\
MiniCPM-o 4.5 & V+A &  20.45 &  25.90 &  23.17 & 21.74 &  22.82 \\
\rowcolor{zebrarow}
HumanSense Omni & V+A &  18.17 &  23.30 &  19.82 &  27.61 &  22.23 \\
Baichuan-Omni 1.5 & V+A & 15.98 & 19.46 & 15.75 &  26.73 & 19.47 \\
\rowcolor{zebrarow}
SALMONN2+ 7B & V+A & 14.00 & 19.67 & 15.91 & 22.37 & 17.99 \\
OmniVinci 9B & V+A & 13.34 & 16.75 & 14.59 & 20.33 & 16.25 \\
\midrule
\multicolumn{7}{c}{\textit{\textbf{Agentic Systems}}} \\
\midrule
\rowcolor{ourscell}
\textbf{LongShOTAgent (Ours)} & V+A & \textbf{55.28} & \textbf{75.69} & \textbf{69.57} & \underline{66.00} & \textbf{66.64} \\
\rowcolor{zebrarow}
VideoMind & V+A & 8.22 & 9.90 & 8.74 & 16.43 & 10.83 \\
Video-RAG & V+A & 5.41 & 5.99 & 7.35 & 7.88 & 6.66 \\
\rowcolor{zebrarow}
Vgent & V+A & 3.44 & 5.43 & 4.29 & 10.15 & 5.83 \\
\midrule
\multicolumn{7}{c}{\textit{\textbf{Proprietary (Closed-source API)}}} \\
\midrule
\rowcolor{zebrarow}
Gemini 3.1 Pro Preview & V+A &  43.01 &  65.82 &  55.57 &  58.15 &  55.63 \\
\bottomrule
\end{tabularx}
\vspace{-2ex}
\end{table}

\vspace{-1ex}
\subsection{Evaluation Pipeline}
\vspace{-1ex}
\textbf{Answer Generation.} Each candidate MLLM is provided with the video and corresponding question; visual, audio, and speech inputs are handled according to each model's supported modalities and inference protocol.\\
\textbf{Answer Evaluation.} Rather than asking a verifier to assign holistic scores, which introduces well-documented variance~\citep{gu2024survey}, we operate in a strictly \emph{grounded} setting: every (question, gold answer, criterion, candidate answer) tuple is reduced to an atomic binary decision, checking whether the candidate matches the human-validated reference with respect to that criterion. \textit{Three} independent verifier models (Qwen3-14B~\citep{yang2025qwen3}, Gemma-4-31B-it, GPT-5-mini) are queried per criterion, and the reported overall in Table~\ref{tab:main_benchmark_table} is the mean of the three verifiers' scores. Agreement with human annotators is detailed in Sec.~\ref{sec:judge_human_alignment_main}.\\
\textbf{Scoring Aggregation.} All scores are on a 0-100 scale using the weighted rubric formula in Eq.~\ref{eq:rubric_score}. Per-category scores are the unweighted mean of their constituent task scores, and the overall score is the unweighted mean of the four category scores. For multi-turn dialogues, we adopt an \textit{ideal trajectory} setting: each turn's response is evaluated independently while the context for subsequent turns is updated with the \textit{ideal response}, isolating intrinsic quality from conversational drift. Per-model single- vs multi-turn comparison is shown in Fig.~\ref{fig:single_multi}.\\
\textbf{Compute.} All generation and evaluation runs used 8$\times$ NVIDIA RTX PRO 6000 Blackwell GPUs with vLLM~\citep{kwon2023efficient} acceleration; configuration scripts are released for reproducibility.

\vspace{-1ex}
\subsection{Benchmark Results}
\label{sec:benchmark_results}
\vspace{-1ex}
 
We evaluate zero-shot performance of 105 video-capable models on LongShOTBench, spanning closed-source APIs, open-source omni-modal and vision-language models, frame-based VLMs, audio LLMs, and agentic pipelines (full leaderboard in Table~\ref{tab:full_benchmark_table}). Table~\ref{tab:main_benchmark_table} reports a curated subset of representative entries per paradigm, ranked by the 3-verifier overall mean. For models that support native video input we pass the raw video directly; for image-only models (e.g., Gemma-3, Kimi-VL) we follow each model's recommended frame-sampling defaults rather than enforcing, so as to better approximate each model's attainable performance under its intended usage.
Below, we summarize main findings revealed by LongShOTBench about the current model landscape.

\textbf{Our LongShOTAgent leads the benchmark, yet LongShOTBench remains far from saturated.}
As shown in Table~\ref{tab:main_benchmark_table}, the strongest closed-source API, Gemini 3.1 Pro Preview, reaches 55.63\%, while the best open-source end-to-end model, Qwen3-Omni 30B-T, reaches 64.05\%. Our training-free LongShOTAgent achieves the highest overall score, 66.64\%, outperforming all 105 evaluated video-capable models. Yet even this result remains far from saturation. Together with the category-level gaps in Table~\ref{tab:main_benchmark_table}, these ceilings indicate that our benchmark is challenging and the hour-length omni-modal video
understanding remains a largely unsolved problem.

\begin{figure}[t]
\centering
\begin{tcolorbox}[colback=white, colframe=black, boxrule=0.5pt, arc=2pt, boxsep=3pt]
\begin{subfigure}[t]{0.50\textwidth}
  \centering
  \includegraphics[width=\textwidth,trim=0 0 0 0,clip]{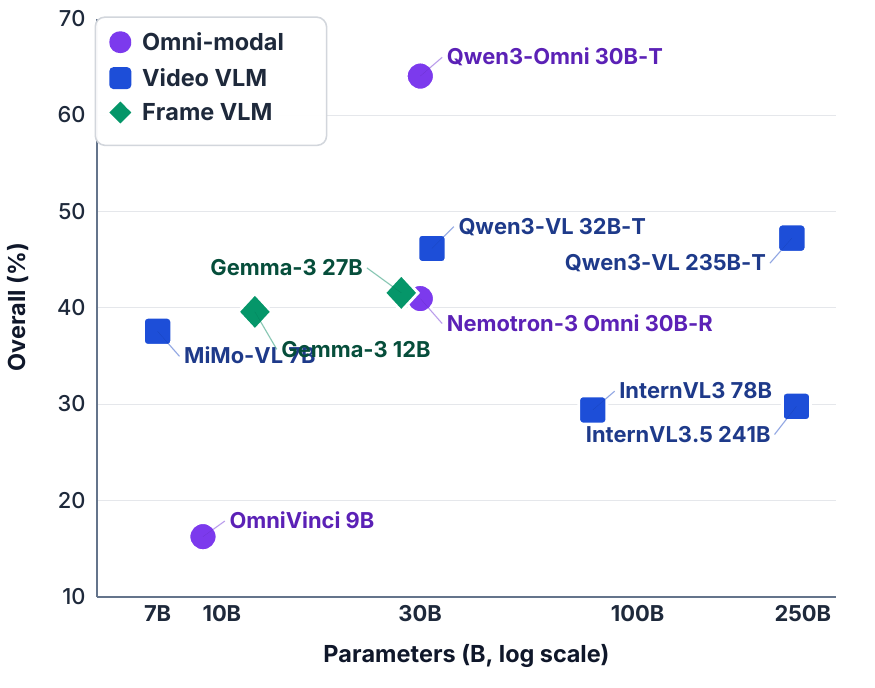}
  \vspace{-3ex}
  \caption{Architecture $>$ scale}
  \label{fig:finding_a}
\end{subfigure}\hfill
\begin{subfigure}[t]{0.50\textwidth}
  \centering
  \includegraphics[width=\textwidth,trim=0 0 0 0,clip]{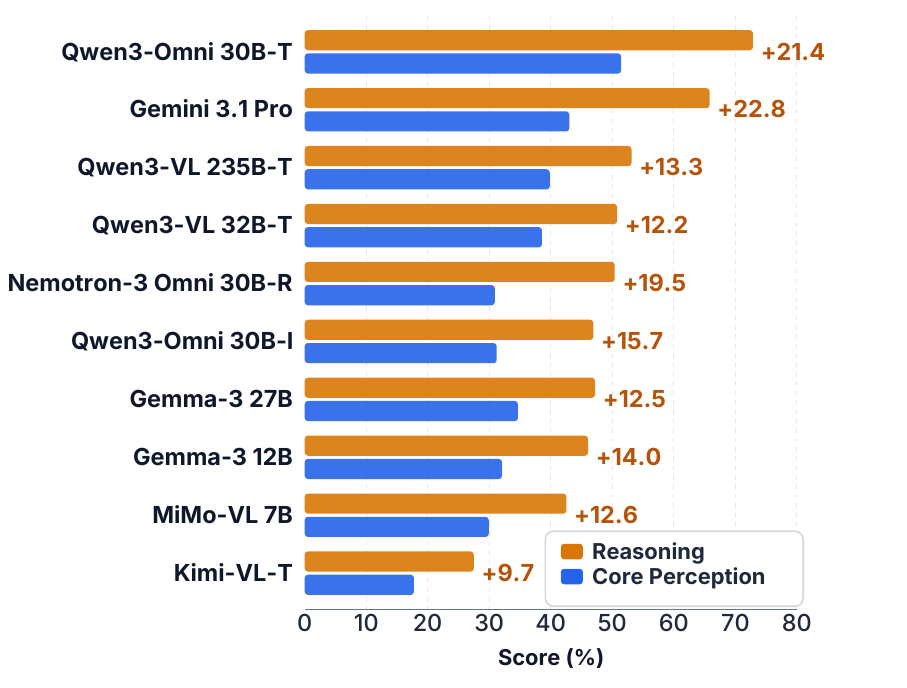}
  \vspace{-3ex}
  \caption{Reasoning $>$ Core Perception}
  \label{fig:finding_b}
\vspace{-3ex}
\end{subfigure}\\[6pt]
\begin{subfigure}[t]{0.50\textwidth}
  \centering
  \includegraphics[width=\textwidth,trim=0 0 0 0,clip]{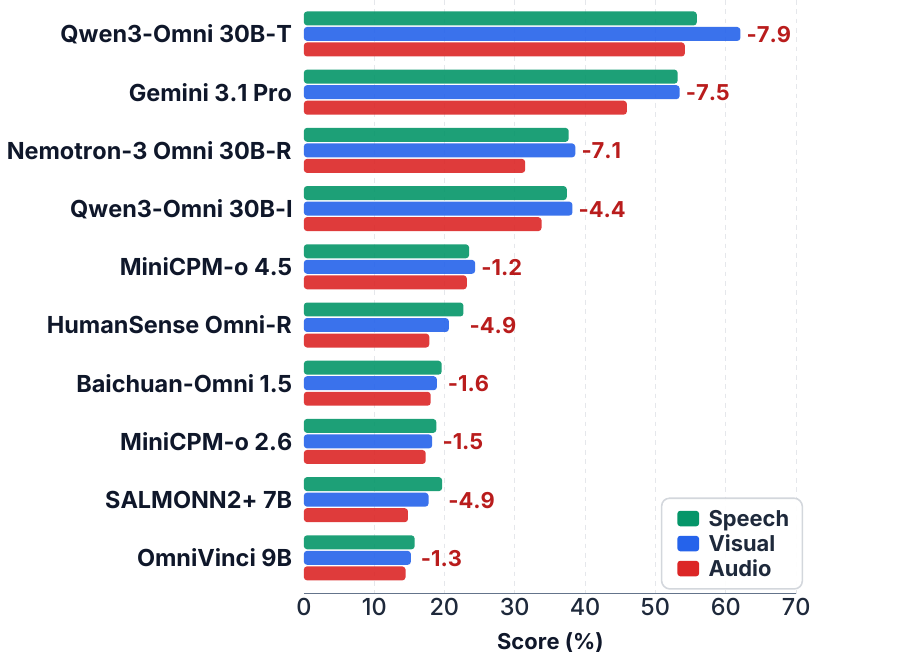}
  \vspace{-3ex}
  \caption{Audio bottleneck}
  \label{fig:finding_c}
\end{subfigure}\hfill
\begin{subfigure}[t]{0.50\textwidth}
  \centering
  \includegraphics[width=\textwidth,trim=0 0 0 0,clip]{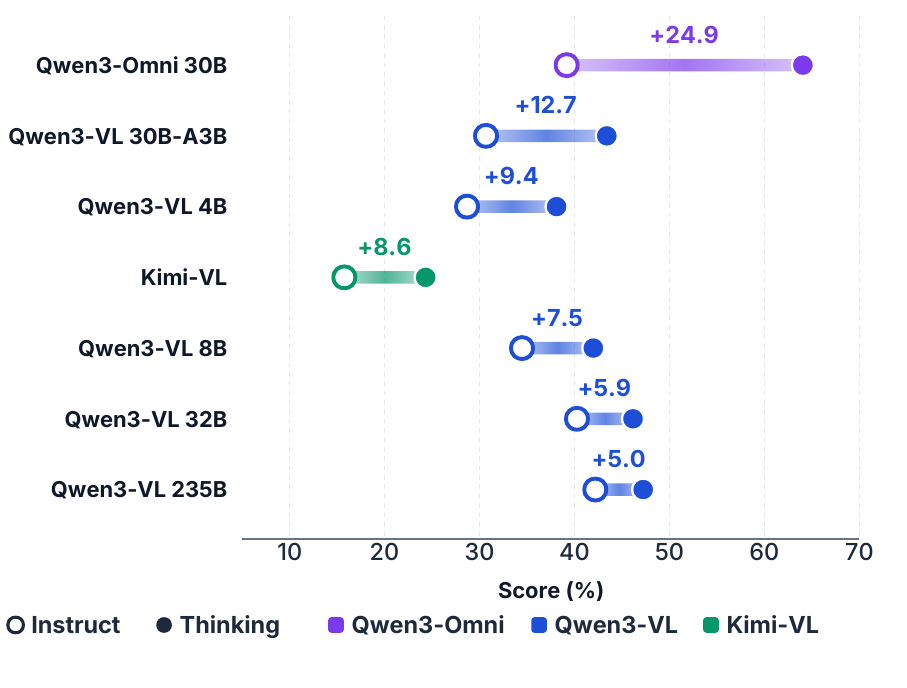}
  \vspace{-3ex}
  \caption{Test-time compute multiplier}
  \label{fig:finding_d}
\end{subfigure}
\vspace{-2ex}
\end{tcolorbox}
\caption{\textbf{Diagnostic findings on LongShOTBench.}
\textbf{(a)} Overall score vs.\ parameter count for open-weight models, colored by paradigm; proprietary Gemini~3.1~Pro~Preview is omitted as its size is undisclosed.
\textbf{(b)} Per-system Core Perception (CP) vs.\ Reasoning (RE), with gap labels reporting RE$-$CP.
\textbf{(c)} Per-modality scores (Visual / Speech / Audio) for ten omni-modal systems; Speech sits behind Visual at near-identical scores (green halo). Gap labels report Audio$\,-\,\max($Visual, Speech$)$.
\textbf{(d)} Instruct vs.\ Thinking siblings within seven model families. All values are 3-verifier mean overall scores from Table~\ref{tab:main_benchmark_table} / Table~\ref{tab:full_benchmark_table}.}
\label{fig:benchmark_findings}
\vspace{-3ex}
\end{figure}

 \vspace{-1ex}
\textbf{Model architecture matters more than scale.}
A clear hierarchy emerges across paradigms: omni-modal models with explicit audio pathways consistently outperform video-native VLMs of comparable or larger size, which in turn lead frame-based systems. The 30B Qwen3-Omni-Thinking (64.05\%) decisively outperforms the 235B Qwen3-VL-Thinking (47.23\%), with an order-of-magnitude smaller model winning by roughly 17 points (Fig.~\ref{fig:benchmark_findings}(a)). Frame-based VLMs at default sampling (e.g., Gemma-3 27B at 41.55\%) also remain competitive with much larger video-native systems (e.g., InternVL3.5 241B at 29.77\%, Table~\ref{tab:full_benchmark_table}). These comparisons suggest that the current bottleneck in long-video understanding lies in temporal indexing and cross-modal alignment rather than in visual or linguistic capacity alone. A dense-vs-MoE breakdown and best-model-per-parameter-tier table are provided in Appendix~\ref{sec:detailed_benchmark_analysis} (Fig.~\ref{fig:param_efficiency}, Table~\ref{tab:cost_eff}).

 \vspace{-1ex}
\textbf{Models reason better than they perceive.}
Counter to the usual expectation that reasoning is harder than perception, top models consistently score higher on Reasoning (RE) than on Core Perception (CP). As Fig.~\ref{fig:benchmark_findings}(b) shows, this gap persists across all top systems: Qwen3-Omni 30B-T scores 72.86\% on RE versus 51.44\% on CP, and Gemini 3.1 Pro Preview shows a comparable 22.8-point spread. One plausible reading is that high-level narrative reasoning can lean on strong language priors and partial evidence, whereas Core Perception demands precise grounding of specific visual or auditory facts at specific moments. This is exactly the regime where long-context dilution and weak temporal indexing hurt most. The inversion suggests that current MLLMs recognize stories more reliably than they perceive the videos those stories are built on, and highlights Core Perception as the priority axis for future model improvement.

 \vspace{-1ex}
\textbf{Non-speech audio is the weakest modality.}
Even within the best omni-modal systems, ambient non-speech audio consistently trails visual and speech performance (Fig.~\ref{fig:benchmark_findings}(c)). Modality breakdowns in Appendix~\ref{sec:detailed_benchmark_analysis} show that Qwen3-Omni 30B-T scores 62.09\% on visual content but only 55.89\% on audio, a gap of over 6 points. The pattern holds even when a dedicated audio specialist is used (LongShOTAgent: visual 64.96\% vs.\ audio 57.61\%). The community has heavily optimized SigLIP-style vision encoders and Whisper-style ASR, but grounding non-speech events (door clicks, music shifts, ambient transitions) to specific timestamps and to the visual stream remains an open problem that LongShOTBench exposes directly.

\vspace{-1ex}
\textbf{Explicit thinking substantially boosts performance.}
Within the same model family, variants that use explicit reasoning or thinking traces (suffix \texttt{-T} or \texttt{-R}) substantially outperform their instruction-tuned counterparts. As shown in Fig.~\ref{fig:benchmark_findings}(d), Qwen3-Omni 30B-T at 64.05\% is more than 24 points above its \texttt{-I} sibling at 39.17\%, and Kimi-VL-T at 24.32\% is roughly 1.5$\times$ the 15.76\% of Kimi-VL-I. The magnitude of these gaps indicates that step-by-step deliberation is doing real work on hour-length inputs, most likely the temporal alignment and evidence gathering that long contexts would otherwise force the model to compress into a single forward pass.
This finding suggests that explicit thinking can serve as an important mechanism for organizing
evidence and sustaining multi-step reasoning under demanding multimodal inputs. Generational progress within the same family at matched scale is broken out in Fig.~\ref{fig:generational}.

 \vspace{-1ex}
\textbf{Existing vision-centric agents do not generalize.}
Three prior agentic systems designed for long-video reasoning (VideoMind~\citep{liu2025videomind}, Video-RAG~\citep{luo2024video}, Vgent~\citep{shen2025vgent}) all score below 15\% on LongShOTBench, far below even modest end-to-end models. These systems operate predominantly over the visual stream and, where available, ASR transcripts, without first-class treatment of non-speech audio. Their failure on LongShOTBench confirms that omni-modal coverage is not optional for hour-scale video understanding. By contrast, our baseline LongShOTAgent, which adds dedicated audio processing alongside visual and speech retrieval in a {search-refine-verify} loop, reaches 66.64\%, significantly outperforming these agentic video systems. This gap between vision-only and omni-modal agents further validates that LongShOTBench is effectively testing multi-sensory integration rather than visual comprehension alone.

\vspace{-1ex}
\begin{wraptable}{r}{0.48\textwidth}
\vspace{-3ex}
\centering
\caption{\textbf{Orchestrator ablation.} Baseline is the standalone score; $\Delta$ is the gain from the agentic loop. Qwen3-14B single-verifier.}
\label{tab:orchestrator_ablation}
\footnotesize
\setlength{\tabcolsep}{8pt}
\renewcommand{\arraystretch}{1.1}
\begin{tabular}{lccc}
\toprule
\textbf{Orchestrator} & \textbf{Base} & \textbf{Agent} & \textbf{$\Delta$} \\
\midrule
\rowcolor{zebrarow}
Qwen3.6-35B-A3B  & 27.17 & \textbf{65.69} & \textcolor{darkgreen}{+38.52} \\
Qwen3-VL-30B-A3B    & 25.82 & 52.54 & \textcolor{darkgreen}{+26.72} \\
\rowcolor{zebrarow}
Gemma-4-31B-IT   & 34.22 & 47.17 & \textcolor{darkgreen}{+12.95} \\
\bottomrule
\end{tabular}
\vspace{-2ex}
\end{wraptable}
\textbf{LongShOTAgent improves diverse orchestrators.}
Table~\ref{tab:orchestrator_ablation} compares three orchestrator LLMs in their standalone and LongShOTAgent configurations (Qwen3-14B single-verifier). All three benefit from the agentic loop, with deltas ranging from +12.95 to +38.52\,pts. Notably, even the lowest-scoring configuration (Gemma-4 at 47.17\%) exceeds most standalone VLMs in Table~\ref{tab:main_benchmark_table}, indicating that structured retrieval and modality-aware re-analysis add value beyond what a single forward pass provides. The wide spread in deltas also suggests that orchestration quality, not just tool availability, is a meaningful factor.

\begin{figure}[t]
\centering
\includegraphics[width=\textwidth]{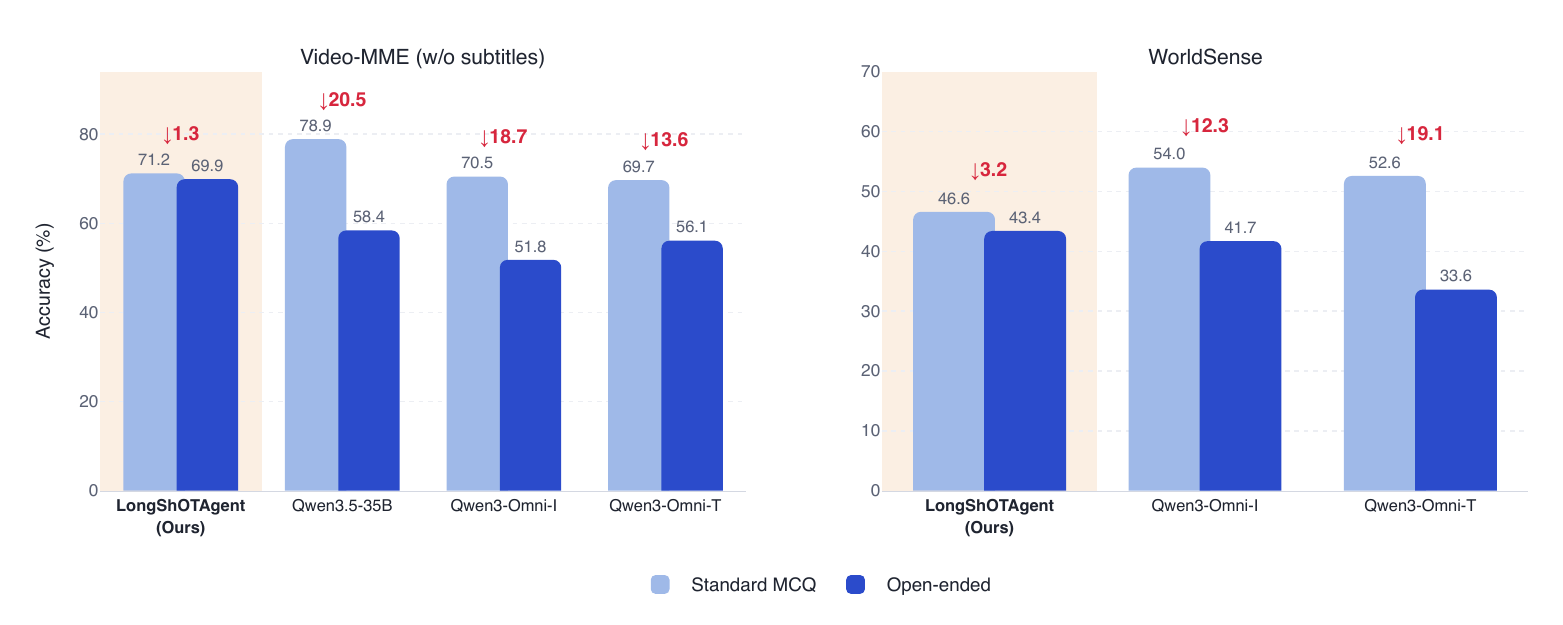}
\vspace{-1.5ex}
\caption{\textbf{Robust where others game the test.} Accuracy under the standard 4-option MCQ protocol (faint bars) versus options-stripped open-ended evaluation (solid bars) on Video-MME and WorldSense; $\Delta$ is the accuracy drop from the standard MCQ protocol to open-ended evaluation. Monolithic models lose 13 to 20 points once option-level shortcuts are removed, while LongShOTAgent, which generates answers from retrieved evidence rather than selecting from a list, drops only 1.3 and 3.2 points and leads every open-ended column. Qwen3.5-35B-A3B is a vision-language model with no audio encoder; since WorldSense supplies the native audio its questions depend on, it is shown on Video-MME only. The full four-protocol results (shuffled options, appended NOTA/AOTA distractors) are reported in Appendix~\ref{sec:robustness}.}
\label{fig:robustness}
\vspace{-3ex}
\end{figure}

\vspace{-1ex}
\textbf{Robust where others game the test.}
Multiple-choice benchmarks reward option-level shortcuts (positional bias, lexical overlap with the question stem, and process-of-elimination) that inflate accuracy without genuine understanding. We probe this by re-evaluating LongShOTAgent and strong monolithic baselines on Video-MME~\citep{fu2024video} and WorldSense~\citep{hong2025worldsense} with the options removed, forcing open-ended answers (Fig.~\ref{fig:robustness}). The monolithic models collapse: on Video-MME, Qwen3.5-35B falls 20.5\,pts (78.9$\rightarrow$58.4\%), Qwen3-Omni-Instruct 18.7\,pts, and Qwen3-Omni-Thinking 13.6\,pts. On the audio-centric WorldSense the two omni-models even start ahead of the agent under MCQ (54.0\% and 52.65\% vs.\ 46.56\%), yet that lead evaporates open-ended, where they shed 12.3 and 19.1\,pts. LongShOTAgent, which generates answers from retrieved evidence rather than selecting from a list, barely moves ($-1.3$ and $-3.2$\,pts) and leads every open-ended column. The advantage is comprehension, not multiple-choice shortcuts: much of the monolithic MCQ accuracy does not survive realistic evaluation, while the open-ended, rubric-graded protocol used throughout LongShOTBench cannot be gamed this way. Adversarial option perturbations (shuffling, and appended NOTA/AOTA distractors) reinforce this and are reported in full in Appendix~\ref{sec:robustness}.

\vspace{-1ex}
\subsection{Verifier-Human Alignment}
\vspace{-1ex}
\label{sec:judge_human_alignment_main}

\begin{table}[t]
\centering
\caption{\textbf{Human-verifier alignment.} Three LLM verifiers and
the 3-verifier ensemble vs.\ the 4 human annotators. Cohen's $\kappa$ and F1 (True-class) are computed on the $400$ binary criterion decisions; Pearson $r$ is on the $100$ aggregated per-(question, model) continuous weighted scores. Inter-human ref.\ is the leave-one-out mean (range) over the annotators.}
\label{tab:human_judge_alignment}
\setlength{\tabcolsep}{8pt}
\renewcommand{\arraystretch}{1.15}
\resizebox{0.62\textwidth}{!}{%
\begin{tabular}{lccc|c|c}
\toprule
\textbf{Metric} & \multicolumn{3}{c|}{\textbf{Individual verifiers}} & \textbf{3-ver.} & \textbf{Inter-human} \\
                  & GPT-5-mini & Gemma-4 & Qwen3 & \textbf{ens.} & \textbf{ref.} \\
\midrule
Cohen's $\kappa$    & 0.677 & 0.735 & 0.701 & \textbf{0.753} & 0.672 (0.61-0.76) \\
F1 (True)     & 0.805 & 0.838 & 0.812    & \textbf{0.847}    & 0.799 (0.76-0.85) \\
\midrule
Pearson $r$   & 0.867 & 0.911 & 0.874 & \textbf{0.923} & 0.832 (0.78-0.90) \\
\bottomrule
\end{tabular}}
\end{table}

To validate the rubric-guided evaluation, four expert annotators independently re-graded a stratified pool of {400 rubric criteria} spanning all task categories, duration buckets, and five representative systems (protocol and per-model breakdowns in Appendix~\ref{sec:human_alignment}). Table~\ref{tab:human_judge_alignment} reports criterion- and sample-level agreement. The 3-verifier ensemble reaches $\kappa{=}0.75$, F1$\,{=}\,0.85$, and sample-level Pearson $r{=}0.92$ against the human consensus; its $\kappa$ and F1 fall within the leave-one-out inter-annotator reference band ($\kappa$ mean $0.67$, range $[0.61, 0.76]$)~\citep{landis1977measurement}, while the continuous $r$ slightly exceeds the human range ($[0.78, 0.90]$) as expected when scoring against the denoised consensus: the ensemble agrees with humans at least as well as the most consistent human agrees with the others. When criterion-level decisions are aggregated into system-level scores, the 3-verifier mean preserves the human ranking up to two near-tied adjacent pairs (Spearman $\rho{=}0.80$, both flips between systems with verifier-side gap $<\!0.04$), with per-system gaps within $\sim 0.1$. Per-system score breakdowns and the full pairwise rater-agreement heatmap are reported in Appendix~\ref{sec:human_alignment} (Table~\ref{tab:per_model_alignment}, Fig.~\ref{fig:alignment_heatmap}). We therefore report the 3-verifier mean as the primary metric throughout.

\vspace{-1ex}
\section{Conclusion}
\label{sec:conclusion}
\vspace{-1ex}
We present LongShOTBench, a comprehensive diagnostic benchmark for evaluating MLLMs on long-form omni-modal video understanding, integrating vision, speech, and audio across hour-long contexts. Its open-ended, intent-driven questions and rubric-guided evaluation provide fine-grained, interpretable diagnostics across perception, temporal reasoning, and omni-modal understanding.
We also propose LongShOTAgent, a training-free agentic framework that reasons over long videos by iteratively searching, refining, and verifying omni-modal evidence. 
We benchmark 105 video-capable MLLMs on LongShOTBench, revealing substantial gaps in current state-of-the-art systems and underscoring the persistent challenges of coherent omni-modal reasoning over extended temporal contexts.
Our work aims to support reproducible research and provide a practical testbed for future model development. Collectively, these contributions help guide the next generation of MLLMs toward robust real-world video understanding and more reliable omni-modal reasoning. Limitations are discussed in Appendix~\ref{sec:limitations}.

\begin{ack}
This work is partially supported by the Meta Regional Research Grant  (Project OMER), the Google Gift Research Award, and the NVIDIA Academic Grant. We also thank Hafsa Hanan, Xu Liu, Dian Jin, and Cailing Han for their contribution to the annotation and validation process.
\end{ack}

{
\small
\bibliographystyle{plainnat}
\bibliography{main}
}

\appendix

\section{Full Benchmark Results}
\label{sec:full_benchmark_results}

Tables~\ref{tab:full_benchmark_table} and~\ref{tab:full_benchmark_modality} report the full per-model results across all 105 evaluated video-capable models, plus our LongShOTAgent, for 106 leaderboard rows in total, complementing the curated subset in Table~\ref{tab:main_benchmark_table} of the main paper. The leaderboard spans six paradigms: video-native VLMs (71), frame-based VLMs (10), omni-modal models with native video and audio input (15), audio-only LLMs (5), agentic systems (4, including our LongShOTAgent), and one proprietary closed-source API. Results are split into two companion tables for readability: category and per-judge scores (Table~\ref{tab:full_benchmark_table}), and per-modality and per-duration breakdowns (Table~\ref{tab:full_benchmark_modality}). Highlighting and verifier definitions follow the main table.

\begingroup
\captionsetup{font=normalsize,labelfont={normalsize,bf},justification=justified,singlelinecheck=false}
\setlength{\tabcolsep}{4pt}
\renewcommand{\arraystretch}{1.25}
\footnotesize
\begin{longtable}{@{}>{\raggedright}p{3.6cm} c c cccc cccc@{}}
\caption{\textbf{Full LongShOTBench leaderboard: Category \& per-judge scores (\%).} CP/RE/Info/MM are 3-judge averages; per-judge columns report each judge's overall, and \textbf{Avg} is their mean. Cells are shaded by tier (deeper\,=\,higher); the peach band marks our agent. \textbf{Bold}: column best; \underline{underline}: second best. Input codes: \textbf{V}=video, \textbf{F}=frames, \textbf{V+A}=video+audio (omni), \textbf{A}=audio-only. Variant suffixes: \texttt{-T}=Thinking, \texttt{-I}=Instruct, \texttt{-R}=Reasoning.}
\label{tab:full_benchmark_table} \\
\toprule
\textbf{Model} & \textbf{Params} & \textbf{Input} & \textbf{CP} & \textbf{RE} & \textbf{Info} & \textbf{MM} & \textbf{Qwen3} & \textbf{Gemma-4} & \textbf{GPT-5} & \textbf{Avg} \\
\midrule
\endfirsthead
\multicolumn{11}{l}{\itshape Continued from previous page} \\
\toprule
\textbf{Model} & \textbf{Params} & \textbf{Input} & \textbf{CP} & \textbf{RE} & \textbf{Info} & \textbf{MM} & \textbf{Qwen3} & \textbf{Gemma-4} & \textbf{GPT-5} & \textbf{Avg} \\
\midrule
\endhead
\multicolumn{11}{r}{\itshape Continued on next page} \\
\endfoot
\bottomrule
\endlastfoot
\midrule
\multicolumn{11}{c}{\textit{\textbf{Video-native VLMs}}} \\
\midrule
\rowcolor{zebrarow}
\makecell[l]{Qwen3-VL-235B-A22B-T\\[-1pt]{\scriptsize \citep{bai2025qwen3}}} & \makecell{235B\\(22B active)} & V & \cT 39.84 & \cT 53.12 & \cT 45.36 & \cT 50.60 & 47.05 & 42.39 & 52.25 & \cT 47.23 \\
\makecell[l]{Qwen3-VL-32B-T\\[-1pt]{\scriptsize \citep{bai2025qwen3}}} & 32B & V & \cT 38.57 & \cT 50.79 & \cT 44.32 & \cT 50.97 & 45.66 & 41.76 & 51.06 & \cT 46.16 \\
\rowcolor{zebrarow}
\makecell[l]{Vero-Qwen3I-8B\\[-1pt]{\scriptsize \citep{sarch2026vero}}} & 8B & V & \cT 40.54 & \cT 49.34 & \cT 42.29 & \cT 50.76 & 41.65 & 42.50 & 53.05 & \cT 45.73 \\
\makecell[l]{Kimi-K2.6\\[-1pt]{\scriptsize \citep{team2026kimi}}} & 1.1T & V & \cT 37.52 & \cT 51.61 & \cT 42.75 & \cT 48.91 & 41.57 & 42.40 & 51.62 & \cT 45.20 \\
\rowcolor{zebrarow}
\makecell[l]{Qwen3-VL-30B-A3B-T\\[-1pt]{\scriptsize \citep{bai2025qwen3}}} & \makecell{30B\\(3B active)} & V & \cT 35.89 & \cT 49.01 & \cT 41.44 & \cT 47.26 & 42.40 & 38.94 & 48.86 & \cT 43.40 \\
\makecell[l]{Qwen3-VL-235B-A22B-I\\[-1pt]{\scriptsize \citep{bai2025qwen3}}} & \makecell{235B\\(22B active)} & V & \cT 35.67 & \cT 45.61 & \cT 40.11 & \cT 47.37 & 39.01 & 38.99 & 48.57 & \cT 42.19 \\
\rowcolor{zebrarow}
\makecell[l]{Qwen3-VL-8B-T\\[-1pt]{\scriptsize \citep{bai2025qwen3}}} & 8B & V & \cT 34.61 & \cT 47.22 & \cT 38.98 & \cT 47.10 & 41.68 & 37.40 & 46.85 & \cT 41.98 \\
\makecell[l]{Intern-S1-mini\\[-1pt]{\scriptsize \citep{bai2025intern}}} & 14B & V & \cT 34.00 & \cT 45.98 & \cT 39.41 & \cT 45.67 & 39.85 & 36.86 & 47.08 & \cT 41.26 \\
\rowcolor{zebrarow}
\makecell[l]{Qwen3-VL-32B-I\\[-1pt]{\scriptsize \citep{bai2025qwen3}}} & 32B & V & \cT 34.55 & \cT 43.72 & \cT 36.92 & \cT 45.84 & 35.99 & 37.43 & 47.35 & \cT 40.26 \\
\makecell[l]{Step3-VL-10B\\[-1pt]{\scriptsize \citep{huang2026step3vl10btechnicalreport}}} & 10B & V & \cT 31.35 & \cT 44.40 & \cT 34.78 & \cT 44.09 & 37.70 & 34.10 & 44.16 & \cT 38.65 \\
\rowcolor{zebrarow}
\makecell[l]{Qwen3-VL-4B-T\\[-1pt]{\scriptsize \citep{bai2025qwen3}}} & 4B & V & \cT 30.77 & \cT 42.94 & \cT 35.44 & \cT 43.28 & 36.26 & 34.67 & 43.39 & \cT 38.11 \\
\makecell[l]{GLM-4.5V\\[-1pt]{\scriptsize \citep{hong2025glm}}} & \makecell{106B\\(12B active)} & V & \cT 31.10 & \cT 42.50 & \cT 34.53 & \cT 42.53 & 34.73 & 33.56 & 44.70 & \cT 37.66 \\
\rowcolor{zebrarow}
\makecell[l]{MiMo-VL-7B-RL-2508\\[-1pt]{\scriptsize \citep{coreteam2025mimovltechnicalreport}}} & 7B & V & \cT 29.92 & \cT 42.54 & \cT 34.77 & \cT 43.03 & 35.12 & 33.23 & 44.34 & \cT 37.56 \\
\makecell[l]{Qwen2.5-VL-32B-I\\[-1pt]{\scriptsize \citep{bai2025qwen2}}} & 32B & V & \cT 30.87 & \cT 39.24 & \cT 33.19 & \cT 40.31 & 31.36 & 32.39 & 43.95 & \cT 35.90 \\
\rowcolor{zebrarow}
\makecell[l]{Qwen3-VL-8B-I\\[-1pt]{\scriptsize \citep{bai2025qwen3}}} & 8B & V & \cT 28.79 & \cH 37.50 & \cH 31.00 & \cT 40.57 & 30.01 & 32.06 & 41.33 & \cT 34.47 \\
\makecell[l]{Ovis2.6-30B-A3B\\[-1pt]{\scriptsize \citep{lu2025ovis25technicalreport}}} & \makecell{30B\\(3B active)} & V & \cH 26.61 & \cH 36.60 & \cT 31.11 & \cT 39.88 & 29.99 & 30.23 & 40.41 & \cH 33.54 \\
\rowcolor{zebrarow}
\makecell[l]{GLM-4.6V-Flash\\[-1pt]{\scriptsize \citep{hong2025glm}}} & 9B & V & \cH 26.72 & \cH 37.11 & \cH 30.64 & \cH 38.93 & 30.46 & 29.75 & 39.84 & \cH 33.35 \\
\makecell[l]{Qwen3.6-27B\\[-1pt]{\scriptsize \citep{qwen36_35b_a3b}}} & 27B & V & \cH 24.59 & \cH 37.61 & \cH 30.36 & \cH 38.92 & 29.08 & 30.52 & 39.01 & \cH 32.87 \\
\rowcolor{zebrarow}
\makecell[l]{Qwen2.5-VL-72B-I\\[-1pt]{\scriptsize \citep{bai2025qwen2}}} & 72B & V & \cH 26.45 & \cH 35.29 & \cH 29.27 & \cH 38.31 & 26.34 & 29.84 & 40.82 & \cH 32.33 \\
\makecell[l]{Qwen3.6-35B-A3B\\[-1pt]{\scriptsize \citep{qwen36_35b_a3b}}} & \makecell{35B\\(3B active)} & V & \cH 22.60 & \cH 36.87 & \cH 27.73 & \cH 39.75 & 27.17 & 30.47 & 37.57 & \cH 31.74 \\
\rowcolor{zebrarow}
\makecell[l]{Qwen3.5-27B\\[-1pt]{\scriptsize \citep{qwen3.5}}} & 27B & V & \cH 22.39 & \cT 37.85 & \cH 28.41 & \cH 36.99 & 26.64 & 29.71 & 37.89 & \cH 31.41 \\
\makecell[l]{Qwen3.5-122B-A10B\\[-1pt]{\scriptsize \citep{qwen3.5}}} & \makecell{122B\\(10B active)} & V & \cH 22.50 & \cH 34.98 & \cH 29.36 & \cH 37.63 & 27.09 & 28.99 & 37.27 & \cH 31.12 \\
\rowcolor{zebrarow}
\makecell[l]{Qwen3.5-35B-A3B\\[-1pt]{\scriptsize \citep{qwen3.5}}} & \makecell{35B\\(3B active)} & V & \cM 22.17 & \cH 36.23 & \cH 27.11 & \cH 38.82 & 26.86 & 29.49 & 36.90 & \cH 31.08 \\
\makecell[l]{R-4B\\[-1pt]{\scriptsize \citep{yang2025r4bincentivizinggeneralpurposeautothinking}}} & 4B & V & \cH 24.32 & \cH 35.77 & \cH 28.04 & \cH 35.95 & 28.73 & 27.94 & 36.39 & \cH 31.02 \\
\rowcolor{zebrarow}
\makecell[l]{Qwen3-VL-30B-A3B-I\\[-1pt]{\scriptsize \citep{bai2025qwen3}}} & \makecell{30B\\(3B active)} & V & \cH 26.80 & \cH 32.82 & \cH 27.78 & \cH 35.32 & 25.82 & 28.79 & 37.42 & \cH 30.68 \\
\makecell[l]{GLM-4.1V-9B-T\\[-1pt]{\scriptsize \citep{hong2025glm}}} & 9B & V & \cH 22.92 & \cH 35.02 & \cH 26.83 & \cH 34.44 & 23.71 & 27.31 & 38.39 & \cH 29.80 \\
\rowcolor{zebrarow}
\makecell[l]{InternVL3.5-241B-A28B\\[-1pt]{\scriptsize \citep{wang2025internvl3_5}}} & \makecell{241B\\(28B active)} & V & \cH 24.63 & \cH 32.74 & \cH 26.63 & \cH 35.07 & 24.33 & 28.02 & 36.95 & \cH 29.77 \\
\makecell[l]{InternVL3-78B\\[-1pt]{\scriptsize \citep{zhu2025internvl3}}} & 78B & V & \cH 23.85 & \cH 31.78 & \cH 26.97 & \cH 35.01 & 23.58 & 27.34 & 37.28 & \cH 29.40 \\
\rowcolor{zebrarow}
\makecell[l]{InternVL3-38B\\[-1pt]{\scriptsize \citep{zhu2025internvl3}}} & 38B & V & \cH 24.93 & \cH 32.98 & \cH 25.81 & \cH 33.72 & 23.48 & 27.71 & 36.89 & \cH 29.36 \\
\makecell[l]{Qwen3-VL-4B-I\\[-1pt]{\scriptsize \citep{bai2025qwen3}}} & 4B & V & \cH 23.84 & \cH 30.85 & \cH 25.38 & \cH 34.65 & 23.34 & 27.09 & 35.62 & \cH 28.68 \\
\rowcolor{zebrarow}
\makecell[l]{VideoRFT-7B\\[-1pt]{\scriptsize \citep{wang2025videorft}}} & 7B & V & \cH 24.37 & \cH 29.57 & \cH 25.92 & \cH 34.38 & 22.13 & 26.18 & 37.38 & \cH 28.56 \\
\makecell[l]{DeepEyes-7B\\[-1pt]{\scriptsize \citep{zheng2025deepeyesincentivizingthinkingimages}}} & 7B & V & \cH 23.40 & \cH 30.82 & \cH 24.18 & \cH 33.90 & 21.22 & 26.05 & 36.95 & \cH 28.07 \\
\rowcolor{zebrarow}
\makecell[l]{Ovis2.5-9B\\[-1pt]{\scriptsize \citep{lu2025ovis25technicalreport}}} & 9B & V & \cH 22.39 & \cH 29.98 & \cH 24.90 & \cH 34.86 & 23.19 & 25.55 & 35.35 & \cH 28.03 \\
\makecell[l]{Video-R2\\[-1pt]{\scriptsize \citep{maaz2025video}}} & 7B & V & \cH 23.01 & \cM 29.28 & \cH 25.22 & \cH 33.79 & 21.49 & 25.01 & 36.98 & \cH 27.83 \\
\rowcolor{zebrarow}
\makecell[l]{Molmo2-8B\\[-1pt]{\scriptsize \citep{clark2026molmo2openweightsdata}}} & 8B & V & \cH 22.62 & \cH 29.48 & \cH 24.11 & \cH 35.01 & 23.10 & 27.08 & 33.24 & \cH 27.81 \\
\makecell[l]{Video-R1-7B\\[-1pt]{\scriptsize \citep{feng2025videor1reinforcingvideoreasoning}}} & 7B & V & \cH 22.44 & \cM 29.04 & \cH 24.67 & \cH 34.73 & 21.40 & 25.44 & 36.33 & \cH 27.72 \\
\rowcolor{zebrarow}
\makecell[l]{LongVT-RL\\[-1pt]{\scriptsize \citep{yang2025longvtincentivizingthinkinglong}}} & 7B & V & \cH 22.61 & \cM 28.80 & \cH 25.28 & \cH 34.05 & 20.91 & 25.06 & 37.08 & \cH 27.68 \\
\makecell[l]{RePlan-Qwen2.5-VL-7B\\[-1pt]{\scriptsize \citep{qu2025replan}}} & 7B & V & \cH 22.78 & \cH 30.21 & \cM 23.40 & \cH 34.16 & 21.09 & 25.21 & 36.60 & \cH 27.63 \\
\rowcolor{zebrarow}
\makecell[l]{InternVL3-14B\\[-1pt]{\scriptsize \citep{zhu2025internvl3}}} & 14B & V & \cH 23.26 & \cH 30.21 & \cH 24.10 & \cM 32.62 & 21.21 & 26.21 & 35.21 & \cH 27.54 \\
\makecell[l]{Qwen3.5-9B\\[-1pt]{\scriptsize \citep{qwen3.5}}} & 9B & V & \cM 18.97 & \cH 33.18 & \cH 24.66 & \cM 33.10 & 23.30 & 26.24 & 32.88 & \cH 27.47 \\
\rowcolor{zebrarow}
\makecell[l]{TimeSearch-R\\[-1pt]{\scriptsize \citep{pan2025timesearch}}} & 7B & V & \cM 22.22 & \cH 30.43 & \cM 23.93 & \cM 32.16 & 19.81 & 25.61 & 36.12 & \cM 27.18 \\
\makecell[l]{ERNIE-4.5-VL-28B-A3B\\[-1pt]{\scriptsize \citep{ernie2025technicalreport}}} & \makecell{28B\\(3B active)} & V & \cH 23.10 & \cH 30.64 & \cM 24.06 & \cM 30.72 & 21.90 & 25.15 & 34.33 & \cM 27.13 \\
\rowcolor{zebrarow}
\makecell[l]{InternVL3.5-8B\\[-1pt]{\scriptsize \citep{wang2025internvl3_5}}} & 8B & V & \cM 20.94 & \cM 27.99 & \cM 23.30 & \cH 34.09 & 21.10 & 25.00 & 33.64 & \cM 26.58 \\
\makecell[l]{VideoChat-R1-7B\\[-1pt]{\scriptsize \citep{li2025videochat}}} & 7B & V & \cM 21.59 & \cM 27.56 & \cM 23.52 & \cM 32.94 & 19.21 & 24.35 & 35.65 & \cM 26.40 \\
\rowcolor{zebrarow}
\makecell[l]{InternVL3.5-38B\\[-1pt]{\scriptsize \citep{wang2025internvl3_5}}} & 38B & V & \cM 21.46 & \cM 28.50 & \cM 22.63 & \cM 32.34 & 20.73 & 24.57 & 33.39 & \cM 26.23 \\
\makecell[l]{PixelReasoner-RL-v1\\[-1pt]{\scriptsize \citep{wang2025pixel}}} & 7B & V & \cM 21.22 & \cH 29.56 & \cM 22.84 & \cM 30.94 & 19.23 & 24.54 & 34.65 & \cM 26.14 \\
\rowcolor{zebrarow}
\makecell[l]{Qwen2.5-VL-7B-I\\[-1pt]{\scriptsize \citep{bai2025qwen2}}} & 7B & V & \cM 21.73 & \cM 27.29 & \cM 23.67 & \cM 31.16 & 18.39 & 23.56 & 35.94 & \cM 25.96 \\
\makecell[l]{InternVL2.5-78B\\[-1pt]{\scriptsize \citep{chen2024expanding}}} & 78B & V & \cM 21.22 & \cM 27.53 & \cM 22.08 & \cM 32.04 & 20.38 & 23.98 & 32.78 & \cM 25.71 \\
\rowcolor{zebrarow}
\makecell[l]{InternVL3-8B\\[-1pt]{\scriptsize \citep{zhu2025internvl3}}} & 8B & V & \cM 20.52 & \cM 25.70 & \cM 21.66 & \cM 30.83 & 18.94 & 23.69 & 31.40 & \cM 24.68 \\
\makecell[l]{InternVL2.5-38B\\[-1pt]{\scriptsize \citep{chen2024expanding}}} & 38B & V & \cM 20.70 & \cM 26.46 & \cM 19.98 & \cM 30.79 & 18.86 & 23.03 & 31.57 & \cM 24.49 \\
\rowcolor{zebrarow}
\makecell[l]{R1-Onevision-7B\\[-1pt]{\scriptsize \citep{yang2025r1}}} & 7B & V & \cM 18.78 & \cM 26.26 & \cM 20.58 & \cM 29.93 & 18.06 & 22.77 & 30.83 & \cM 23.89 \\
\makecell[l]{Phi-4-R-Vision-15B\\[-1pt]{\scriptsize \citep{aneja2026phi4reasoningvision15btechnicalreport}}} & 15B & V & \cM 19.33 & \cM 25.90 & \cM 19.21 & \cM 30.01 & 19.51 & 22.25 & 29.07 & \cM 23.61 \\
\rowcolor{zebrarow}
\makecell[l]{Qwen2.5-VL-7B-VRAG\\[-1pt]{\scriptsize \citep{wang2025vrag}}} & 7B & V & \cM 20.45 & \cM 23.28 & \cM 20.74 & \cM 29.38 & 17.32 & 22.03 & 31.04 & \cM 23.46 \\
\makecell[l]{AURA-32B\\[-1pt]{\scriptsize \citep{aura2026}}} & 32B & V & \cM 19.57 & \cM 24.25 & \cM 20.24 & \cM 28.70 & 17.40 & 22.91 & 29.26 & \cM 23.19 \\
\rowcolor{zebrarow}
\makecell[l]{InternVL3.5-4B-I\\[-1pt]{\scriptsize \citep{wang2025internvl3_5}}} & 4B & V & \cM 19.06 & \cM 24.15 & \cM 19.73 & \cM 29.02 & 17.11 & 22.03 & 29.82 & \cM 22.99 \\
\makecell[l]{Qwen3.5-4B\\[-1pt]{\scriptsize \citep{qwen3.5}}} & 4B & V & 16.26 & \cM 26.56 & \cM 20.31 & 26.48 & 17.78 & 21.13 & 28.29 & \cM 22.40 \\
\rowcolor{zebrarow}
\makecell[l]{Keye-VL-1.5-8B\\[-1pt]{\scriptsize \citep{yang2025kwaikeyevl15technical}}} & 8B & V & 17.19 & \cM 24.83 & \cM 18.97 & \cM 28.34 & 17.09 & 21.23 & 28.68 & \cM 22.33 \\
\makecell[l]{NVIDIA Nemotron Nano 12B VL\\[-1pt]{\scriptsize \citep{nvidia2025nvidianemotronnanov2}}} & 12B & V & \cM 17.38 & 22.95 & 18.78 & \cM 29.05 & 16.71 & 21.55 & 27.87 & \cM 22.04 \\
\rowcolor{zebrarow}
\makecell[l]{MiniCPM-V 4.5\\[-1pt]{\scriptsize \citep{yao2024minicpm}}} & 8B & V & \cM 17.59 & \cM 23.63 & 18.61 & \cM 26.80 & 16.29 & 20.44 & 28.24 & 21.66 \\
SpaceThinker-3B & 3B & V & 16.94 & 21.69 & 18.71 & 23.52 & 13.52 & 19.27 & 27.85 & 20.21 \\
\rowcolor{zebrarow}
\makecell[l]{Qwen3-VL-2B-I\\[-1pt]{\scriptsize \citep{bai2025qwen3}}} & 2B & V & 17.35 & 20.07 & 17.30 & 25.22 & 13.84 & 19.46 & 26.64 & 19.98 \\
\makecell[l]{Qwen2.5-VL-3B-I\\[-1pt]{\scriptsize \citep{bai2025qwen2}}} & 3B & V & 16.26 & 21.10 & 16.61 & 25.15 & 12.88 & 19.65 & 26.80 & 19.78 \\
\rowcolor{zebrarow}
\makecell[l]{Qwen3.5-2B\\[-1pt]{\scriptsize \citep{qwen3.5}}} & 2B & V & 17.25 & 19.12 & 18.52 & 24.07 & 14.19 & 19.45 & 25.59 & 19.74 \\
\makecell[l]{Qwen3.5-0.8B\\[-1pt]{\scriptsize \citep{qwen3.5}}} & 0.8B & V & 15.10 & 16.60 & 15.67 & 22.39 & 11.89 & 17.68 & 22.75 & 17.44 \\
\rowcolor{zebrarow}
\makecell[l]{VideoLLaMA3-7B\\[-1pt]{\scriptsize \citep{zhang2025videollama3frontiermultimodal}}} & 7B & V & 12.47 & 16.72 & 13.62 & 23.05 & 11.38 & 15.77 & 22.25 & 16.47 \\
\makecell[l]{Monet-7B\\[-1pt]{\scriptsize \citep{wang2025monetreasoninglatentvisual}}} & 7B & V & 10.74 & 14.83 & 11.28 & 18.70 & 8.52 & 13.96 & 19.18 & 13.89 \\
\rowcolor{zebrarow}
\makecell[l]{LLaVA-OneVision-7B\\[-1pt]{\scriptsize \citep{li2024llavaonevisioneasyvisualtask}}} & 7B & V & 11.12 & 13.43 & 11.42 & 16.31 & 9.14 & 12.55 & 17.51 & 13.07 \\
\makecell[l]{Tempo-6B\\[-1pt]{\scriptsize \citep{fei2026small}}} & 6B & V & 6.32 & 9.66 & 7.16 & 16.74 & 6.13 & 10.05 & 13.73 & 9.97 \\
\rowcolor{zebrarow}
\makecell[l]{SenseNova-SI-1.3-8B\\[-1pt]{\scriptsize \citep{sensenova-si}}} & 8B & V & 7.62 & 10.03 & 7.48 & 13.97 & 5.64 & 9.19 & 14.51 & 9.78 \\
\makecell[l]{OneThinker-8B\\[-1pt]{\scriptsize \citep{feng2025onethinker}}} & 8B & V & 6.67 & 7.72 & 8.55 & 5.39 & 7.50 & 5.84 & 7.92 & 7.09 \\
\rowcolor{zebrarow}
\makecell[l]{VideoScore2\\[-1pt]{\scriptsize \citep{he2025videoscore2thinkscoregenerative}}} & 7B & V & 2.64 & 3.27 & 2.59 & 2.60 & 1.02 & 2.40 & 4.90 & 2.77 \\
\midrule
\multicolumn{11}{c}{\textit{\textbf{Frame-based VLMs}}} \\
\midrule
\rowcolor{zebrarow}
\makecell[l]{Gemma-3-27B\\[-1pt]{\scriptsize \citep{gemmateam2025gemma3technicalreport}}} & 27B & F & \cT 34.67 & \cT 47.20 & \cT 39.14 & \cT 45.22 & 37.86 & 36.67 & 50.13 & \cT 41.55 \\
\makecell[l]{Gemma-3-12B\\[-1pt]{\scriptsize \citep{gemmateam2025gemma3technicalreport}}} & 12B & F & \cT 32.08 & \cT 46.06 & \cT 36.69 & \cT 43.46 & 36.49 & 34.84 & 47.40 & \cT 39.58 \\
\rowcolor{zebrarow}
Gemma-4-31B & 31B & F & \cT 31.23 & \cT 43.59 & \cT 36.01 & \cT 43.14 & 34.22 & 35.96 & 45.29 & \cT 38.49 \\
\makecell[l]{Gemma-3-4B\\[-1pt]{\scriptsize \citep{gemmateam2025gemma3technicalreport}}} & 4B & F & \cT 28.89 & \cT 40.42 & \cT 32.87 & \cT 40.27 & 33.23 & 30.19 & 43.42 & \cT 35.61 \\
\rowcolor{zebrarow}
Gemma-4-26B-A4B & \makecell{25B\\(3.8B active)} & F & \cT 29.44 & \cT 39.37 & \cT 31.70 & \cH 36.61 & 30.52 & 31.31 & 41.02 & \cT 34.28 \\
\makecell[l]{Ovis2-16B\\[-1pt]{\scriptsize \citep{lu2024ovis}}} & 16B & F & \cM 20.28 & \cM 28.16 & \cM 20.87 & \cM 31.98 & 19.84 & 23.77 & 32.35 & \cM 25.32 \\
\rowcolor{zebrarow}
\makecell[l]{Kimi-VL-A3B-T\\[-1pt]{\scriptsize \citep{team2025kimi}}} & \makecell{8B\\(3B active)} & F & \cM 17.76 & \cM 27.49 & \cM 21.24 & \cM 30.77 & 17.89 & 22.75 & 32.31 & \cM 24.32 \\
\makecell[l]{Ovis2-8B\\[-1pt]{\scriptsize \citep{lu2024ovis}}} & 8B & F & 15.81 & 21.51 & 17.26 & 25.14 & 15.16 & 18.75 & 25.89 & 19.93 \\
\rowcolor{zebrarow}
\makecell[l]{Ovis2-4B\\[-1pt]{\scriptsize \citep{lu2024ovis}}} & 4B & F & 14.41 & 18.11 & 14.97 & 23.54 & 13.07 & 17.19 & 23.01 & 17.76 \\
\makecell[l]{Kimi-VL-A3B-I\\[-1pt]{\scriptsize \citep{team2025kimi}}} & \makecell{8B\\(3B active)} & F & 12.38 & 16.58 & 12.29 & 21.79 & 11.35 & 15.40 & 20.53 & 15.76 \\
\midrule
\multicolumn{11}{c}{\textit{\textbf{Omni-modal Models (native video + audio)}}} \\
\midrule
\rowcolor{zebrarow}
\makecell[l]{Qwen3-Omni-30B-A3B-T\\[-1pt]{\scriptsize \citep{xu2025qwen3}}} & \makecell{30B\\(3B active)} & V+A & \cT \underline{51.44} & \cT \underline{72.86} & \cT \underline{65.40} & \cT \textbf{66.50} & \underline{61.52} & \underline{61.95} & \underline{68.67} & \cT \underline{64.05} \\
\makecell[l]{Nemotron-3-Nano-Omni-30B-A3B-R\\[-1pt]{\scriptsize \citep{nvidia2026nemotron3nanoomni}}} & \makecell{30B\\(3B active)} & V+A & \cT 30.90 & \cT 50.39 & \cT 42.00 & \cT 40.55 & 34.16 & 39.93 & 48.80 & \cT 40.96 \\
\rowcolor{zebrarow}
\makecell[l]{Qwen3-Omni-30B-A3B-I\\[-1pt]{\scriptsize \citep{xu2025qwen3}}} & \makecell{30B\\(3B active)} & V+A & \cT 31.21 & \cT 46.92 & \cT 36.96 & \cT 41.59 & 33.76 & 37.57 & 46.18 & \cT 39.17 \\
Gemma-4-E4B & \makecell{8B\\(4.5B eff)} & V+A & \cH 22.27 & \cM 27.56 & \cH 24.41 & \cH 33.34 & 21.68 & 25.09 & 33.93 & \cM 26.90 \\
\rowcolor{zebrarow}
\makecell[l]{MiniCPM-o 4.5\\[-1pt]{\scriptsize \citep{yao2024minicpm}}} & 8B & V+A & \cM 20.45 & \cM 25.90 & \cM 23.17 & 21.74 & 17.05 & 20.97 & 30.44 & \cM 22.82 \\
Gemma-4-E2B & \makecell{5B\\(2.3B eff)} & V+A & \cM 19.20 & \cM 26.03 & \cM 20.88 & 23.67 & 17.63 & 20.20 & 29.50 & \cM 22.44 \\
\rowcolor{zebrarow}
\makecell[l]{EchoInk-R1-7B\\[-1pt]{\scriptsize \citep{xing2025echoinkr1exploringaudiovisualreasoning}}} & 7B & V+A & \cM 18.10 & \cM 23.02 & \cM 19.95 & \cM 28.28 & 16.65 & 21.45 & 28.91 & \cM 22.34 \\
\makecell[l]{HumanSense Omni-R\\[-1pt]{\scriptsize \citep{qin2025humansense}}} & 7B & V+A & \cM 18.17 & \cM 23.30 & \cM 19.82 & \cM 27.61 & 16.54 & 21.62 & 28.52 & \cM 22.23 \\
\rowcolor{zebrarow}
\makecell[l]{Qwen2.5-Omni-7B\\[-1pt]{\scriptsize \citep{xu2025qwen25omni}}} & 7B & V+A & \cM 17.69 & 22.72 & \cM 20.27 & \cM 27.20 & 15.89 & 21.19 & 28.82 & 21.97 \\
\makecell[l]{Baichuan-Omni 1.5\\[-1pt]{\scriptsize \citep{li2025baichuan}}} & 7B & V+A & 15.98 & 19.46 & 15.75 & \cM 26.73 & 14.10 & 18.78 & 25.54 & 19.47 \\
\rowcolor{zebrarow}
\makecell[l]{Qwen2.5-Omni-3B\\[-1pt]{\scriptsize \citep{xu2025qwen25omni}}} & 3B & V+A & 15.86 & 19.39 & 16.79 & 25.01 & 13.51 & 19.01 & 25.27 & 19.26 \\
\makecell[l]{MiniCPM-o 2.6\\[-1pt]{\scriptsize \citep{yao2024minicpm}}} & 8B & V+A & 15.38 & 19.23 & 16.22 & 25.94 & 14.41 & 18.24 & 24.92 & 19.19 \\
\rowcolor{zebrarow}
\makecell[l]{SALMONN2+ 7B\\[-1pt]{\scriptsize \citep{tang2025videosalmonn2captionenhancedaudiovisual}}} & 7B & V+A & 14.00 & 19.67 & 15.91 & 22.37 & 11.29 & 17.50 & 25.17 & 17.99 \\
\makecell[l]{OmniVinci-9B\\[-1pt]{\scriptsize \citep{ye2025omnivinci}}} & 9B & V+A & 13.34 & 16.75 & 14.59 & 20.33 & 12.35 & 14.17 & 22.23 & 16.25 \\
\rowcolor{zebrarow}
\makecell[l]{OmniAtlas-Qwen2.5-7B\\[-1pt]{\scriptsize \citep{li2026omnigaia}}} & 7B & V+A & 14.12 & 15.95 & 15.86 & 15.79 & 12.06 & 14.44 & 19.79 & 15.43 \\
\midrule
\multicolumn{11}{c}{\textit{\textbf{Agentic Systems}}} \\
\midrule
\rowcolor{ourscell}
\textbf{LongShOTAgent (Ours)} & Agent & V+A & \textbf{55.28} & \textbf{75.69} & \textbf{69.57} & \textbf{66.00} & \textbf{65.69} & \textbf{62.46} & \textbf{71.76} & \textbf{66.64} \\
\rowcolor{zebrarow}
\makecell[l]{VideoMind\\[-1pt]{\scriptsize \citep{liu2025videomind}}} & Agent & V+A & 8.22 & 9.90 & 8.74 & 16.43 & 7.36 & 10.60 & 14.52 & 10.83 \\
\makecell[l]{Video-RAG\\[-1pt]{\scriptsize \citep{luo2024video}}} & Agent & V+A & 5.41 & 5.99 & 7.35 & 7.88 & 4.06 & 6.05 & 9.87 & 6.66 \\
\rowcolor{zebrarow}
\makecell[l]{Vgent\\[-1pt]{\scriptsize \citep{shen2025vgent}}} & Agent & V+A & 3.44 & 5.43 & 4.29 & 10.15 & 4.68 & 5.74 & 7.07 & 5.83 \\
\midrule
\multicolumn{11}{c}{\textit{\textbf{Audio-Only LLMs}}} \\
\midrule
\rowcolor{zebrarow}
\makecell[l]{Voxtral-Mini-3B-2507\\[-1pt]{\scriptsize \citep{mistralai2026voxtraltts}}} & 3B & Audio & \cT 33.92 & \cT 53.31 & \cT 42.90 & \cT 40.41 & 42.63 & 47.16 & 58.03 & \cT 49.27 \\
\makecell[l]{Voxtral-Small-24B-2507\\[-1pt]{\scriptsize \citep{mistralai2026voxtraltts}}} & 24B & Audio & \cT 32.63 & \cT 51.65 & \cT 41.38 & \cT 43.53 & 42.30 & 46.13 & 56.80 & \cT 48.41 \\
\rowcolor{zebrarow}
\makecell[l]{Fun-Audio-Chat-8B\\[-1pt]{\scriptsize \citep{tongyifunteam2026funaudiochattechnicalreport}}} & 8B & Audio & \cH 19.92 & \cH 28.85 & \cH 22.93 & \cH 30.69 & 25.60 & 29.53 & 37.69 & \cH 30.94 \\
\makecell[l]{Audio Flamingo 3\\[-1pt]{\scriptsize \citep{goel2025audio}}} & 8B & Audio & \cM 13.50 & \cM 18.18 & \cM 14.54 & \cM 22.67 & 17.22 & 22.00 & 29.49 & \cM 22.90 \\
\rowcolor{zebrarow}
\makecell[l]{Qwen2-Audio-7B-I\\[-1pt]{\scriptsize \citep{chu2024qwen2}}} & 7B & Audio & \cM 6.04 & \cM 7.61 & \cM 6.16 & \cM 12.47 & 8.07 & 13.27 & 18.21 & \cM 13.18 \\
\midrule
\multicolumn{11}{c}{\textit{\textbf{Proprietary (Closed-source API)}}} \\
\midrule
\rowcolor{zebrarow}
\makecell[l]{Gemini 3.1 Pro Preview\\[-1pt]{\scriptsize \citep{google2025gemini3}}} & API & V+A & \cT 43.01 & \cT 65.82 & \cT 55.57 & \cT 58.15 & 51.78 & 53.81 & 61.31 & \cT 55.63 \\
\end{longtable}
\endgroup

\begingroup
\captionsetup{font=normalsize,labelfont={normalsize,bf},justification=justified,singlelinecheck=false}
\setlength{\tabcolsep}{5pt}
\renewcommand{\arraystretch}{1.25}
\footnotesize
\begin{longtable}{@{}>{\raggedright}p{4.2cm} ccc ccc@{}}
\caption{\textbf{Full LongShOTBench leaderboard: Per-modality \& per-duration scores (\%).} All values are 3-judge means. Modality columns are restricted to questions tagged with that modality; duration columns to videos in that bucket. \textbf{V}: Visual, \textbf{A}: Audio (non-speech), \textbf{S}: Speech. \textbf{Short}: $<$\,20\,min, \textbf{Med}: 20-45\,min, \textbf{Long}: $>$\,45\,min. Variant suffixes: \texttt{-T}=Thinking, \texttt{-I}=Instruct, \texttt{-R}=Reasoning.}
\label{tab:full_benchmark_modality} \\
\toprule
& \multicolumn{3}{c}{\textbf{Modality}} & \multicolumn{3}{c}{\textbf{Duration}} \\
\cmidrule(lr){2-4}\cmidrule(lr){5-7}
\textbf{Model} & \textbf{Visual} & \textbf{Audio} & \textbf{Speech} & \textbf{Short} & \textbf{Medium} & \textbf{Long} \\
\midrule
\endfirsthead
\multicolumn{7}{l}{\itshape Continued from previous page} \\
\toprule
& \multicolumn{3}{c}{\textbf{Modality}} & \multicolumn{3}{c}{\textbf{Duration}} \\
\cmidrule(lr){2-4}\cmidrule(lr){5-7}
\textbf{Model} & \textbf{Visual} & \textbf{Audio} & \textbf{Speech} & \textbf{Short} & \textbf{Medium} & \textbf{Long} \\
\midrule
\endhead
\multicolumn{7}{r}{\itshape Continued on next page} \\
\endfoot
\bottomrule
\endlastfoot
\midrule
\multicolumn{7}{c}{\textit{\textbf{Video-native VLMs}}} \\
\midrule
\rowcolor{zebrarow}
\makecell[l]{Qwen3-VL-235B-A22B-T\\[-1pt]{\scriptsize \citep{bai2025qwen3}}} & \cT 45.48 & \cT 43.99 & \cT 46.22 & \cT 43.68 & \cT 44.31 & \cT 48.04 \\
\makecell[l]{Qwen3-VL-32B-T\\[-1pt]{\scriptsize \citep{bai2025qwen3}}} & \cT 44.13 & \cT 43.10 & \cT 44.81 & \cT 43.05 & \cT 43.53 & \cT 45.49 \\
\rowcolor{zebrarow}
\makecell[l]{Vero-Qwen3I-8B\\[-1pt]{\scriptsize \citep{sarch2026vero}}} & \cT 43.57 & \cT 43.12 & \cT 43.99 & \cT 44.58 & \cT 43.32 & \cT 44.09 \\
\makecell[l]{Kimi-K2.6\\[-1pt]{\scriptsize \citep{team2026kimi}}} & \cT 42.88 & \cT 41.49 & \cT 43.62 & \cT 43.32 & \cT 43.15 & \cT 42.12 \\
\rowcolor{zebrarow}
\makecell[l]{Qwen3-VL-30B-A3B-T\\[-1pt]{\scriptsize \citep{bai2025qwen3}}} & \cT 41.29 & \cT 39.78 & \cT 42.06 & \cT 40.37 & \cT 40.55 & \cT 43.00 \\
\makecell[l]{Qwen3-VL-235B-A22B-I\\[-1pt]{\scriptsize \citep{bai2025qwen3}}} & \cT 40.05 & \cT 38.80 & \cT 40.54 & \cT 45.42 & \cT 39.44 & \cT 41.20 \\
\rowcolor{zebrarow}
\makecell[l]{Qwen3-VL-8B-T\\[-1pt]{\scriptsize \citep{bai2025qwen3}}} & \cT 39.48 & \cT 38.27 & \cT 40.12 & \cT 41.73 & \cT 38.92 & \cT 40.69 \\
\makecell[l]{Intern-S1-mini\\[-1pt]{\scriptsize \citep{bai2025intern}}} & \cT 39.14 & \cT 37.55 & \cT 39.70 & \cT 38.78 & \cT 38.48 & \cT 40.72 \\
\rowcolor{zebrarow}
\makecell[l]{Qwen3-VL-32B-I\\[-1pt]{\scriptsize \citep{bai2025qwen3}}} & \cT 37.78 & \cT 36.35 & \cT 38.16 & \cT 42.04 & \cT 37.78 & \cT 37.43 \\
\makecell[l]{Step3-VL-10B\\[-1pt]{\scriptsize \citep{huang2026step3vl10btechnicalreport}}} & \cT 36.10 & \cT 34.35 & \cT 36.60 & \cT 36.20 & \cT 35.21 & \cT 38.04 \\
\rowcolor{zebrarow}
\makecell[l]{Qwen3-VL-4B-T\\[-1pt]{\scriptsize \citep{bai2025qwen3}}} & \cT 35.61 & \cT 34.39 & \cT 36.07 & \cT 37.76 & \cT 35.14 & \cT 36.51 \\
\makecell[l]{GLM-4.5V\\[-1pt]{\scriptsize \citep{hong2025glm}}} & \cT 35.41 & \cT 32.66 & \cT 36.01 & \cT 37.34 & \cT 34.54 & \cT 37.27 \\
\rowcolor{zebrarow}
\makecell[l]{MiMo-VL-7B-RL-2508\\[-1pt]{\scriptsize \citep{coreteam2025mimovltechnicalreport}}} & \cT 34.86 & \cT 33.19 & \cT 35.45 & \cT 35.13 & \cT 34.36 & \cT 35.90 \\
\makecell[l]{Qwen2.5-VL-32B-I\\[-1pt]{\scriptsize \citep{bai2025qwen2}}} & \cT 33.96 & \cT 31.70 & \cT 34.39 & \cT 35.77 & \cT 33.35 & \cT 35.25 \\
\rowcolor{zebrarow}
\makecell[l]{Qwen3-VL-8B-I\\[-1pt]{\scriptsize \citep{bai2025qwen3}}} & \cT 31.36 & \cT 30.27 & \cT 31.77 & \cT 35.63 & \cT 31.15 & \cT 31.69 \\
\makecell[l]{Ovis2.6-30B-A3B\\[-1pt]{\scriptsize \citep{lu2025ovis25technicalreport}}} & \cH 31.15 & \cH 29.24 & \cH 31.63 & \cT 36.11 & \cH 30.77 & \cH 31.64 \\
\rowcolor{zebrarow}
\makecell[l]{GLM-4.6V-Flash\\[-1pt]{\scriptsize \citep{hong2025glm}}} & \cH 30.91 & \cH 29.47 & \cH 31.42 & \cH 32.55 & \cH 30.08 & \cT 32.75 \\
\makecell[l]{Qwen3.6-27B\\[-1pt]{\scriptsize \citep{qwen36_35b_a3b}}} & \cH 29.69 & \cH 28.30 & \cH 30.26 & \cH 28.04 & \cH 29.45 & \cH 30.24 \\
\rowcolor{zebrarow}
\makecell[l]{Qwen2.5-VL-72B-I\\[-1pt]{\scriptsize \citep{bai2025qwen2}}} & \cH 30.01 & \cH 27.48 & \cH 30.42 & \cH 29.77 & \cH 29.36 & \cH 31.44 \\
\makecell[l]{Qwen3.6-35B-A3B\\[-1pt]{\scriptsize \citep{qwen36_35b_a3b}}} & \cH 27.63 & \cH 25.93 & \cH 28.22 & \cH 32.05 & \cH 27.39 & \cH 27.98 \\
\rowcolor{zebrarow}
\makecell[l]{Qwen3.5-27B\\[-1pt]{\scriptsize \citep{qwen3.5}}} & \cH 28.21 & \cH 25.50 & \cH 28.82 & \cH 29.34 & \cH 27.65 & \cH 29.49 \\
\makecell[l]{Qwen3.5-122B-A10B\\[-1pt]{\scriptsize \citep{qwen3.5}}} & \cH 27.82 & \cH 25.44 & \cH 28.42 & \cH 33.27 & \cH 27.37 & \cH 28.56 \\
\rowcolor{zebrarow}
\makecell[l]{Qwen3.5-35B-A3B\\[-1pt]{\scriptsize \citep{qwen3.5}}} & \cH 26.98 & \cH 24.83 & \cH 27.59 & \cH 29.30 & \cH 26.45 & \cH 28.14 \\
\makecell[l]{R-4B\\[-1pt]{\scriptsize \citep{yang2025r4bincentivizinggeneralpurposeautothinking}}} & \cH 28.49 & \cH 27.02 & \cH 28.91 & \cT 34.85 & \cH 27.36 & \cH 30.82 \\
\rowcolor{zebrarow}
\makecell[l]{Qwen3-VL-30B-A3B-I\\[-1pt]{\scriptsize \citep{bai2025qwen3}}} & \cH 29.01 & \cH 28.08 & \cH 29.36 & \cH 32.94 & \cH 28.72 & \cH 29.45 \\
\makecell[l]{GLM-4.1V-9B-T\\[-1pt]{\scriptsize \citep{hong2025glm}}} & \cH 27.33 & \cH 25.82 & \cH 27.73 & \cH 27.22 & \cH 26.74 & \cH 28.72 \\
\rowcolor{zebrarow}
\makecell[l]{InternVL3.5-241B-A28B\\[-1pt]{\scriptsize \citep{wang2025internvl3_5}}} & \cH 27.31 & \cH 25.41 & \cH 27.71 & \cH 30.66 & \cH 26.81 & \cH 28.30 \\
\makecell[l]{InternVL3-78B\\[-1pt]{\scriptsize \citep{zhu2025internvl3}}} & \cH 26.76 & \cH 24.84 & \cH 27.11 & \cH 30.72 & \cH 26.35 & \cH 27.38 \\
\rowcolor{zebrarow}
\makecell[l]{InternVL3-38B\\[-1pt]{\scriptsize \citep{zhu2025internvl3}}} & \cH 26.78 & \cH 24.80 & \cH 27.09 & \cH 30.17 & \cH 26.37 & \cH 27.59 \\
\makecell[l]{Qwen3-VL-4B-I\\[-1pt]{\scriptsize \citep{bai2025qwen3}}} & \cH 26.06 & \cH 24.88 & \cH 26.34 & \cH 31.86 & \cH 25.71 & \cH 26.72 \\
\rowcolor{zebrarow}
\makecell[l]{VideoRFT-7B\\[-1pt]{\scriptsize \citep{wang2025videorft}}} & \cH 26.89 & \cH 25.32 & \cH 27.11 & \cH 28.85 & \cH 26.65 & \cH 27.27 \\
\makecell[l]{DeepEyes-7B\\[-1pt]{\scriptsize \citep{zheng2025deepeyesincentivizingthinkingimages}}} & \cH 26.03 & \cH 24.50 & \cH 26.32 & \cH 29.52 & \cH 25.47 & \cH 27.17 \\
\rowcolor{zebrarow}
\makecell[l]{Ovis2.5-9B\\[-1pt]{\scriptsize \citep{lu2025ovis25technicalreport}}} & \cM 25.16 & \cH 23.99 & \cM 25.48 & \cH 29.24 & \cM 24.58 & \cH 26.40 \\
\makecell[l]{Video-R2\\[-1pt]{\scriptsize \citep{maaz2025video}}} & \cH 25.91 & \cH 24.78 & \cH 26.19 & \cM 24.40 & \cH 25.64 & \cH 26.63 \\
\rowcolor{zebrarow}
\makecell[l]{Molmo2-8B\\[-1pt]{\scriptsize \citep{clark2026molmo2openweightsdata}}} & \cH 25.32 & \cH 24.21 & \cH 25.50 & \cH 31.93 & \cH 25.10 & \cM 25.45 \\
\makecell[l]{Video-R1-7B\\[-1pt]{\scriptsize \citep{feng2025videor1reinforcingvideoreasoning}}} & \cH 25.43 & \cH 24.42 & \cH 25.68 & \cM 25.88 & \cH 24.97 & \cH 26.51 \\
\rowcolor{zebrarow}
\makecell[l]{LongVT-RL\\[-1pt]{\scriptsize \citep{yang2025longvtincentivizingthinkinglong}}} & \cH 25.59 & \cH 25.47 & \cH 25.77 & \cH 27.38 & \cH 25.46 & \cM 25.69 \\
\makecell[l]{RePlan-Qwen2.5-VL-7B\\[-1pt]{\scriptsize \citep{qu2025replan}}} & \cH 25.34 & \cH 23.91 & \cH 25.61 & \cH 27.49 & \cH 25.05 & \cH 25.91 \\
\rowcolor{zebrarow}
\makecell[l]{InternVL3-14B\\[-1pt]{\scriptsize \citep{zhu2025internvl3}}} & \cH 25.41 & \cM 23.48 & \cH 25.71 & \cM 25.98 & \cM 24.88 & \cH 26.58 \\
\makecell[l]{Qwen3.5-9B\\[-1pt]{\scriptsize \citep{qwen3.5}}} & \cM 23.57 & \cM 21.81 & \cM 24.04 & \cM 25.88 & \cM 23.31 & \cM 24.05 \\
\rowcolor{zebrarow}
\makecell[l]{TimeSearch-R\\[-1pt]{\scriptsize \citep{pan2025timesearch}}} & \cH 25.28 & \cM 23.70 & \cH 25.53 & \cM 24.37 & \cH 24.97 & \cH 26.05 \\
\makecell[l]{ERNIE-4.5-VL-28B-A3B\\[-1pt]{\scriptsize \citep{ernie2025technicalreport}}} & \cH 25.72 & \cH 24.58 & \cH 26.08 & \cM 26.65 & \cH 25.02 & \cH 27.29 \\
\rowcolor{zebrarow}
\makecell[l]{InternVL3.5-8B\\[-1pt]{\scriptsize \citep{wang2025internvl3_5}}} & \cM 23.60 & \cM 22.12 & \cM 23.84 & \cM 25.47 & \cM 23.32 & \cM 24.08 \\
\makecell[l]{VideoChat-R1-7B\\[-1pt]{\scriptsize \citep{li2025videochat}}} & \cM 24.75 & \cM 23.30 & \cM 24.99 & \cM 22.13 & \cM 24.26 & \cH 25.96 \\
\rowcolor{zebrarow}
\makecell[l]{InternVL3.5-38B\\[-1pt]{\scriptsize \citep{wang2025internvl3_5}}} & \cM 23.92 & \cM 22.51 & \cM 24.21 & \cH 27.53 & \cM 23.52 & \cM 24.67 \\
\makecell[l]{PixelReasoner-RL-v1\\[-1pt]{\scriptsize \citep{wang2025pixel}}} & \cM 24.02 & \cM 22.60 & \cM 24.27 & \cM 25.58 & \cM 23.64 & \cM 24.89 \\
\rowcolor{zebrarow}
\makecell[l]{Qwen2.5-VL-7B-I\\[-1pt]{\scriptsize \citep{bai2025qwen2}}} & \cM 24.24 & \cM 23.09 & \cM 24.53 & \cH 26.76 & \cM 23.93 & \cM 24.85 \\
\makecell[l]{InternVL2.5-78B\\[-1pt]{\scriptsize \citep{chen2024expanding}}} & \cM 22.99 & \cM 21.56 & \cM 23.29 & \cM 25.70 & \cM 22.49 & \cM 24.05 \\
\rowcolor{zebrarow}
\makecell[l]{InternVL3-8B\\[-1pt]{\scriptsize \citep{zhu2025internvl3}}} & \cM 22.36 & \cM 21.38 & \cM 22.52 & \cM 25.33 & \cM 21.99 & \cM 23.09 \\
\makecell[l]{InternVL2.5-38B\\[-1pt]{\scriptsize \citep{chen2024expanding}}} & \cM 22.07 & \cM 20.85 & \cM 22.44 & \cM 22.62 & \cM 21.49 & \cM 23.44 \\
\rowcolor{zebrarow}
\makecell[l]{R1-Onevision-7B\\[-1pt]{\scriptsize \citep{yang2025r1}}} & \cM 21.99 & \cM 20.73 & \cM 22.24 & \cM 22.01 & \cM 21.71 & \cM 22.62 \\
\makecell[l]{Phi-4-R-Vision-15B\\[-1pt]{\scriptsize \citep{aneja2026phi4reasoningvision15btechnicalreport}}} & \cM 21.18 & \cM 20.59 & \cM 21.43 & \cM 25.45 & \cM 20.39 & \cM 22.74 \\
\rowcolor{zebrarow}
\makecell[l]{Qwen2.5-VL-7B-VRAG\\[-1pt]{\scriptsize \citep{wang2025vrag}}} & \cM 22.32 & \cM 22.28 & \cM 22.49 & 19.86 & \cM 21.95 & \cM 23.26 \\
\makecell[l]{AURA-32B\\[-1pt]{\scriptsize \citep{aura2026}}} & \cM 21.09 & \cM 20.59 & \cM 21.25 & \cM 21.38 & \cM 20.81 & \cM 21.73 \\
\rowcolor{zebrarow}
\makecell[l]{InternVL3.5-4B-I\\[-1pt]{\scriptsize \citep{wang2025internvl3_5}}} & \cM 21.03 & \cM 19.51 & \cM 21.27 & 20.88 & \cM 20.58 & \cM 22.12 \\
\makecell[l]{Qwen3.5-4B\\[-1pt]{\scriptsize \citep{qwen3.5}}} & \cM 19.99 & \cM 18.66 & \cM 20.30 & \cM 22.52 & \cM 20.07 & 19.48 \\
\rowcolor{zebrarow}
\makecell[l]{Keye-VL-1.5-8B\\[-1pt]{\scriptsize \citep{yang2025kwaikeyevl15technical}}} & \cM 19.90 & \cM 19.48 & \cM 20.17 & \cM 22.56 & \cM 19.35 & \cM 21.00 \\
\makecell[l]{NVIDIA Nemotron Nano 12B VL\\[-1pt]{\scriptsize \citep{nvidia2025nvidianemotronnanov2}}} & 19.46 & \cM 18.83 & 19.79 & 16.58 & 18.93 & \cM 20.79 \\
\rowcolor{zebrarow}
\makecell[l]{MiniCPM-V 4.5\\[-1pt]{\scriptsize \citep{yao2024minicpm}}} & 19.37 & \cM 19.00 & 19.70 & \cM 21.79 & 18.92 & \cM 20.43 \\
SpaceThinker-3B & 18.89 & 17.47 & 19.07 & 19.98 & 18.25 & \cM 20.33 \\
\rowcolor{zebrarow}
\makecell[l]{Qwen3-VL-2B-I\\[-1pt]{\scriptsize \citep{bai2025qwen3}}} & 18.53 & 18.00 & 18.65 & \cM 21.24 & 18.07 & 19.53 \\
\makecell[l]{Qwen2.5-VL-3B-I\\[-1pt]{\scriptsize \citep{bai2025qwen2}}} & 17.97 & 17.36 & 18.15 & 14.63 & 17.61 & 19.04 \\
\rowcolor{zebrarow}
\makecell[l]{Qwen3.5-2B\\[-1pt]{\scriptsize \citep{qwen3.5}}} & 18.14 & 17.55 & 18.31 & 18.23 & 17.91 & 18.61 \\
\makecell[l]{Qwen3.5-0.8B\\[-1pt]{\scriptsize \citep{qwen3.5}}} & 16.05 & 15.73 & 16.18 & 15.69 & 15.84 & 16.57 \\
\rowcolor{zebrarow}
\makecell[l]{VideoLLaMA3-7B\\[-1pt]{\scriptsize \citep{zhang2025videollama3frontiermultimodal}}} & 13.74 & 13.32 & 13.88 & 13.20 & 13.31 & 14.76 \\
\makecell[l]{Monet-7B\\[-1pt]{\scriptsize \citep{wang2025monetreasoninglatentvisual}}} & 12.27 & 12.15 & 12.43 & 10.84 & 11.76 & 13.54 \\
\rowcolor{zebrarow}
\makecell[l]{LLaVA-OneVision-7B\\[-1pt]{\scriptsize \citep{li2024llavaonevisioneasyvisualtask}}} & 11.79 & 11.32 & 11.98 & 13.04 & 11.62 & 12.20 \\
\makecell[l]{Tempo-6B\\[-1pt]{\scriptsize \citep{fei2026small}}} & 7.28 & 5.45 & 7.33 & 11.07 & 7.23 & 7.38 \\
\rowcolor{zebrarow}
\makecell[l]{SenseNova-SI-1.3-8B\\[-1pt]{\scriptsize \citep{sensenova-si}}} & 8.50 & 8.81 & 8.69 & 6.00 & 8.01 & 9.73 \\
\makecell[l]{OneThinker-8B\\[-1pt]{\scriptsize \citep{feng2025onethinker}}} & 7.36 & 7.50 & 7.48 & 6.17 & 6.78 & 8.69 \\
\rowcolor{zebrarow}
\makecell[l]{VideoScore2\\[-1pt]{\scriptsize \citep{he2025videoscore2thinkscoregenerative}}} & 2.76 & 2.54 & 2.68 & 4.50 & 2.86 & 2.46 \\
\midrule
\multicolumn{7}{c}{\textit{\textbf{Frame-based VLMs}}} \\
\midrule
\rowcolor{zebrarow}
\makecell[l]{Gemma-3-27B\\[-1pt]{\scriptsize \citep{gemmateam2025gemma3technicalreport}}} & \cT 39.34 & \cT 37.41 & \cT 39.99 & \cT 40.63 & \cT 38.26 & \cT 41.71 \\
\makecell[l]{Gemma-3-12B\\[-1pt]{\scriptsize \citep{gemmateam2025gemma3technicalreport}}} & \cT 36.92 & \cT 35.44 & \cT 37.60 & \cH 34.42 & \cT 36.01 & \cT 39.11 \\
\rowcolor{zebrarow}
Gemma-4-31B & \cT 35.99 & \cT 34.67 & \cT 36.51 & \cT 40.83 & \cT 35.37 & \cT 36.85 \\
\makecell[l]{Gemma-3-4B\\[-1pt]{\scriptsize \citep{gemmateam2025gemma3technicalreport}}} & \cT 33.19 & \cT 31.27 & \cT 33.72 & \cT 35.60 & \cT 32.45 & \cT 34.77 \\
\rowcolor{zebrarow}
Gemma-4-26B-A4B & \cT 32.34 & \cT 31.38 & \cT 32.69 & \cT 34.93 & \cT 32.61 & \cH 31.29 \\
\makecell[l]{Ovis2-16B\\[-1pt]{\scriptsize \citep{lu2024ovis}}} & \cM 22.86 & \cM 21.84 & \cM 23.17 & \cM 25.72 & \cM 22.51 & \cM 23.57 \\
\rowcolor{zebrarow}
\makecell[l]{Kimi-VL-A3B-T\\[-1pt]{\scriptsize \citep{team2025kimi}}} & \cM 21.23 & \cM 19.57 & \cM 21.67 & \cM 23.32 & \cM 21.04 & \cM 21.43 \\
\makecell[l]{Ovis2-8B\\[-1pt]{\scriptsize \citep{lu2024ovis}}} & 17.65 & 17.55 & 17.84 & 18.15 & 16.99 & 19.15 \\
\rowcolor{zebrarow}
\makecell[l]{Ovis2-4B\\[-1pt]{\scriptsize \citep{lu2024ovis}}} & 15.56 & 15.52 & 15.65 & 16.01 & 14.88 & 17.11 \\
\makecell[l]{Kimi-VL-A3B-I\\[-1pt]{\scriptsize \citep{team2025kimi}}} & 13.53 & 13.26 & 13.76 & 11.27 & 13.23 & 14.26 \\
\midrule
\multicolumn{7}{c}{\textit{\textbf{Omni-modal Models (native video + audio)}}} \\
\midrule
\rowcolor{zebrarow}
\makecell[l]{Qwen3-Omni-30B-A3B-T\\[-1pt]{\scriptsize \citep{xu2025qwen3}}} & \cT \underline{62.09} & \cT \underline{55.89} & \cT \underline{63.74} & \cT \underline{62.22} & \cT \underline{62.93} & \cT \underline{60.07} \\
\makecell[l]{Nemotron-3-Nano-Omni-30B-A3B-R\\[-1pt]{\scriptsize \citep{nvidia2026nemotron3nanoomni}}} & \cT 38.89 & \cT 32.89 & \cT 39.98 & \cT 43.52 & \cT 38.91 & \cT 38.62 \\
\rowcolor{zebrarow}
\makecell[l]{Qwen3-Omni-30B-A3B-I\\[-1pt]{\scriptsize \citep{xu2025qwen3}}} & \cT 36.27 & \cT 32.77 & \cT 37.20 & \cH 29.64 & \cT 36.18 & \cT 36.89 \\
Gemma-4-E4B & \cM 25.18 & \cH 24.77 & \cM 25.48 & \cH 30.36 & \cH 24.90 & \cM 25.47 \\
\rowcolor{zebrarow}
\makecell[l]{MiniCPM-o 4.5\\[-1pt]{\scriptsize \citep{yao2024minicpm}}} & \cM 22.38 & \cM 21.43 & \cM 22.67 & \cM 22.03 & \cM 22.07 & \cM 23.12 \\
Gemma-4-E2B & \cM 22.02 & \cM 20.89 & \cM 22.27 & \cH 31.66 & \cM 21.89 & \cM 21.72 \\
\rowcolor{zebrarow}
\makecell[l]{EchoInk-R1-7B\\[-1pt]{\scriptsize \citep{xing2025echoinkr1exploringaudiovisualreasoning}}} & \cM 20.28 & \cM 18.29 & \cM 20.65 & \cM 22.70 & \cM 20.40 & 20.06 \\
\makecell[l]{HumanSense Omni-R\\[-1pt]{\scriptsize \citep{qin2025humansense}}} & \cM 20.34 & 17.96 & \cM 20.71 & \cM 22.81 & \cM 20.11 & \cM 20.99 \\
\rowcolor{zebrarow}
\makecell[l]{Qwen2.5-Omni-7B\\[-1pt]{\scriptsize \citep{xu2025qwen25omni}}} & \cM 20.12 & 18.17 & \cM 20.48 & \cM 21.80 & \cM 20.22 & 20.00 \\
\makecell[l]{Baichuan-Omni 1.5\\[-1pt]{\scriptsize \citep{li2025baichuan}}} & 17.31 & 16.60 & 17.47 & 18.35 & 16.80 & 18.44 \\
\rowcolor{zebrarow}
\makecell[l]{Qwen2.5-Omni-3B\\[-1pt]{\scriptsize \citep{xu2025qwen25omni}}} & 17.35 & 15.74 & 17.66 & 19.31 & 17.07 & 18.05 \\
\makecell[l]{MiniCPM-o 2.6\\[-1pt]{\scriptsize \citep{yao2024minicpm}}} & 17.00 & 16.27 & 17.22 & 14.31 & 16.80 & 17.62 \\
\rowcolor{zebrarow}
\makecell[l]{SALMONN2+ 7B\\[-1pt]{\scriptsize \citep{tang2025videosalmonn2captionenhancedaudiovisual}}} & 15.91 & 13.40 & 16.30 & 16.17 & 15.73 & 16.49 \\
\makecell[l]{OmniVinci-9B\\[-1pt]{\scriptsize \citep{ye2025omnivinci}}} & 14.09 & 13.56 & 14.38 & 15.53 & 14.03 & 14.17 \\
\rowcolor{zebrarow}
\makecell[l]{OmniAtlas-Qwen2.5-7B\\[-1pt]{\scriptsize \citep{li2026omnigaia}}} & 15.51 & 15.03 & 15.75 & 16.83 & 15.55 & 15.43 \\
\midrule
\multicolumn{7}{c}{\textit{\textbf{Agentic Systems}}} \\
\midrule
\rowcolor{ourscell}
\textbf{LongShOTAgent (Ours)} & \textbf{64.96} & \textbf{57.61} & \textbf{66.76} & \textbf{66.43} & \textbf{65.60} & \textbf{63.50} \\
\rowcolor{zebrarow}
\makecell[l]{VideoMind\\[-1pt]{\scriptsize \citep{liu2025videomind}}} & 8.77 & 7.03 & 8.76 & 13.37 & 8.82 & 8.56 \\
\makecell[l]{Video-RAG\\[-1pt]{\scriptsize \citep{luo2024video}}} & 6.12 & 4.50 & 6.34 & 9.48 & 6.07 & 6.31 \\
\rowcolor{zebrarow}
\makecell[l]{Vgent\\[-1pt]{\scriptsize \citep{shen2025vgent}}} & 4.13 & 3.07 & 4.24 & 6.50 & 4.23 & 3.98 \\
\midrule
\multicolumn{7}{c}{\textit{\textbf{Audio-Only LLMs}}} \\
\midrule
\rowcolor{zebrarow}
\makecell[l]{Voxtral-Mini-3B-2507\\[-1pt]{\scriptsize \citep{mistralai2026voxtraltts}}} & \cT 41.43 & \cT 35.08 & \cT 42.85 & \cT 37.58 & \cT 41.22 & \cT 42.40 \\
\makecell[l]{Voxtral-Small-24B-2507\\[-1pt]{\scriptsize \citep{mistralai2026voxtraltts}}} & \cT 40.39 & \cT 33.10 & \cT 41.63 & \cT 35.87 & \cT 40.19 & \cT 41.13 \\
\rowcolor{zebrarow}
\makecell[l]{Fun-Audio-Chat-8B\\[-1pt]{\scriptsize \citep{tongyifunteam2026funaudiochattechnicalreport}}} & \cH 24.00 & \cH 22.13 & \cH 24.56 & \cH 24.90 & \cH 23.83 & \cH 24.40 \\
\makecell[l]{Audio Flamingo 3\\[-1pt]{\scriptsize \citep{goel2025audio}}} & \cM 15.32 & \cM 12.35 & \cM 15.76 & \cM 19.35 & \cM 15.37 & \cM 15.17 \\
\rowcolor{zebrarow}
\makecell[l]{Qwen2-Audio-7B-I\\[-1pt]{\scriptsize \citep{chu2024qwen2}}} & \cM 7.00 & \cM 6.88 & \cM 7.14 & \cM 6.19 & \cM 6.64 & \cM 7.92 \\
\midrule
\multicolumn{7}{c}{\textit{\textbf{Proprietary (Closed-source API)}}} \\
\midrule
\rowcolor{zebrarow}
\makecell[l]{Gemini 3.1 Pro Preview\\[-1pt]{\scriptsize \citep{google2025gemini3}}} & \cT 52.05 & \cT 46.15 & \cT 53.65 & \cT 50.05 & \cT 51.96 & \cT 52.57 \\
\end{longtable}
\endgroup

\section{Details of LongShOTBench}
\subsection{Dataset Statistics}

\begin{figure}[h]
  \centering
  \includegraphics[width=0.85\linewidth]{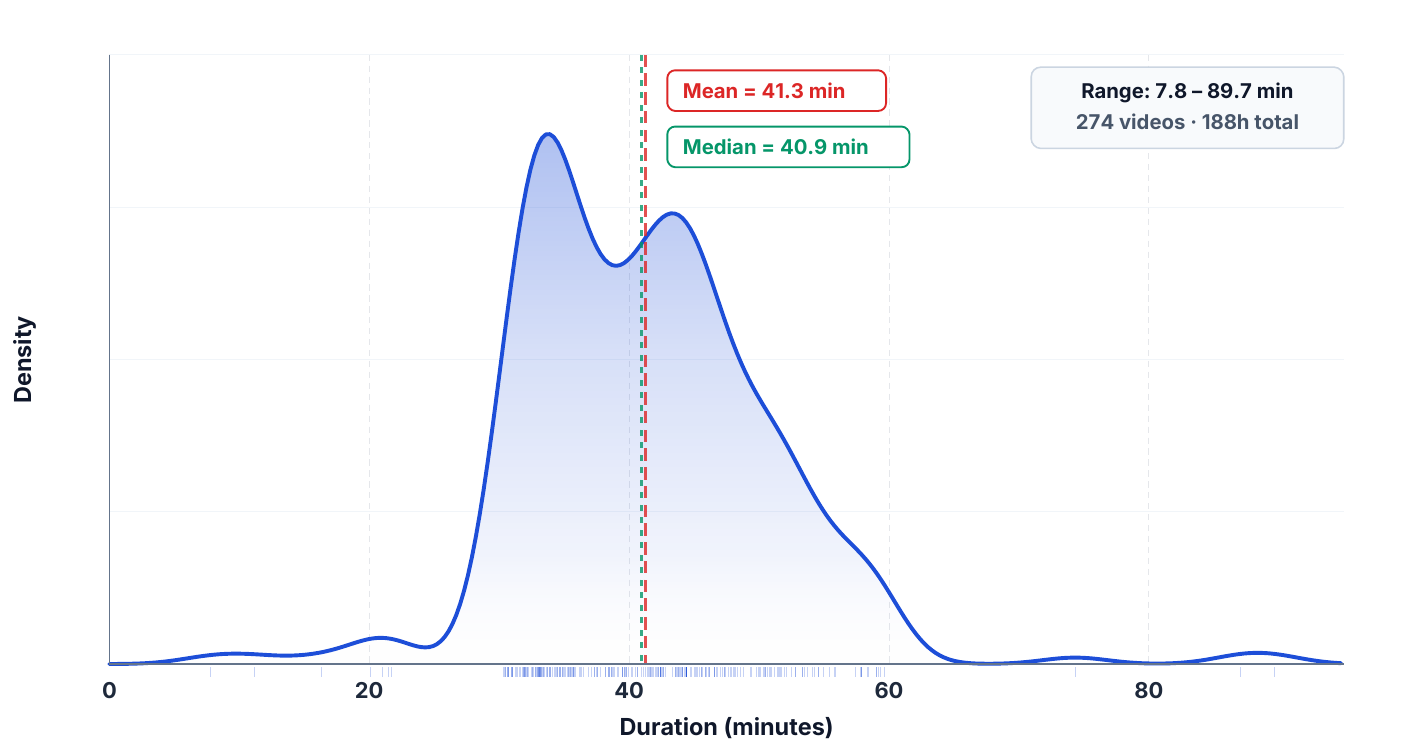}
  \caption{\textbf{Distribution of video durations (minutes)} in our validated sample set ($n{=}274$). Kernel density estimate with a per-video rug plot; dashed and dash-dot lines mark the mean (41.3 min) and median (40.9 min). Videos span 7.8-89.7 minutes, totaling $\sim$188 hours.}
  \label{fig:duration-distribution}
\end{figure}

The LongShOTBench dataset is composed of 274 long-form videos, specifically curated to evaluate the long-context understanding capabilities of Multimodal Large Language Models (MLLMs). The benchmark consists of 3{,}401 samples encompassing 4{,}893 question-answer turns. As illustrated in Fig.~\ref{fig:duration-distribution}, the benchmark has a distinct focus on videos of substantial length to rigorously test model performance on extended temporal sequences.
The video durations are heavily concentrated around a central peak, with a mean of 41.3 minutes and a median of 40.9 minutes, and span a range of 7.8-89.7 minutes for a total of about 188 hours of content. The close proximity of mean and median indicates a relatively symmetric distribution; the kernel density estimate exhibits a primary mode around 33-35 minutes and a secondary mode near 43-45 minutes, with the vast majority of videos lying between 30 and 60 minutes. This characteristic makes LongShOTBench a challenging testbed for evaluating multimodal integration, temporal reasoning, and information retention over significant durations.

\begin{figure}[h]
  \centering
  \includegraphics[width=0.85\linewidth]{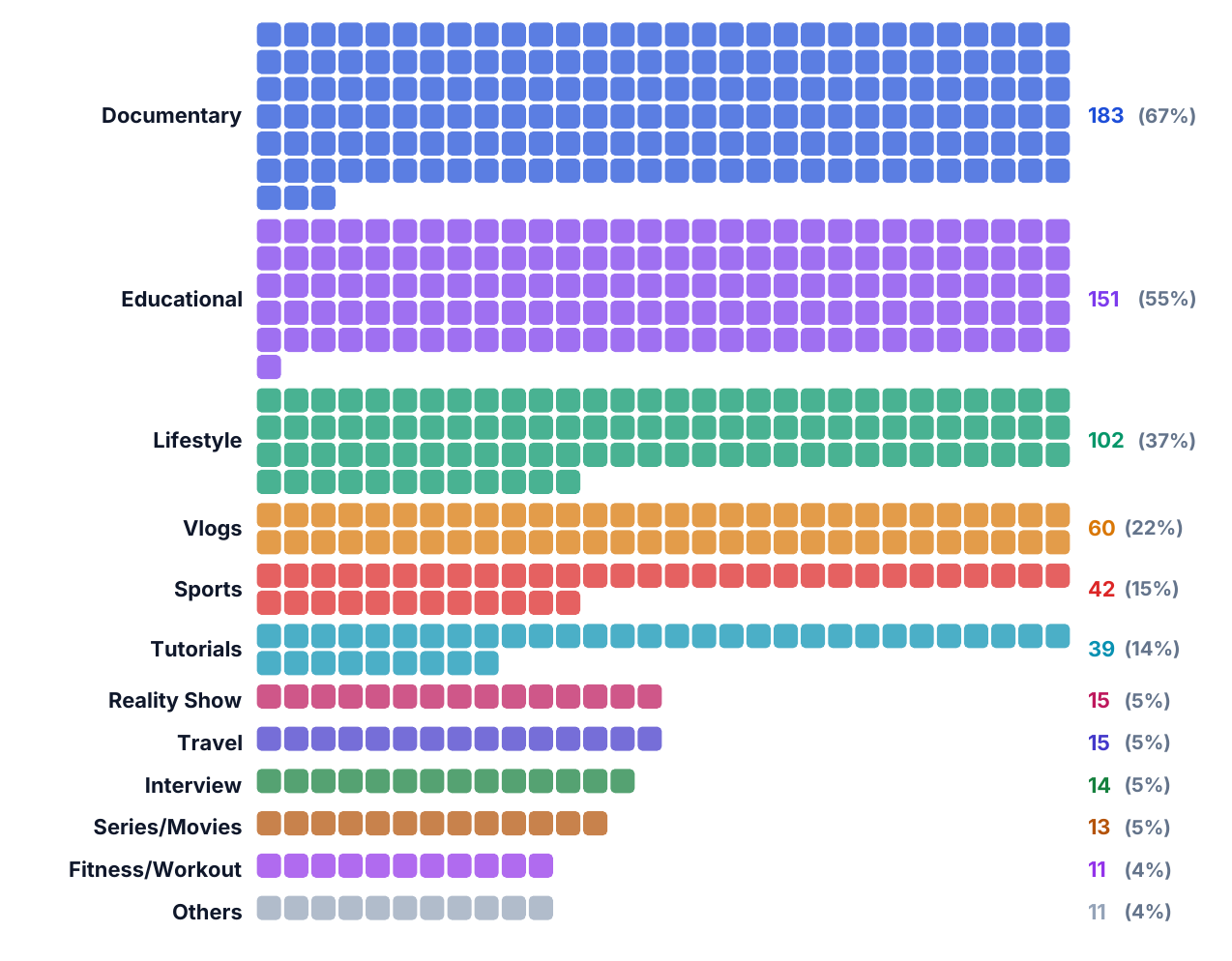}
  \caption{\textbf{Top-12 video genre distribution.} Each video may belong to multiple genres, so percentages are computed over the 274 unique videos and do not sum to 100\%. The benchmark is dominated by knowledge-rich genres (Documentary at 67\%, Educational at 55\%, and Lifestyle at 37\%), while still covering Vlogs, Sports, Tutorials, Reality Shows, Travel, Interviews, Series/Movies, and Fitness content, ensuring broad coverage across narrative styles.}
  \label{fig:sample-category-dist}
\end{figure}

\begin{figure}[h]
  \centering
  \includegraphics[width=0.85\linewidth]{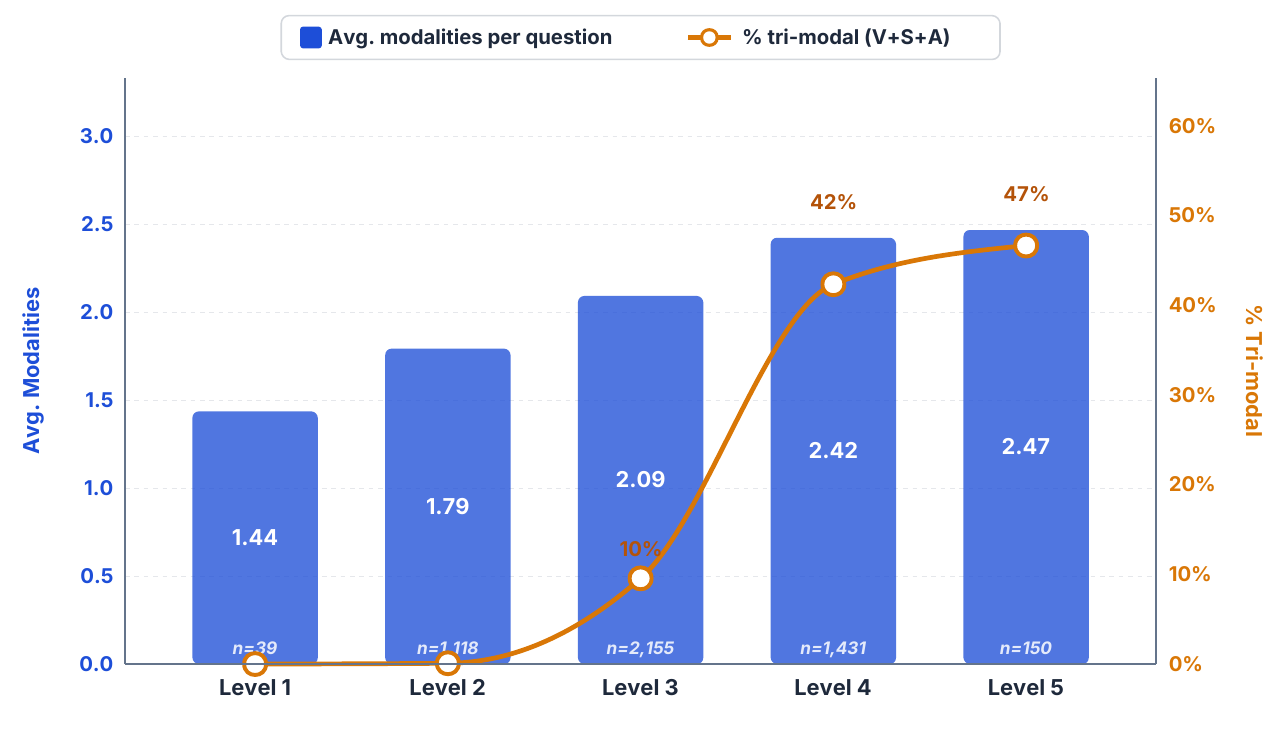}
  \caption{\textbf{Multimodal demand grows with question difficulty.} Bars (left axis) show the average number of canonical modalities (Visual, Speech, Audio) referenced by each question at each difficulty level; the line (right axis) shows the fraction of questions that require all three modalities simultaneously. Average modality count rises monotonically from 1.44 at Level~1 to 2.47 at Level~5, while tri-modal questions jump from $<$1\% at Levels~1-2 to 42.3\% and 46.7\% at Levels~4 and~5, directly evidencing that LongShOTBench's harder questions demand genuine cross-modal integration rather than single-stream perception.}
  \label{fig:difficulty-modality}
\end{figure}

\begin{figure}[h]
  \centering
  \includegraphics[width=0.85\linewidth]{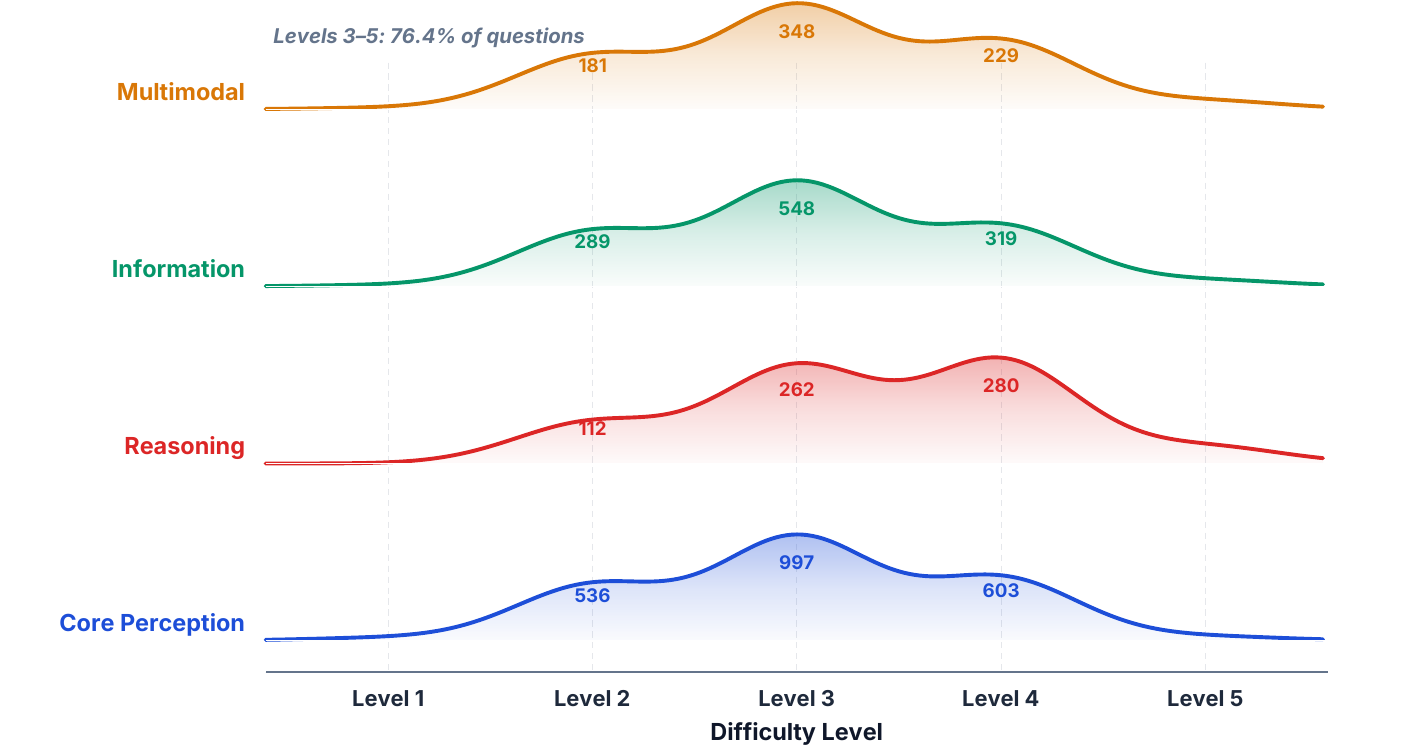}
  \caption{\textbf{Difficulty distribution by task category.} Smoothed ridge plot of question counts across difficulty Levels 1-5, separately for the four reporting categories (Core Perception, Reasoning, Information, Multimodal). All four categories peak at the moderate-to-hard end (Levels 3-4), with 76.4\% of questions overall falling in Levels 3-5. Reasoning questions skew hardest (peak at Level~4), Core Perception and Information are centered at Level~3, and the Multimodal category is also hard-skewed, complementing Fig.~\ref{fig:difficulty-modality}.}
  \label{fig:difficulty-ridge}
\end{figure}

Fig.~\ref{fig:sample-category-dist} visualizes the genre distribution of the source videos, Fig.~\ref{fig:difficulty-modality} characterizes how multimodal demand grows with question difficulty, and Fig.~\ref{fig:difficulty-ridge} shows the per-category difficulty distribution.

\section{Human Alignment Study (Extended)}
\label{sec:human_alignment}

\paragraph{Validation pool.}
The validation pool comprises \textbf{$400$ rubric criteria} drawn from LongShOTBench's hierarchical rubrics, stratified to span all four task categories (Core Perception, Reasoning, Information, Multimodal) and all three video-duration buckets (short / medium / long). The criteria are spread across $5$ representative systems (one closed-source API: Gemini-3.1-Pro-Preview; two omni-modal open-source models: Qwen3-Omni-30B-A3B-Thinking and Nemotron-Nano-Omni-30B-Reasoning; one frame-based omni-modal baseline: MiniCPM-o-4.5; and our LongShOTAgent), selected to cover the full performance range of the leaderboard so that the alignment study probes the verifiers across the regimes they will be deployed in. Each of the $400$ criteria is an independent atomic decision (``does the candidate answer match the gold reference answer with respect to this rubric criterion?''), so the effective sample size for criterion-level statistics is $400$, not the number of underlying questions. Four expert annotators independently graded the full pool. Annotators see exactly the same inputs as the LLM verifiers (question, gold reference answer, criterion list, and candidate answer) and toggle each criterion as met / not met based on the candidate-gold match, not on their own world knowledge.

\paragraph{Metrics and what they tell us.}
We focus on two metrics that operate directly on the $400$ criterion-level decisions and one metric that asks whether the resulting per-system ranking matches humans:
\begin{itemize}
  \item \textbf{Cohen's $\kappa$.} Chance-corrected agreement between two raters on the binary ``criterion met'' label across all $400$ decisions. $\kappa{=}0$ is chance, $\kappa{=}1$ is identical labels; the Landis-Koch ranges are $0.41$-$0.60$ \emph{moderate}, $0.61$-$0.80$ \emph{substantial}, $0.81$-$1.00$ \emph{almost perfect}~\citep{landis1977measurement}. \emph{What it tells us:} how often the rater's yes/no decision matches the reference, after correcting for the rate at which they would agree by chance.
  \item \textbf{F1 on the True class.} Harmonic mean of precision and recall when ``criterion met'' is the positive label, computed on the same $400$ decisions. \emph{What it tells us:} whether a verifier is systematically ``stingy'' (marks too many criteria as unmet) or ``lenient'' (marks too many as met); a failure mode that high accuracy alone can hide.
  \item \textbf{Model-ranking Spearman $\rho$.} After aggregating each rater's $400$ criterion-level decisions into a per-system score, $\rho$ asks whether verifiers and humans order the $5$ systems the same way. \emph{What it tells us} (and this is the leaderboard-level question): whether the choice of verifier can flip the leaderboard.
\end{itemize}
For human-vs-verifier comparisons we use the per-criterion human \emph{majority} as the binary label. The 3-verifier ensemble uses majority voting at the criterion level.

\paragraph{Per-system agreement.}
Table~\ref{tab:per_model_alignment} aggregates the $400$ criterion-level decisions into per-system scores and reports the 3-verifier mean against the human mean. Across all $5$ systems the absolute gap stays within $\sim 0.1$, with the 3-verifier mean systematically slightly more lenient than humans (every gap is positive, ranging from $+0.013$ on Qwen3-Omni-30B-A3B-T to $+0.090$ on Gemini-3.1-Pro-Preview). The system ranking is preserved up to two adjacent flips on near-tied pairs (Gemini vs LongShOTAgent and Nemotron-Nano-Omni vs Qwen3-Omni; both verifier-side gaps are $<\!0.04$, well below the per-system noise band), yielding Spearman $\rho{=}0.80$. Because the bias is approximately constant in sign, relative comparisons between systems with non-trivial gaps remain unbiased.
\begin{table}[t]
\centering
\small
\caption{\textbf{Per-model mean weighted score.} Mean over the 20-unit
human-validation subset. The 3-verifier mean tracks the human mean within
$\sim$0.1
across all 5 systems and preserves the human ranking up to near-tied adjacent pairs
(Spearman $\rho=0.800$).}
\label{tab:per_model_alignment}
\setlength{\tabcolsep}{4pt}
\resizebox{\textwidth}{!}{%
\begin{tabular}{lccccc}
\toprule
\textbf{Model} & GPT-5-mini & Gemma-4-31B & Qwen3-14B & \textbf{3-verifier mean} & \textbf{Human mean} \\
\midrule
LongShOTAgent & 0.591 & 0.592 & 0.591 & \textbf{0.591} & \textbf{0.534} \\
Qwen3-Omni-30B-A3B-T & 0.455 & 0.423 & 0.323 & \textbf{0.400} & \textbf{0.387} \\
Gemini-3.1-Pro-Preview & 0.660 & 0.617 & 0.544 & \textbf{0.607} & \textbf{0.517} \\
Nemotron-Nano-Omni-30B-R & 0.427 & 0.383 & 0.421 & \textbf{0.410} & \textbf{0.371} \\
MiniCPM-o-4.5 & 0.259 & 0.273 & 0.209 & \textbf{0.247} & \textbf{0.222} \\
\bottomrule
\end{tabular}}
\end{table}

\begin{figure}[t]
  \centering
  \includegraphics[width=1.0\linewidth]{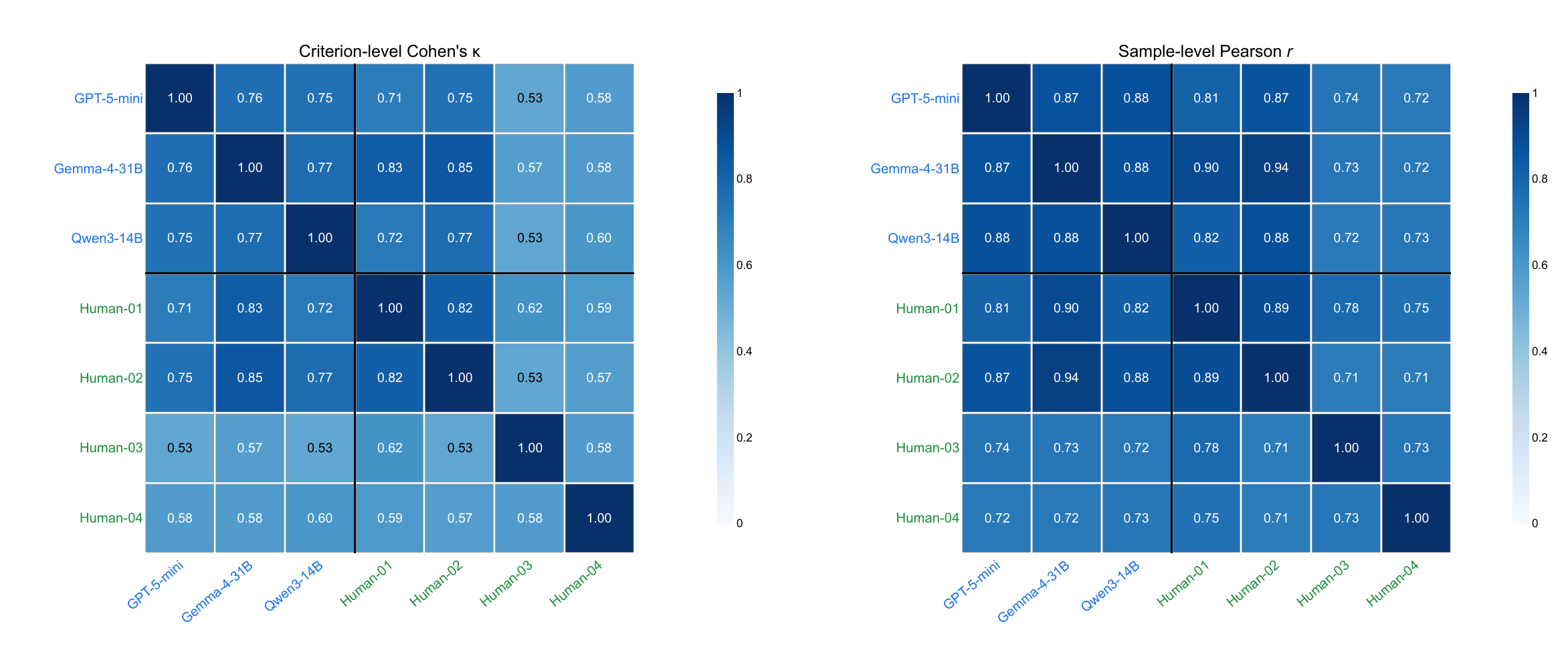}
  \caption{\textbf{Pairwise rater agreement.} Left: criterion-level Cohen's $\kappa$ on the $400$ rubric-criterion decisions. Right: continuous-score Pearson $r$ on the same set after aggregating into per-(question, model) weighted scores. Black lines separate the inter-verifier block (top-left $3\!\times\!3$) from the inter-human block (bottom-right $4\!\times\!4$); off-diagonal cells are cross verifier-human pairs. Inter-verifier cells sit in $[0.75, 0.77]$ and span the same range as the inter-human block ($[0.53, 0.82]$); the strongest cross verifier-human cell ($\kappa{=}0.85$) exceeds every inter-human pair, supporting the use of any single verifier for ranking and the 3-verifier mean for absolute scoring.}
  \label{fig:alignment_heatmap}
\end{figure}

\paragraph{Pairwise inter-rater agreement.}
Figure~\ref{fig:alignment_heatmap} shows the full $7\!\times\!7$ pairwise agreement matrix among the three LLM verifiers and the four dense human annotators, separated by a black divider into inter-verifier (top-left), inter-human (bottom-right), and cross verifier-human (off-diagonal) blocks. Three observations follow on the criterion-level $\kappa$ side. \emph{(1) Inter-verifier $\kappa$} sits in $[0.75, 0.77]$, substantial agreement that spans the same range as the pairwise inter-human band ($\kappa\!\in\![0.53, 0.82]$, mean $0.65$); after ensembling against the human majority, verifier-vs-human agreement ($\kappa{=}0.75$) sits at the upper edge of the inter-human distribution. \emph{(2) Cross verifier-human agreement} is uniformly substantial-to-strong on the more consistent annotators ($\kappa\!\in\![0.71, 0.85]$ on two of the four humans) and drops only on the noisier annotators ($\kappa\!\in\![0.53, 0.60]$ on the other two); the strongest cross-block cell ($\kappa{=}0.85$) is the highest agreement in the entire matrix, exceeding every inter-human pair. \emph{(3) No verifier is an outlier:} every verifier row stays in the same regime, so the validity conclusion does not hinge on any single verifier. The right panel ($r$) corroborates the same picture on the continuous side and is included for completeness.

\paragraph{Why this justifies the 3-verifier mean.}
Three independent pieces of evidence support reporting the 3-verifier mean as the headline metric while keeping individual verifiers as a fallback. (i) \emph{No single verifier dominates.} Gemma is the highest-$\kappa$ verifier on this $400$-criterion pool ($\kappa{=}0.74$), but its rank-correlation with humans across the $5$ systems is identical to Qwen3 and GPT-5-mini ($\rho{=}0.90$); in a different draw of criteria another verifier could win, and averaging removes that lottery. (ii) \emph{Ensembling tightens absolute calibration onto humans.} The ensemble lifts agreement above every individual verifier ($\kappa$: best individual $0.74\!\rightarrow$ ensemble $0.75$, F1: $0.84\!\rightarrow0.85$) and keeps system-level ranking shifts confined to near-tied adjacent pairs whose verifier-side gaps are below $0.04$ (Spearman $\rho{=}0.80$ at the $5$-system level). Because only $5$ systems are ranked, Spearman is coarse here (each adjacent transposition moves it by $\sim$0.1), so the ensemble's $\rho{=}0.80$ relative to the $0.90$ of the individual verifiers corresponds to a single additional swap between near-tied systems (verifier-side gap $<\!0.04$) rather than weaker ranking agreement. This is the property the leaderboard claim relies on: relative comparisons between non-tied systems are preserved. (iii) \emph{Inter-verifier agreement is substantial in its own right} ($\kappa\!\geq\!0.75$ pairwise, individual verifier ranking $\rho{=}0.90$), so single-verifier scores remain interpretable for cost-constrained ablations; the small numerical differences across verifiers do not change the qualitative conclusions.

\paragraph{Caveats.}
(1) The dense-human pool is small ($n{=}4$) because each annotation requires the same matched-input protocol given to the LLM verifiers, with every annotator grading the full $400$-criterion pool; the $400$ underlying criterion decisions and the explicit reporting of the full inter-human range alongside every aggregate are intended to keep the comparison interpretable despite the small rater count, and the pairwise heatmap further shows that no single annotator is an outlier. (2) Annotators evaluate under an identical setting to the LLM verifiers, receiving the same rubric, gold reference answer, and candidate answer, and judging match-to-gold rather than world-knowledge correctness; this matched-input protocol isolates the variable of interest, the fidelity of the rubric verification step itself, while video-grounding is enforced upstream during dataset construction by our hierarchical rubric pipeline.

\section{Detailed Benchmark Analysis}
\label{sec:detailed_benchmark_analysis}

The main text presents five diagnostic findings drawn from the 105-model evaluation on LongShOTBench. This section extends that analysis with additional breakdowns that further characterize the landscape of long-video understanding. All scores below are 3-verifier means unless otherwise noted.

\paragraph{Scaling Laws vs.\ Architectures.}
The results indicate that scaling laws in dense VLMs face severe diminishing returns for long-video tasks when compared to fundamentally shifting the architecture. For instance, the 235B parameter \texttt{Qwen3-VL 235B-T} reaches 47.23\%. In contrast, \texttt{Qwen3-Omni 30B-T}, a model an order of magnitude smaller but with explicit audio pathways, scores decisively higher at 64.05\%. Similarly, frame-based models like \texttt{Gemma-3 27B} (41.55\%) closely rival massive video-native models like \texttt{InternVL3.5 241B} (29.77\%). This strongly suggests the current bottleneck in long-video understanding lies in high-quality temporal indexing and multi-sensory alignment, rather than raw linguistic or visual semantic scaling.

\paragraph{The Audio Bottleneck in Omni-models.}
While omni-models successfully integrate audio, non-speech ambient audio remains their weakest link. For instance, breaking down the modality scores for \texttt{Qwen3-Omni 30B-T} shows Visual: 62.09\% vs Audio: 55.89\%. Even LongShOTAgent, utilizing dedicated audio specialists, exhibits a gap (Visual: 64.96\% vs Audio: 57.61\%). The industry has highly optimized ViT/SigLIP for vision and Whisper for implicit ASR, but grounding abstract acoustic data (like doors clicking or shifts in environmental tone) to specific visual timelines remains a major open challenge.

\begin{figure}[t]
\centering
\includegraphics[width=\textwidth]{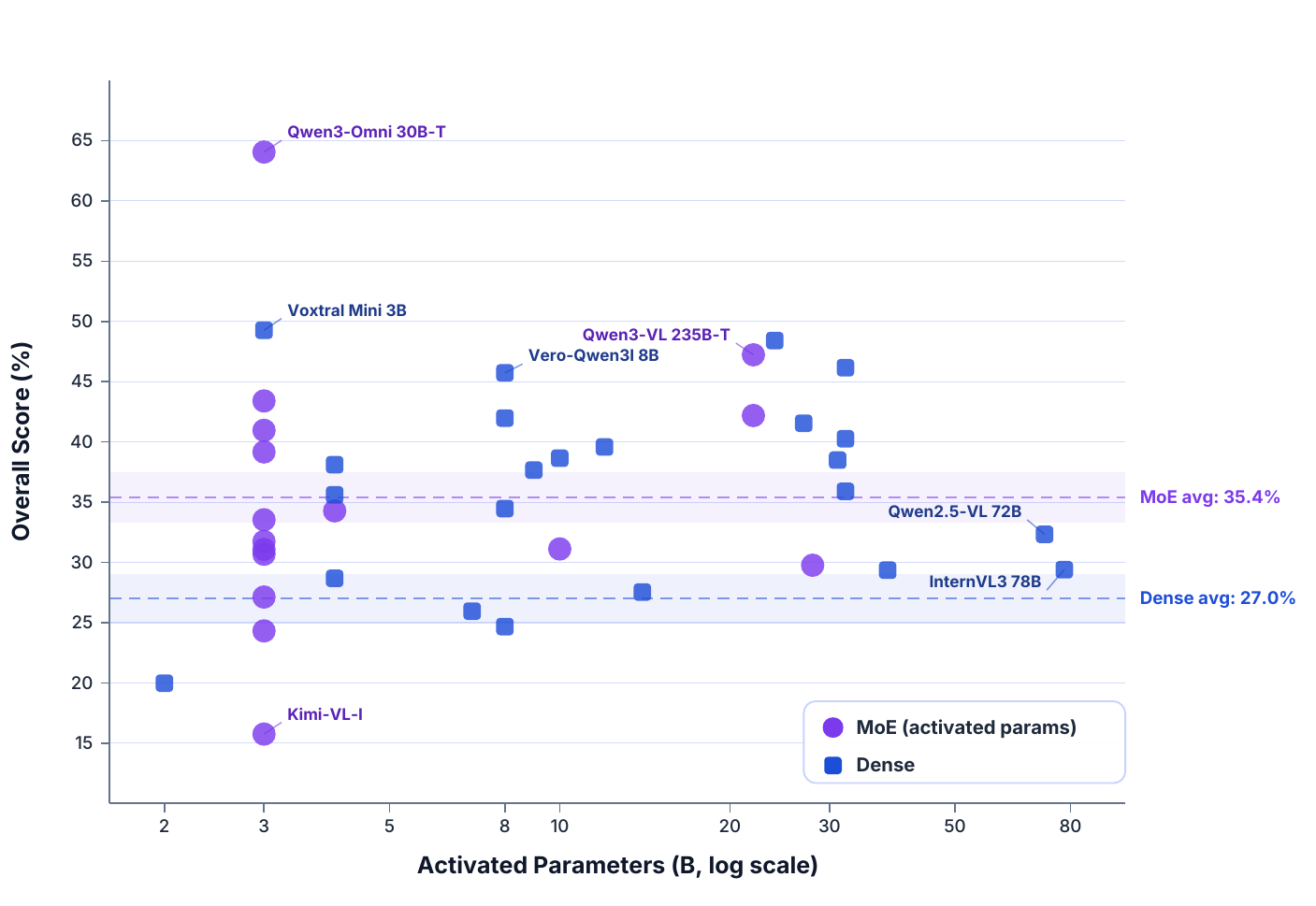}
\caption{\textbf{Parameter efficiency: Dense vs.\ MoE architectures.} Each point represents a model plotted by its activated parameter count (log scale) and overall 3-verifier mean score. Purple circles denote MoE models and blue squares denote dense models. Dashed horizontal lines mark the architecture-level averages. MoE models consistently outperform dense models at matched inference cost, and dense scaling shows diminishing returns beyond approximately 30B parameters.}
\label{fig:param_efficiency}
\end{figure}

\paragraph{Parameter Efficiency: Dense vs.\ Sparse Architectures.}
A natural follow-up to the architecture-versus-scale finding is whether the type of scaling itself matters. Figure~\ref{fig:param_efficiency} plots overall score against \emph{activated} parameters for all open-weight models in the benchmark, distinguishing dense architectures from Mixture-of-Experts (MoE) models. Across the full leaderboard, MoE models average 35.4\% compared to 27.0\% for dense models, an absolute gap of over eight points. This advantage is not merely a function of total parameter count. At three billion activated parameters, the MoE model \texttt{Qwen3-Omni 30B-T} reaches 64.05\%, while no dense model below 30B exceeds 50\%. Even at 72B parameters, the dense \texttt{Qwen2.5-VL 72B} reaches only 32.33\%, falling well short of several MoE models that activate only a fraction of their weights.

Two observations stand out. First, the MoE advantage is especially pronounced for models that combine sparse routing with thinking-mode training. \texttt{Qwen3-Omni 30B-T} and \texttt{Qwen3-VL 30B-A3B-T} both activate only 3B parameters yet rank among the top five open-weight systems, suggesting that sparse routing and test-time reasoning are complementary rather than redundant. Second, the dense trend line is essentially flat beyond 30B activated parameters, indicating that simply widening or deepening a dense model provides diminishing returns for long-video understanding. Together, these patterns suggest that sparse architectures offer a more efficient path toward long-video competence, likely because the additional expert capacity can be devoted to modality-specific or temporally-specialized sub-networks without inflating inference cost.

\paragraph{Best Models per Parameter Tier.}
To offer practical guidance for deployment under different compute budgets, Table~\ref{tab:cost_eff} identifies the best-performing model within each activated-parameter tier. The results reveal a striking inversion of the expected cost-performance relationship. At the lowest inference tier (under 4B activated parameters), MoE models with thinking-mode training dominate. \texttt{Qwen3-Omni 30B-T}, which activates only 3B parameters per token, achieves 64.05\% and outperforms every model in every larger tier. In the 8-15B range, \texttt{Vero-Qwen3I 8B} (45.7\%) offers strong performance at moderate cost. The 40B+ tier is notably weak: the best model, \texttt{Qwen2.5-VL 72B} at 32.3\%, scores roughly half of the best sub-4B model.

This inversion reinforces the central lesson of the benchmark. For long-video understanding, the combination of architecture choice (MoE vs.\ dense), training strategy (thinking vs.\ instruct), and modality coverage (omni-modal vs.\ vision-only) matters far more than raw model size. The table serves as a reference for practitioners selecting models under real-world deployment constraints, and further demonstrates that LongShOTBench differentiates models along dimensions that parameter count alone cannot predict.

\begin{table}[t]
\centering
\caption{\textbf{Best model per activated-parameter tier.} For each tier, we report the top-scoring model by 3-verifier mean overall. The sub-4B tier outperforms all larger tiers, driven by MoE models with thinking-mode training.}
\label{tab:cost_eff}
\footnotesize
\setlength{\tabcolsep}{5pt}
\begin{tabular}{llccl}
\toprule
\textbf{Tier} & \textbf{Best Model} & \textbf{Act.\ Params} & \textbf{Overall (\%)} & \textbf{Architecture} \\
\midrule
$<$4B   & Qwen3-Omni 30B-T        & 3B   & 64.05 & MoE + Thinking \\
4-8B    & Qwen3-VL 4B-T           & 4B   & 38.11 & Dense + Thinking \\
8-15B   & Vero-Qwen3I 8B          & 8B   & 45.73 & Dense + Thinking \\
15-40B  & Voxtral-Small 24B       & 24B  & 48.41 & Dense + Audio \\
40B+    & Qwen2.5-VL 72B-I        & 72B  & 32.33 & Dense \\
\bottomrule
\end{tabular}
\end{table}

\begin{figure}[t]
\centering
\includegraphics[width=\textwidth]{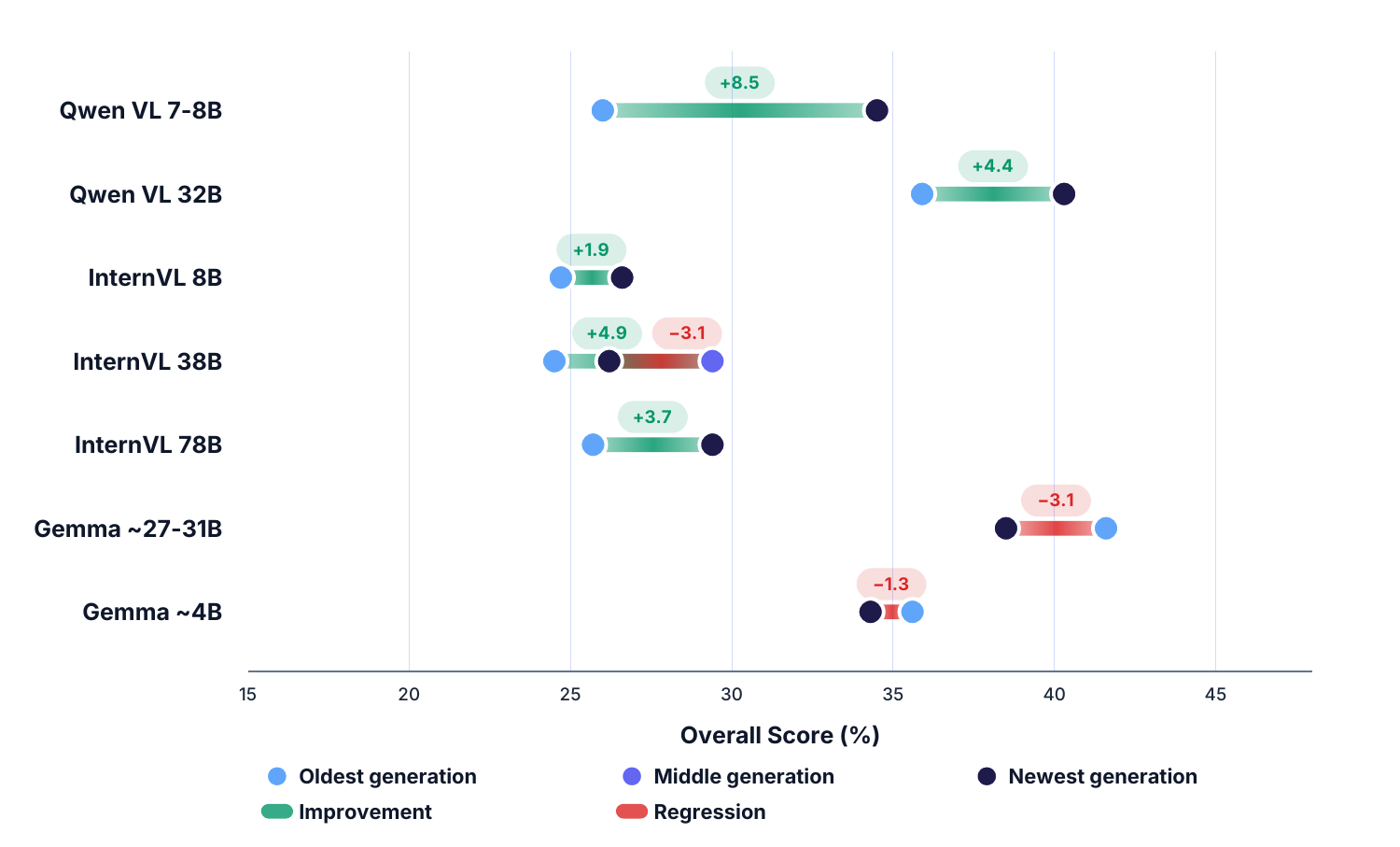}
\caption{\textbf{Generational progress across model families.} Connected dots compare successive model generations at matched parameter scales. Delta annotations (green = improvement, red = regression) show the score change between generations. Qwen and InternVL show consistent gains, while Gemma-4 regresses relative to Gemma-3 at both scales tested.}
\label{fig:generational}
\end{figure}

\paragraph{Generational Progress Across Model Families.}
LongShOTBench covers multiple generations of the same model family at comparable parameter counts, enabling controlled measurement of how ongoing model development translates into long-video performance. Figure~\ref{fig:generational} presents seven such head-to-head comparisons.

The Qwen family shows the most consistent improvement. At the 7-8B scale, \texttt{Qwen3-VL 8B-I} (34.5\%) improves by 8.5 points over \texttt{Qwen2.5-VL 7B-I} (26.0\%). The 32B pair shows a gain of 4.4 points. These improvements are directionally consistent with gains reported on short-video and image benchmarks, but the magnitude on LongShOTBench is larger, suggesting that newer Qwen models have made targeted improvements in long-context processing. The InternVL series also shows moderate but consistent gains from version 2.5 to 3.0, with the 38B variant improving by 4.9 points and the 78B variant by 3.7 points.

A notable finding concerns the Gemma family. \texttt{Gemma-3 27B} (41.6\%) outscores \texttt{Gemma-4 31B} (38.5\%) by 3.1 points, and the 4B comparison shows a similar reversal of 1.3 points. This regression may reflect different training-data compositions or architectural trade-offs in the newer generation that improve short-context benchmarks at the expense of sustained temporal reasoning. These results highlight that generational progress is not uniform across families and that long-video performance is a useful independent axis for tracking model development, one that does not always correlate with improvements on established short-context benchmarks.

\begin{figure}[t]
\centering
\includegraphics[width=\textwidth]{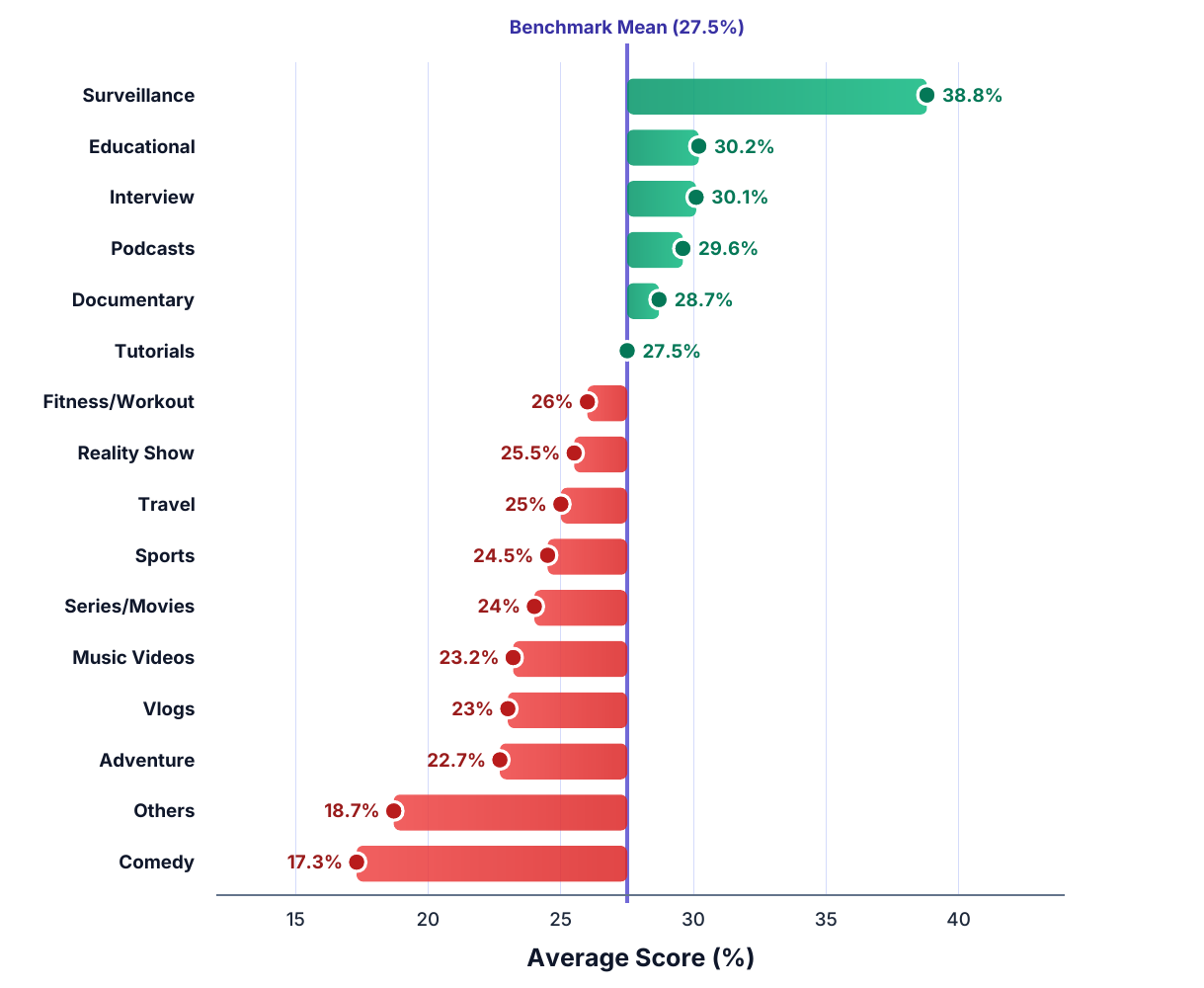}
\caption{\textbf{Performance by video genre.} Diverging bars show the average 3-verifier mean across all 105 models for each genre, sorted from highest to lowest. The vertical line marks the benchmark mean (27.5\%). The 21.5-point spread between Surveillance and Comedy highlights genre as a meaningful difficulty axis.}
\label{fig:genre_perf}
\end{figure}

\paragraph{Content Genre as a Difficulty Axis.}
Video genre introduces a natural source of difficulty variation that is orthogonal to task type, modality, and duration. Figure~\ref{fig:genre_perf} reports the average score across all 105 models for each of the 16 genres represented in LongShOTBench.

Surveillance videos are the easiest genre (38.8\%), likely because they feature repetitive and predictable scenes with limited semantic complexity. Educational content (30.2\%) and Interviews (30.1\%) also fall above the benchmark mean of 27.5\%, consistent with the structured and explicit information delivery typical of these formats. At the other end, Comedy is the hardest genre by a wide margin at 17.3\%. Understanding humor requires reasoning about timing, cultural context, and the interplay between audio cues and visual gags, all of which present challenges for current multimodal models. Adventure (22.7\%) and Vlogs (23.0\%) are similarly difficult, as they feature fast-paced editing, ambient audio that carries narrative weight, and visual content that shifts rapidly between settings.

This spread of over 20 points between the easiest and hardest genres demonstrates that LongShOTBench captures real-world difficulty variation beyond what task labels alone convey. A model that performs well on Documentary content is not guaranteed to transfer that competence to Comedy or Vlogs, which reinforces the value of genre diversity in long-video evaluation and provides an additional diagnostic lens for understanding model capabilities.

\begin{figure}[t]
\centering
\includegraphics[width=0.9\textwidth]{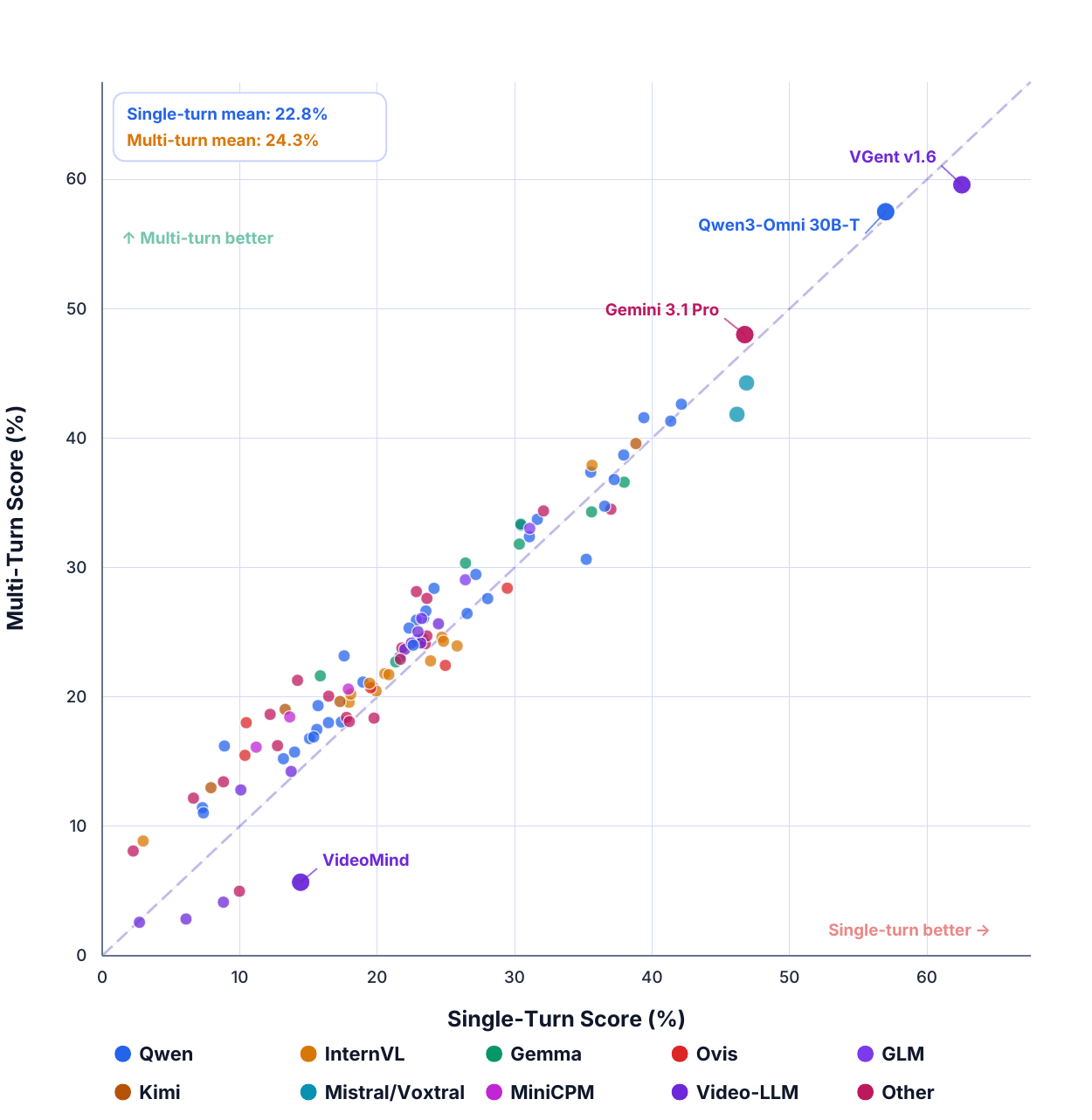}
\caption{\textbf{Single-turn vs.\ multi-turn performance.} Each point is a model, colored by family. The dashed diagonal represents equal performance across both formats. Most models cluster near parity, consistent with the ideal-trajectory evaluation design. Top-tier models perform consistently across both protocols.}
\label{fig:single_multi}
\end{figure}

\paragraph{Conversational Robustness: Single-Turn vs.\ Multi-Turn.}
LongShOTBench includes both single-turn and multi-turn evaluation protocols, enabling a direct comparison of how models handle isolated questions versus multi-step conversational probing over the same video content. Figure~\ref{fig:single_multi} plots each model's single-turn score against its multi-turn score.

The aggregate picture is one of near-parity. The single-turn mean across all 105 models is 22.8\% and the multi-turn mean is 24.3\%, a gap of less than two points. Most models cluster tightly around the diagonal in Figure~\ref{fig:single_multi}, indicating that the two formats are comparably difficult on average. This near-parity is partly by design. Multi-turn evaluation in LongShOTBench uses an ideal-trajectory setting where each follow-up turn receives the gold response from the previous turn as context, rather than the model's own output. This choice isolates per-turn response quality from error accumulation across turns, ensuring that the multi-turn protocol tests the model's ability to build on correct context rather than to recover from its own mistakes.

Several informative outliers emerge. On the multi-turn side, \texttt{OmniAtlas 7B} and \texttt{Ovis2 8B} show multi-turn scores 7-9 points above their single-turn performance, suggesting that these models benefit from the additional conversational context provided by the ideal trajectory. On the single-turn side, \texttt{VideoMind} shows an 11.6-point drop in multi-turn, possibly reflecting limitations in maintaining coherence across sequential tool-calling rounds. Top-tier models such as LongShOTAgent and \texttt{Qwen3-Omni 30B-T} perform consistently across both formats, indicating that strong long-video reasoning generalizes well across evaluation protocols. This consistency further validates the benchmark's design: the single-turn and multi-turn components measure complementary but non-redundant aspects of video understanding, and together they provide a more complete picture of model capability than either format alone.

\section{MCQ Robustness Analysis}
\label{sec:robustness}

Standard multiple-choice benchmarks such as Video-MME~\citep{fu2024video} for video understanding and WorldSense~\citep{hong2025worldsense} for omni-modal (video, audio, and speech) understanding are widely used to rank multimodal models.
However, MCQ formats allow models to exploit option-level shortcuts such as positional bias, lexical overlap between the question stem and the correct option, and process-of-elimination heuristics, which inflate accuracy without requiring genuine understanding.
To quantify this effect, we evaluate LongShOTAgent and three strong baselines under four progressively harder evaluation protocols on both benchmarks.
\textbf{Standard MCQ} keeps the original 4-option format with options in their canonical order.
\textbf{Open-ended} uses a two-pass procedure: in the first pass the model is shown the question with all options removed and produces a free-form answer; in the second pass the model is shown its own first-pass answer together with the original 4 options and selects the option that best matches it. This keeps scoring under the standard MCQ exact-match protocol while still preventing the model from exploiting option-level cues during the actual reasoning step.
\textbf{Shuffled options} randomly permutes the 4 options to neutralize positional bias.
\textbf{Same order + NOTA/AOTA at end} preserves the original option order but appends ``None of the Above'' (NOTA) and ``All of the Above'' (AOTA) as the last two choices; by construction the gold answer remains one of the original four, so this protocol probes whether models can resist the lure of plausible-sounding novel options rather than testing additional knowledge.

\noindent
Table~\ref{tab:robustness} reports accuracy (\%) under each protocol, with each cell annotated by its $\Delta$ relative to that method's Standard-MCQ score (\textcolor{red!70!black}{$\downarrow$} = drop, \textcolor{green!50!black}{$\uparrow$} = gain). 
{Qwen3.5-35B-A3B is a vision-language model with no audio encoder; on WorldSense, which supplies native audio, it is therefore evaluated without audio.}

\begin{table}[t]
\centering
\caption{\textbf{Robustness under four evaluation protocols (\%).}
Std.\ MCQ is the canonical baseline; other columns show $\Delta$ from that baseline (\textcolor{red!70!black}{$\downarrow$}: drop, \textcolor{green!50!black}{$\uparrow$}: gain). \textbf{Bold} / \underline{underline} mark per-column best / second-best; ``n/a'' indicates not evaluated. \textsuperscript{$\dagger$}Author-published Std.\ MCQ scores; all other values are from our runs.}
\label{tab:robustness}
\setlength{\tabcolsep}{3pt}
\renewcommand{\arraystretch}{1.05}
\scriptsize
\newcommand{\dn}[1]{\,\textcolor{red!70!black}{\tiny$\downarrow$#1}}
\newcommand{\up}[1]{\,\textcolor{green!50!black}{\tiny$\uparrow$#1}}
\newcolumntype{Y}{>{\centering\arraybackslash}X}
\begin{tabularx}{\textwidth}{@{}>{\raggedright\arraybackslash}p{2.9cm}|YYYY|YYYY@{}}
\toprule
& \multicolumn{4}{c|}{\textbf{Video-MME (w/o sub)}} & \multicolumn{4}{c}{\textbf{WorldSense}} \\
\cmidrule(lr){2-5}\cmidrule(lr){6-9}
\textbf{Method}
& \makecell{\textbf{Std.}\\\textbf{MCQ}} & \makecell{\textbf{Open-}\\\textbf{ended}} & \textbf{Shuf.} & \textbf{+NOTA}
& \makecell{\textbf{Std.}\\\textbf{MCQ}} & \makecell{\textbf{Open-}\\\textbf{ended}} & \textbf{Shuf.} & \textbf{+NOTA} \\
\midrule
\textbf{LongShOTAgent (Ours)}
& \underline{71.24}
& \textbf{69.91}\dn{1.3}
& \underline{68.11}\dn{3.1}
& \textbf{66.09}\dn{5.2}
& 46.56
& \textbf{43.38}\dn{3.2}
& 50.70\up{4.1}
& \underline{45.43}\dn{1.1} \\
Qwen3.5-35B-A3B
& \textbf{78.9}\textsuperscript{$\dagger$}
& \underline{58.40}\dn{20.5}
& \textbf{69.41}\dn{9.5}
& \underline{63.22}\dn{15.7}
& 42.78 & 36.41\dn{6.4} & 36.16\dn{6.6} & 41.55\dn{1.2} \\
Qwen3-Omni-30B Instruct
& 70.5\textsuperscript{$\dagger$}
& 51.81\dn{18.7}
& 33.19\dn{37.3}
& 29.67\dn{40.8}
& \textbf{54.0}\textsuperscript{$\dagger$}
& \underline{41.68}\dn{12.3}
& \textbf{52.43}\dn{1.6}
& \textbf{45.78}\dn{8.2} \\
Qwen3-Omni-30B Thinking
& 69.7\textsuperscript{$\dagger$}
& 56.11\dn{13.6}
& 62.93\dn{6.8}
& 35.04\dn{34.7}
& \underline{52.65}
& 33.58\dn{19.1}
& \underline{50.90}\dn{1.8}
& 41.30\dn{11.4} \\
\bottomrule
\end{tabularx}
\end{table}

\paragraph{MCQ scores overestimate monolithic VLMs.}
On standard Video-MME, the strongest monolithic baseline (Qwen3.5-35B-A3B) reports 78.9\%, almost 8 points above LongShOTAgent's 71.24\%.
However, when we strip the options and evaluate open-ended, Qwen3.5 drops to 58.4\% ($\Delta{=}-20.5$\,pp), Qwen3-Omni-Instruct to 51.81\% ($-18.7$\,pp), and Qwen3-Omni-Thinking to 56.11\% ($-13.6$\,pp), while LongShOTAgent retains 69.91\%, a drop of only 1.3\,pp from its MCQ score and the smallest open-ended degradation among all evaluated systems by a wide margin.
A substantial fraction of end-to-end VLM accuracy on MCQ benchmarks therefore derives from option-level cues rather than from genuine video comprehension; agentic composition, in which the final answer is generated from retrieved evidence rather than selected from a list, is far less affected.

\paragraph{Monolithic models are fragile under option perturbation.}
Shuffling options on Video-MME collapses Qwen3-Omni-Instruct from 70.5\% to 33.19\% ($-53\%$ relative), an indication that the model is heavily anchored on canonical option positions.
Appending NOTA/AOTA distractors while preserving the original order is even more punishing for the omni-models: Qwen3-Omni-Instruct falls to 29.67\% ($-58\%$ relative) and Qwen3-Omni-Thinking to 35.04\% ($-50\%$ relative).
The vision-language model Qwen3.5-35B-A3B is moderately robust to shuffling ($-9.5$\,pp) but still loses 15.7\,pp under NOTA/AOTA, confirming that NOTA-resistance is far from automatic even for strong vision-language models.
LongShOTAgent is the most robust system in every perturbation column on Video-MME: it retains 68.11\% under shuffling ($-3.1$\,pp) and 66.09\% under Same+NOTA ($-5.2$\,pp), drops an order of magnitude smaller than the monolithic baselines.
This robustness stems from the agent answering from grounded multimodal evidence first and only mapping to the option set as a final, isolated step.

\paragraph{WorldSense reveals a modality gap that closes under realistic evaluation.}
On the audio-centric WorldSense, both Omni models leverage their native audio encoders to outperform the agent under Standard MCQ (54.0\% and 52.65\% vs.\ 46.56\%).
This advantage disappears under open-ended evaluation: LongShOTAgent leads with 43.38\%, while Qwen3-Omni-Instruct drops to 41.68\% (closest competitor; $-12.3$\,pp from its MCQ score) and Qwen3-Omni-Thinking collapses to 33.58\% ($-19.1$\,pp).
Strikingly, the agent is the only method whose accuracy actually \emph{rises} under shuffled options on WorldSense ($46.56\rightarrow 50.70$\%, $+4.1$\,pp), suggesting that its pipeline does not depend on canonical option order at all, while the omni-models lose 1.6 to 1.8\,pp.
Under Same+NOTA the agent again has the smallest degradation ($-1.1$\,pp) compared with Qwen3-Omni-Instruct ($-8.2$\,pp) and Qwen3-Omni-Thinking ($-11.4$\,pp).
The agent's modular audio path (faster-whisper ASR plus Audio Flamingo 3) closes most of the native-audio gap once option-level shortcuts are removed.

These results indicate that MCQ accuracy alone is a poor proxy for video understanding.
Models that appear strong on standard MCQ benchmarks may be substantially less capable than their scores imply, relying on option-level heuristics that do not transfer to realistic, open-ended settings.
Open-ended and adversarial MCQ protocols, such as the rubric-graded, free-form evaluation used throughout LongShOTBench, are essential for reliable model comparison.
The leaderboard position of LongShOTAgent on our benchmark (Table~\ref{tab:main_benchmark_table}, 66.64\% overall) is consistent with this analysis: LongShOTBench cannot be gamed through option exploitation, and under that protocol the agentic approach outperforms all monolithic alternatives we evaluated.

\section{Additional Details of LongShOTAgent}
\label{sec:agent_details}

\paragraph{Motivation \& Advantages.}
The proposed agentic pipeline is not solely a baseline but a demonstration of how structured orchestration can overcome the limitations of massive monolithic scaling. Its fundamental advantages lie in:
\begin{itemize}
    \item \textbf{Bypassing the brute-force context window.} Rather than attempting to compress hours of video representations into a single long-context pass (which suffers from catastrophic "lost-in-the-middle" effects and exorbitant inference costs), LongShOTAgent utilizes a semantic index. It retrieves and refines only the specific 60-second clips required by the prompt, preserving high-resolution temporal granularity.
    \item \textbf{Native weaponization of Test-Time Compute.} While monolithic models force a single forward-pass deduction, LongShOTAgent operates an explicit \textit{search-refine-verify} loop. It mimics human-like analytical workflows, gathering specialized evidence and verifying claims before committing to a final answer.
    \item \textbf{Domain Specialization.} Rather than diluting a single set of model weights to learn both visual spatial mapping and acoustic ambiance, it farms the work sequentially to unimodal experts (SigLIP, Whisper, Audio Flamingo), only using the main LLM (Qwen3.6 35B) as an intelligent orchestrator.
\end{itemize}
These architectural decisions allow LongShOTAgent (66.64\%) to outperform both state-of-the-art open omni-models and proprietary APIs without any additional training or private data.

\paragraph{System Overview.}
LongShOTAgent is a modular, training-free system in which an LLM orchestrator issues structured tool calls in a ReAct-style loop over a per-video multimodal index. The same Qwen3.6-35B-A3B checkpoint serves both as the reasoning orchestrator and as the vision-language backbone used during segment refinement; an audio-language model handles audio/speech segments. Each component corresponds to a well-defined sub-capability for long-horizon video understanding:

\begin{itemize}
    \item \textbf{Reasoning Orchestrator:} Qwen3.6-35B-A3B served via vLLM. The orchestrator emits OpenAI-style function calls and maintains a stateful conversation history per video; tool calls may be batched and executed concurrently.
    \item \textbf{Vision-Language Module:} The same Qwen3.6-35B-A3B checkpoint is invoked through a separate vLLM endpoint for clip-level semantic grounding and dense scene description on segments returned by retrieval.
    \item \textbf{Audio-Language Module:} Audio Flamingo 3 (AF3)~\citep{goel2025audio} performs speech and non-speech acoustic understanding on retrieved audio segments. The orchestrator selects the visual or audio branch based on the inferred dominant modality of the query.
    \item \textbf{Speech Transcription:} faster-whisper Large-v3 produces timestamped transcripts during indexing; long-form transcription runs in batched mode on GPU when available.
    \item \textbf{Retrieval and Memory Module:} A per-video vector database (ChromaDB) holds visual frame embeddings from ViT-B-16 SigLIP-512 (sampled at 5\,FPS at $512{\times}512$) together with text embeddings (sentence-transformers \texttt{all-MiniLM-L6-v2}) of speech transcripts and non-speech audio descriptions. Cross-modal queries are encoded into the appropriate embedding space at retrieval time.
    \item \textbf{Tool Surface:} The orchestrator is exposed exactly three tools: \texttt{search\_video} (top-$k$ semantic retrieval over the audio/visual indices), \texttt{refine\_video} (extracts a $\le$60-second segment and re-analyses it with the VLM or ALM), and \texttt{verify\_claim} (verifies a textual claim against the visual scene). The ReAct loop is bounded by a configurable iteration limit and de-duplicates repeated retrievals across rounds.
\end{itemize}

\paragraph{Execution Dynamics.}
The agent follows a two-stage process. First, during \textit{indexing} (run once per video), audio is extracted with FFmpeg and transcribed by faster-whisper while frames are sampled at 5\,FPS / $512{\times}512$ and embedded by SigLIP; both streams populate the per-video vector database in a producer-consumer pipeline so that ingestion can run without buffering the full video. Second, during \textit{tool-calling reasoning}, the orchestrator iteratively calls \texttt{search\_video} to surface candidate segments, optionally drills in with \texttt{refine\_video} on a $\le$60\,s window, and uses \texttt{verify\_claim} to check existence claims; intermediate outputs are kept in the conversation history so the orchestrator can refine hypotheses across rounds before producing a final answer.

\paragraph{Fairness and Scope.}
The agentic pipeline is not part of the primary model benchmarking. It serves as a system-level demonstration of compositional reasoning under the same video-only input protocol and within the same computational constraints as the benchmarked models. Because modules are invoked conditionally and never run concurrently, parameter summation across components is not a meaningful fairness metric. The key measure of efficiency lies in selective activation and modular coordination rather than in total parameter count. All components are frozen and publicly available, ensuring reproducibility and transparency. The improvement over single end-to-end models arises solely from efficient orchestration, not from additional data or tuning.

\paragraph{Interpretation.}
The results demonstrate that agentic modularity can effectively compensate for model scale. By decomposing reasoning into targeted, verifiable sub-steps, the system achieves competence close to large proprietary models while remaining interpretable and resource-efficient. This suggests that progress in multimodal understanding can emerge not only from larger architectures but also from advances in structured coordination and agentic design.

\section{Limitations}
\label{sec:limitations}

While LongShOTBench advances long-form, omni-modal video evaluation, several limitations remain.
First, evaluation relies on a three-model rubric-verification ensemble (Qwen3-14B, Gemma-4-31B-it, GPT-5-mini) that grounds every decision against the human-validated gold answer rather than the verifier's own world knowledge; although the per-criterion match-to-gold protocol and verifier averaging improve reliability over single free-form LLM grading, residual model-specific verifier biases and prompt sensitivity may still affect absolute scores.
Second, due to the substantial cost of long-video inference, we run a single generation pass per model and report the mean of the three verifiers' overall scores; per-verifier values and inter-verifier standard deviation are provided in the supplementary, but model-side variance from sampling is not explicitly quantified.
Third, our native video-only protocol delegates frame sampling to each model's documented defaults, which excludes commercial APIs that require client-side frame extraction (e.g., GPT-4o); this prevents us from directly benchmarking those systems under identical conditions.
Fourth, the dataset is sourced from Video-MME and YouTube and is predominantly English; coverage of other languages, cultures, and low-resource domains remains limited.
Finally, the construction pipeline itself depends on a single LLM family (Qwen3) for scenario framing, question, answer, and rubric generation, so any systematic biases in those models can propagate into the benchmark; the human-validation stage is intended to mitigate but not fully eliminate this risk.

\end{document}